\documentclass[acmsmall]{acmart}

\AtBeginDocument{%
  \providecommand\BibTeX{{%
    \normalfont B\kern-0.5em{\scshape i\kern-0.25em b}\kern-0.8em\TeX}}}

\setcopyright{cc}
\setcctype{by-nc-sa}
\acmJournal{CSUR}
\acmYear{2025} \acmVolume{1} \acmNumber{1} \acmArticle{1} \acmMonth{1} \acmPrice{}\acmDOI{10.1145/3716846}

\newcommand{\etal}{\textit{et al}.}
\newcommand{\etc}{\textit{etc}.}
\newcommand{\ie}{\textit{i}.\textit{e}.,}
\newcommand{\eg}{\textit{e}.\textit{g}.,}
\newcommand{\STAB}[1]{\begin{tabular}{@{}c@{}}#1\end{tabular}}

\usepackage{url}
\usepackage{breakurl}
\usepackage{amsthm}
\usepackage{enumitem}
\usepackage{wrapfig}
\usepackage{booktabs}
\usepackage{graphicx}
\usepackage{multirow}
\usepackage{float}
\usepackage{array}
\usepackage{longtable}
\usepackage{makecell}
\usepackage{breakcites}
\usepackage[flushleft]{threeparttable}
\usepackage{afterpage}
\usepackage{caption}

\theoremstyle{definition}
\newtheorem{definition}{Definition}[section]

\begin{document}

\title[Hallucination Detection in Foundation Models for Decision-Making]{Hallucination Detection in Foundation Models for Decision-Making: A Flexible Definition and Review of the State of the Art}

\author{Neeloy Chakraborty}
\email{neeloyc2@illinois.edu}
\affiliation{%
  \institution{University of Illinois Urbana-Champaign}
  \department{Department of Electrical and Computer Engineering}
  \department{Coordinated Science Laboratory}
  \city{Urbana}
  \state{IL}
  \country{USA}
}
\author{Melkior Ornik}
\email{mornik@illinois.edu}
\affiliation{%
  \institution{University of Illinois Urbana-Champaign}
  \department{Department of Aerospace Engineering}
  \department{Talbot Laboratory}
  \city{Urbana}
  \state{IL}
  \country{USA}
}
\author{Katherine Driggs-Campbell}
\email{krdc@illinois.edu}
\affiliation{%
  \institution{University of Illinois Urbana-Champaign}
  \department{Department of Electrical and Computer Engineering}
  \department{Coordinated Science Laboratory}
  \city{Urbana}
  \state{IL}
  \country{USA}
}

\renewcommand{\shortauthors}{Chakraborty~\etal}

\begin{abstract}
Autonomous systems are soon to be ubiquitous, spanning manufacturing, agriculture, healthcare, entertainment, and other industries.
Most of these systems are developed with modular sub-components for decision-making, planning, and control that may be hand-engineered or learning-based.
While these approaches perform well under the situations they were specifically designed for, they can perform especially poorly in out-of-distribution scenarios that will undoubtedly arise at test-time.
The rise of foundation models trained on multiple tasks with impressively large datasets has led researchers to believe that these models may provide ``common sense'' reasoning that existing planners are missing, bridging the gap between algorithm development and deployment.
While researchers have shown promising results in deploying foundation models to decision-making tasks, these models are known to hallucinate and generate decisions that may sound reasonable, but are in fact poor.
We argue there is a need to step back and simultaneously design systems that can quantify the certainty of a model's decision, and detect when it may be hallucinating.
In this work, we discuss the current use cases of foundation models for decision-making tasks, provide a general definition for hallucinations with examples, discuss existing approaches to hallucination detection and mitigation with a focus on decision problems, present guidelines, and explore areas for further research in this exciting field.
\end{abstract}

\thanks{This work is supported by the \grantsponsor{0}{Office of Naval Research}{} under Grant No.:~\grantnum{0}{N00014-23-1-2651}.}



\received{29 April 2024}
\received[revised]{13 January 2025}
\received[accepted]{3 February 2025}

\maketitle

\section{Introduction}
\label{sec:intro}

A great deal of progress has been made in the last decade and a half with regards to the efficacy and efficiency of models for perception, decision-making, planning, and control~\cite{soori2023artificial, janai2020computer}.
Broadly speaking, approaches to these problems fall under one of two umbrellas: hand-engineered model-based systems and data-driven learning-based models~\cite{formentin2013model}.
With some deployment scenario in mind, developers may hand-engineer rules~\cite{hayes1985rule} or tune a controller~\cite{borase2021review} to be tested, or in the case of learning-based models, collect training data and craft some reward function to fit a model to an objective, given said data~\cite{henderson2018deep}.
In practice, these methods work particularly well in the scenarios that they were specifically designed and trained for, but may produce undesirable results in previously unseen out-of-distribution cases~\cite{wen2023road}.
Designers may choose to add more rules, re-tune their controller, fine-tune their model to a more representative dataset, fix the reward function to handle edge cases, or even add a detector (which may itself be rule-based or data-driven) at test-time to identify out-of-distribution scenarios before calling on the decision-maker~\cite{singer2021framework, schreiber2023attentional, chakraborty2023structural}.
However, even with these changes, there will always be other situations that designers had not previously considered which will come about during deployment, leading to sub-optimal performance or critical failures.
Furthermore, the modifications made to the model may have unforeseen effects at test-time like undesired conflicting rules~\cite{ekenberg2000logic} or catastrophic forgetting of earlier learned skills~\cite{kemker2018measuring}. 

Informally, classical methods and data-driven approaches lack some form of \emph{common sense} that humans use to adapt in unfamiliar circumstances~\cite{fu2023drive}.
More recently, there has been work towards developing multi-modal large language models, which take inputs in the form of images, videos, audio, and text, to tackle more complex language understanding and reasoning tasks~\cite{bommasani2022opportunities, bai2024hallucination}.
These models are developed by collecting and cleaning an enormous natural language dataset, pre-training to reconstruct sentences on said dataset, fine-tuning on specific tasks (\eg~question-answering), and applying human-in-the-loop reinforcement learning to produce more reasonable responses~\cite{achiam2023gpt}.
Even though these models are another form of data-driven learning that attempt to maximize the likelihood of generated text conditioned on a given context, researchers have shown that they have the ability to generalize to tasks they have not been trained on, and reason about their decisions. 
Many researchers are specifically exploring the use of large (visual) language models, L(V)LMs, to fill the knowledge gap found in earlier works~\cite{cui2024survey}.
As such, these \emph{foundation} models are being tested in tasks like simulated decision-making~\cite{huang2024understanding} and real-world robotics~\cite{zeng2023large} to take the place of perception, planning, and control modules.
Even so, foundation models are not without their limitations.
Specifically, these models have a tendency to \emph{hallucinate},~\ie~generate decisions or reasoning that sound plausible, but are in fact inaccurate or would result in undesired effects in the world~\cite{ji2023survey,rawte2023survey,dai2023plausible}. 
This phenomenon has led to the beginning of a new research direction that attempts to detect when L(V)LMs hallucinate so as to produce more trustworthy and reliable systems. 
Before these large black-box systems are applied in safety-critical situations, there need to be methods to detect and mitigate hallucinations.
Thus, this survey collects and discusses current hallucination mitigation techniques for foundation models in decision-making tasks, and presents potential research directions.

Existing surveys particularly focus on presenting methods for hallucination detection and mitigation in question-answering (QA)~\cite{ji2023survey, rawte2023survey, zhang2023siren, ye2023cognitive} or object detection tasks~\cite{li2023evaluating}.
For example,~\citet{ji2023survey} summarize metrics and hallucination detection and mitigation methods for abstractive summarization, dialogue generation, generative question-answering, data-to-text generation, and machine translation for natural language generation models in general.
\citet{ye2023cognitive} and~\citet{rawte2023survey} take this work a step further by classifying hallucination mitigation strategies for foundation models in particular, in generative text and multi-modal settings respectively.
\citet{zhang2023siren} similarly tackle hallucination detection for language foundation models, but limit their definition of hallucinations to conflicts between generations and inputs, contexts, and facts.
The authors also discuss other possible problems in generated outputs, like ambiguity and bias. 
In the domain of image captioning,~\citet{li2023evaluating} provide examples of hallucinations and present a new metric to evaluate object hallucinations.
More recent surveys, like the ones from~\citet{bai2024hallucination} and~\citet{liu2024survey}, dive deeper into hallucination mitigation methods for multi-modal foundation models applied to image captioning tasks, but lack a general definition for hallucinations that encompasses decision-making applications.
The extensive survey from~\citet{huang2023survey} primarily focuses on methods for evaluating the trustworthiness of language models from the fronts of factuality and faithfulness.
Their taxonomy also briefly touches on hallucinations in object detection using LVLMs, but like other surveys, ignores broader decision-making deployments.
There are also other works that provide examples of current use cases of L(V)LMs in autonomous vehicles~\cite{yang2023llm4drive} and robotics~\cite{zeng2023large, zhang2023large}.
\citet{wang2023decoding} perform a deep analysis of the trustworthiness of a variety of foundation models and~\citet{chen2024can} provide a taxonomy of hallucinations within LLMs, but both exclude applications to general decision problems.
To the best of our knowledge, we are the first to propose a general definition of hallucinations that can be flexibly tuned to any particular deployment setting, including commonly found applications to QA or information retrieval, and more recent developments in planning or control.
Furthermore, there is no existing work that summarizes state of the art methods for hallucination detection and mitigation approaches within decision-making and planning tasks.
Additionally, to the best of our knowledge, ours is the first review to provide guidelines for choosing and designing hallucination intervention algorithms across different application areas.

In the remainder of this work, we 
discuss the current uses of foundation models for decision-making tasks in Section~\ref{sec:foundation_models}, 
define and provide examples of hallucinations in Section~\ref{sec:hallucinations}, 
identify current detection methods in Section~\ref{sec:detection}, 
present guidelines in Section~\ref{sec:guidelines},
and explore research directions in Section~\ref{sec:future}.

\section{Foundation Models Making Decisions}
\label{sec:foundation_models}

Originally coined by~\citet{bommasani2022opportunities}, the term \emph{foundation models} refers to models that are ``trained on broad data at scale such that they can be adapted to a wide range of downstream tasks.''
This approach is in contrast to works that design and train models on a smaller subset of data for the purpose of being deployed to a specific task~\cite{yang2024harnessing}.
The key difference is that foundation models undergo a pre-training procedure on a large-scale dataset containing information from a variety of possible deployment fields, through which they are expected to learn more general features and correspondences that may be useful at test-time on a broader set of tasks~\cite{zhou2023comprehensive, zhao2023survey}.
Examples of existing pre-trained foundation models span language~\cite{devlin2019bert, brown2020language, touvron2023llama}, vision~\cite{caron2021emerging, oquab2024dinov2, kirillov2023segment}, and multi-modal~\cite{radford2021learning, achiam2023gpt} inputs. 
In this section, we give a brief overview of existing use cases for foundation models in robotics and autonomous vehicles, and we provide a discussion of other decision-making systems in Appendix~\ref{sec:app_other_areas}.
We also succinctly point out hallucinations found in these works and leave a lengthier discussion in Section~\ref{sec:hal_examples}.
Readers should refer to works from~\citet{cui2024survey},~\citet{yang2023llm4drive},~\citet{zeng2023large}, and~\citet{zhang2023large} for a deeper review of application areas.

\subsection{Autonomous Driving}
\label{sec:found_ad}

\begin{figure*}[t]
  \centering
  \includegraphics[width=0.9\linewidth]{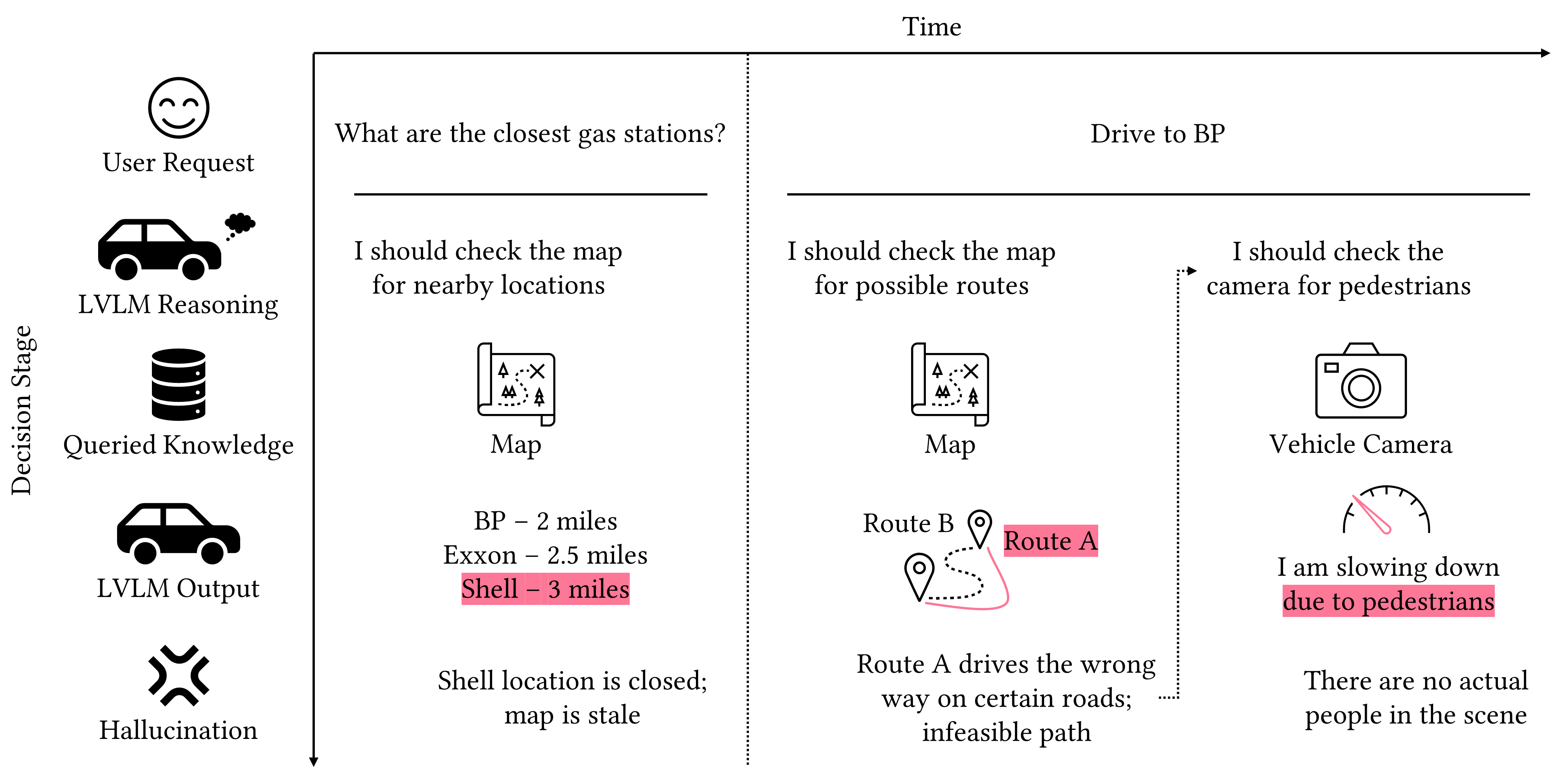}
  \caption{\textbf{Example deployment of an LVLM foundation model in an autonomous driving setting.} Hallucinations (pink) may arise at any point in the decision-making pipeline, including information retrieval, planning, perception, and control. In this example, the LVLM correctly queries the map for possible destinations of gas stations, but lists a location that is no longer open. Then, when navigating to one of the locations, the model predicts a path on the map that goes in the wrong direction of traffic flow. Finally, when applied to perception tasks for detecting possible pedestrians in front of the vehicle, the model hallucinates nearby people, causing an improper control action.}
  \label{fig:hall_example}
\end{figure*}

For the autonomous vehicle domain, researchers have formulated the use of language foundation models as a fine-tuning and prompt engineering problem~\cite{wen2023road, wen2024dilu}.
An external sub-system is usually designed with (1) a perception module to process signals from raw sensors, (2) a memory bank of prior important experiences and its corresponding similarity function to find alike scenarios, and (3) a prompt generator to convert current sensor data and relevant memories into natural language that can be input to the foundation model.
Currently, works either fine-tune LLMs with a few examples, or directly apply the model in a zero-shot manner, on a QA task with driving related questions.
By framing the task in a QA form, researchers have been able to provide context to the L(V)LM to probe for high-level natural language decisions~\cite{wen2023road, wen2024dilu}, path planning~\cite{mao2023language, sima2023drivelm}, vehicle tracking and trajectory prediction~\cite{wu2023language, keysan2023can}, descriptions of the surroundings of the vehicle~\cite{chen2023driving, xu2024drivegpt4}, and low-level control~\cite{liu2023mtd}.
Figure~\ref{fig:hall_example} is an example of how a foundation model may be used in an autonomous driving setting, with possible hallucinations in deployment.

\paragraph{High-level Decisions}
\citet{wen2024dilu} propose DiLu, a framework consisting of reasoning, reflection, and memory modules that support an LLM in producing high-level decisions for autonomous driving, and they test their method within a driving simulator environment.
Specifically, the reasoning module views the current observation of the vehicle, queries the memory module for any similar situations that were encountered in the past, and converts the experience into a prompt, which is input to the LLM.
The prompt is formatted such that it elicits chain-of-thought reasoning~\cite{wei2022chain} from the LLM, which is shown to improve the accuracy of the model.
The generated text output by the LLM is summarized by the reflection module, and is used to update the memory bank of experiences.
A separate decision decoder model converts the summary into a discrete high-level decision (\eg~idle, turn right, accelerate,~\etc) to take in the simulator.
The same authors have also experimented with prompting the LVLM model, GPT-4V~\cite{achiam2023gpt}, in a zero-shot manner to describe the surroundings of the vehicle, take high-level decisions, and explain why it believes it would be a good action to take~\cite{wen2023road}.
They find that GPT-4V is capable of identifying safety-critical scenarios and suggests driving more conservatively in those situations. 
However, like other existing vision models~\cite{suganuma2022current}, it has difficulty in detecting traffic light states.
We discuss other hallucination examples in traffic scenarios in Section~\ref{sec:hal_examples}.

\paragraph{Path Planning}
Agent-Driver from~\citet{mao2023language} utilizes a tool library of functions that communicate with neural modules that are responsible for object detection, trajectory prediction, occupancy estimation, and mapping.
The LLM is asked to reason about what information would be helpful to plan a path of the ego vehicle, and calls on functions from the tool library to build up relevant context.
Like~\citet{wen2024dilu}, the authors use a memory bank of prior driving experiences to bolster the context provided to the LLM.
With this context, the LLM predicts a suitable path for the ego vehicle to follow.
If a collision is detected between the predicted trajectory and surrounding objects in the scene, the LLM undergoes self-reflection, like Reflexion~\cite{shinn2023reflexion}, another hallucination mitigation technique, to fine-tune its prediction.
Through a short study, the authors test the frequency of invalid, hallucinated outputs from the model, and find that their self-reflection approach results in zero invalid generations at test-time.
\citet{sima2023drivelm} build up context before predicting a path for the ego vehicle by asking a VLM questions about its perception of surrounding vehicles, predicting their behaviors, planning high-level vehicle decisions, converting to lower level discrete actions, and finally, estimating a coordinate-level trajectory.
Their method, DriveLM-Agent, predicts paths from images in an end-to-end manner using a multi-modal approach, whereas Agent-Driver requires sensor modules to process context separately.

\paragraph{Trajectory Prediction}
\citet{wu2023language} propose PromptTrack as a method to predict bounding boxes and trajectories of vehicles in multi-view camera scenes conditioned on a text prompt.
PromptTrack is another end-to-end method that encodes multi-view images with an image encoder, decodes previously tracked boxes and current detections into new tracks, uses a language embedding branch to predict 3D bounding boxes, and updates the memory of current tracks using past \& future reasoning branches.
Rather than using ego vehicle point of view images for object tracking,~\citet{keysan2023can} propose an approach to convert rasterized images from a birds-eye view of the scene into a prompt describing the past trajectories of vehicles with Bézier curves. 
This method combines vision and language encoders to generate the Bézier curve-based scene description, and elicits a language model to predict trajectories in a similar format.

\paragraph{Scene Understanding}
Works using foundation models for generating scene descriptions given multi-modal information frame the task as a QA problem.
For example,~\citet{chen2023driving} use a reinforcement learning (RL) agent pre-trained on the driving task in simulation to collect a dataset containing the vehicle's state, environment observation (assumed to be ground truth from simulator,) low-level action, and the ego's percentage of attention placed on different surrounding agents (if using an attention-based network).
The authors ask ChatGPT~\cite{openai2022chatgpt} to act as a professional driving instructor who generates questions and answers given general driving rules and the vectorized description of the scene from the RL agent.
Then, combining a pre-trained LLM with a vector embedder and former model, the architecture is trained end-to-end to answer questions from the scene.
Examples of questions the architecture is posed at test-time include, ``What objects are you observing,'' ``How are you going to drive in this situation and why,'' and, ``What are the best tourist spots in London'' (the last of which is considered out of scope by the driving model).
The authors acknowledge that the LLM may generate undesired hallucinated outputs during deployment, so they augment their training instruction dataset with out of scope questions that the model should learn to refuse to answer.
DriveGPT4, proposed by \citet{xu2024drivegpt4}, is a multi-modal LLM using LLaMA 2~\cite{touvron2023llama2} which encodes video sequences with an encoder model, and projects question and video embeddings into a text form to be input to the LLM.
A decoder model converts the tokenized output of the LLM into a low-level action with a corresponding high-level reason to take the action. 
Like~\citeauthor{chen2023driving}, the authors collect a driving QA dataset to align and fine-tune the architecture to driving tasks. 

\paragraph{Control}
\citet{liu2023mtd} tackle low-level control within unsignalized intersections by formulating the problem as a multi-task decision-making process.
Specifically, they train expert policies with RL to perform individual control tasks (\eg~turning left, going forward, and turning right), with which they collect an expert demonstration dataset.
An LLM model, GPT-2~\cite{radford2019language}, is fine-tuned to predict next actions given past trajectories of state and action pairs from the expert dataset. 
The authors showcase that their method, MTD-GPT, is able to achieve higher success rates across each of the control tasks over the original RL policies alone, showcasing the promise of foundation models to leverage their general knowledge in a multi-task problem setting.

\subsection{Robotics}

Foundation models have also been used in the robotics domain for object detection, affordance prediction, grounding, navigation, and communication.
An example of a robot deployed with LVLM capabilities and potential hallucinations is shown in Figure~\ref{fig:app_hall_robot_example} of Appendix~\ref{sec:app_robotics}.
\citet{ichter2023do} are motivated by the issue of misalignment between the capabilities of a robot and what an LLM believes it is capable of performing.
Because LLMs may not specifically be trained with data from the robot it is to be deployed on, there is a gap in the model's understanding and the true capacity of the robot, which could lead to hallucinated generations that cannot feasibly be used at runtime.
The authors propose SayCan as a method to combine the general knowledge of LLMs with the specific capabilities of a robot in the real-world.
Specifically, an LLM is given a task in text form, and is asked to output a list of smaller actions to take in order to complete said task successfully.
To constrain the LLM to generate possible actions available to the robot, they assume access to (1) the probability distribution of next tokens to generate from the model, and (2) a set of available skills on the robot, with which they compute the probability of the LLM generating each of the skills next.
SayCan greedily selects the action that has the highest product of the next token probability from the LLM and the probability of the action actually successfully being executed in the environment, until the model predicts it has completed the task.

Rather than relying purely on textual context, PaLM-E, proposed by~\citet{driess2023palme}, is a multi-modal model that converts various sensor inputs (\eg~images) to a token-space embedding that is combined with instruction embeddings to be input to a PaLM LLM~\cite{chowdhery2023palm}.
PaLM is used to either answer questions about the surroundings of the robot, or to plan a sequence of actions to perform to complete a task. 
\citeauthor{driess2023palme} further acknowledge that the multi-modality of their PaLM-E architecture leads to increased risk of hallucinations.

Inspired by recent promising findings in using foundation models to generate programs~\cite{chen2021evaluating}, other works deploy foundation models to write low-level code to be run on robots.
\citet{liang2023code} present Code as Policies, which uses LLMs to hierarchically generate interactive code and functions that can be called.
As the model writes main code to be run on a robot given an instructive prompt of the task from the user, it identifies functions to call within the higher level code to complete the task successfully.
The authors show that LLMs can leverage third party libraries for existing functions, or develop their own library of functions dynamically with custom methods for the task.
While the functionality of Code as Policies can be tested easily for low-level skill definitions, longer multi-step problems require testing whether all requested conditions have been met by running the generated code on the robot.
As such,~\citet{hu2024deploying} propose the RoboEval performance benchmark for testing robot-agnostic LLM-generated code.
Specifically, the CodeBotler platform provides an LLM access to abstract functions like ``pick,'' ``place,'' and ``get\_current\_location'' that have the same external interface regardless of the robot to be deployed on.
Like Code as Policies, CodeBotler is provided a text instruction from the user and generates code to be tested. 
Then the RoboEval benchmark uses \emph{RoboEval Temporal Logic} (RTL) to test whether the generated code meets task and temporal ordering constraints provided in the original prompt.
Furthermore, they test the robustness of the LLM by passing in several paraphrased prompts to check for consistency across inputs. 
We discuss similar consistency-checking strategies for identifying hallucinations in decision-making tasks further in Section~\ref{sec:det_analyze}.

In the space of robot navigation, LM-Nav leverages a VLM and attempts to predict a sequence of waypoints for a robot to follow and visit landmarks described within a language command~\cite{shah2023lm}.
Here, the authors use in-context learning~\cite{dong2023survey} to teach GPT-3~\cite{brown2020language} to extract desired landmarks from a natural language instruction.
Assuming there are images of the possible landmarks the robot can navigate to in its environment, LM-Nav uses CLIP~\cite{radford2021learning} to predict the closest matching pairs of extracted landmark descriptions and waypoint images.
Finally, dynamic programming is applied on the complete graph of the environment to optimize the path of landmarks to visit.
The overall predicted path is optimized to maximize the likelihood of successfully completing the instruction input to the model.

\section{Hallucinations}
\label{sec:hallucinations}

Even with all their success on a multitude of deployment areas, foundation models still produce inconsistent outputs, or \emph{hallucinate}, at test-time.
Here, we provide a general definition for hallucinations that can be applied to any foundation model deployment task, including various autonomous systems.
Additionally, we give examples of hallucinations encountered in literature, and discuss how they come about during testing.

\subsection{What are hallucinations?}
\label{sec:what_are_hallu}

Across current literature on foundation models, there exist similar patterns and themes that can be used to develop a unified definition for hallucinations.
With the majority of works studying this problem within QA tasks, where ground truth answers are available, several authors explain hallucinations as producing text that includes details/facts/claims that are fictional/misleading/fabricated rather than truthful or reliable~\cite{rawte2023survey}.
Works making use of a dedicated knowledge-base further describe hallucinations as generating nonsensical or false claims that are unsubstantiated or incorrectly cited~\cite{mundler2024self, chen2023purr, zhang2023language, li2023halueval}.
\citet{varshney2023stitch} also present the idea that foundation models may sound syntactically correct, or \emph{coherent}, while simultaneously being incorrect.
\citet{gallifant2024peer}, who perform a peer review of the GPT-4 technical paper, state that hallucinations include responses that are irrelevant to the original prompt.
\citet{li2023evaluating}, who specifically explore hallucinations of LVLMs in detecting and classifying objects within images, define hallucinations as generating object descriptions inconsistent with target images. 
A common theme among existing hallucination definitions for QA, information retrieval, and image captioning domains is that, while the generation may sound coherent, either the output is incorrect, or the model's reasoning behind the generated text is incorrect.
However, we find these characteristics on their own do not completely encompass the hallucinations found in decision-making tasks in literature, thus requiring additional nuances.

Within papers that apply foundation models to decision-making tasks specifically, researchers have encountered similar problems of hallucinations impacting performance.
\citet{park2024clara} describe hallucinations as predicting an incorrect feasibility of an autonomous system when generating an explanation behind the uncertainty of an action to take.
Similarly,~\citet{kwon2023reward} find that language models may provide incoherent reasoning behind their actions.
\citet{wang2024conformal} and~\citet{ren2023robots} believe that these generative models also have a sense of high (false) confidence when generating incorrect or unreasonable plans.
In the case of robot navigation and object manipulation,~\citet{hu2024deploying} and~\citet{liang2024introspective} refer to hallucinations as attempting to interact with non-existent locations or objects.

In the code generation task,~\citet{chen2021evaluating} use the term ``alignment failure,'' with similar effects to those of hallucinations discussed above.
More specifically, the authors informally describe an alignment failure as an outcome where a model is \emph{capable} of performing a task, but \emph{chooses} not to.
If a model is able to complete a task successfully within its latent space (perhaps through additional prompt engineering or fine-tuning), one may ask, ``Why would the model \emph{choose} not to?''
As foundation models are trained with the next-token reconstruction objective on a training set, they attempt to maximize the likelihood of the next token appearing at test-time as well.
Consequently, if the test-time prompt includes even minor mistakes,~\citeauthor{chen2021evaluating} find that LLMs will continue to generate buggy code to match the input prompt.
This issue is further described in Section~\ref{sec:hal_why}.

We realize existing definitions for hallucinations are extremely disparate depending on the deployment area.
Nevertheless, we have identified three distinct features that are commonly missing within hallucinated generations:~\emph{compliance},~\emph{desirability}, and~\emph{relevancy}.
Additionally, we find hallucinations may seem to be~\emph{plausible} while they are in fact unacceptable.
A compliance metric checks that the generation meets hard constraints while the desirability metric measures how well the generation meets soft constraints defined by the model engineer.
Irrelevant generations refer to predictions that contain details outside of a set of requested topics.
Notice that irrelevant predictions can still be considered compliant and desirable depending on how the hard and soft constraints are defined.
Plausibility compares the generation's syntax to those of a set of known, unhallucinated samples.
While relevancy is evaluating the content of a prediction, plausibility is assessing its phrasing.
Table~\ref{tab:sub_defs} provides complete definitions for each of the four characteristics.

\begin{table}[t]
  \begin{center}
    \caption{\textbf{Definitions for compliance, desirability, relevancy, and plausibility --- the four characteristics of hallucinations.}}
    \label{tab:sub_defs}
    \resizebox{\textwidth}{!}{
    \begin{tabular}{ll} 
    \toprule
    \textbf{Characteristic}  & \textbf{Definition} \\
      \midrule
      Compliance    & Generation meets hard constraints that cannot be ignored\\
      Desirability  & Generation attempts to meet soft constraints measured by some cost or reward function\\
      Relevancy     & Contents of generation do not fall outside of a defined set of topics\\
      Plausibility  & The syntactic similarity of a generation and in-domain normal samples measured via a critic function\\
      \bottomrule
    \end{tabular}
    }
  \end{center}
\end{table}

Then, to bridge definitions from existing QA application areas, decision-making tasks, and all other possible test scenarios for foundation models, we combine these findings and define the term hallucination as follows:

\begin{definition}
\label{def:hal}
\emph{A hallucination is a generated output from a model that conflicts with constraints or deviates from desired behavior in actual deployment, or is completely irrelevant to the task at hand, but could be deemed syntactically plausible under the circumstances.}
\end{definition}

There are three key pieces to this definition:

\begin{enumerate}[noitemsep]
  \item A generated output from a model.
  \item A deployment scenario to evaluate model outputs with any of the following:
  \begin{itemize}
    \item A list of constraints that must be \emph{compliant} within the generation.
    \item A loose interpretation of a \emph{desired behavior} the generation should meet.
    \item A set of topics \emph{relevant} to the task.
  \end{itemize} 
  \item Metrics measuring compliance, desirability, relevancy, and syntactic soundness (\emph{plausibility}) of generations.
\end{enumerate}

In practice, this definition generally encapsulates the qualities of hallucinations discussed earlier. 
For example, in QA or image captioning tasks, one may define a set of relevant topics that a generation should not stray from, and constraints may be held in the form a knowledge-base of ground truth facts.
The desired behavior of the generation may be to be phrased in an informative manner, rather than sarcastic.
On the other hand, in robot manipulation settings, a developer may have a set of constrained actions feasible on the robot, and the desired behavior could be to complete a task with as few actions as possible.
Relevancy may be measured in relation to the specific task to be deployed on (\eg~a prompt requesting a recipe to make pasta would find it irrelevant if the model also suggested a song to play while cooking).
Finally, plausibility informally relates to a measure of how believable an output is to a critic.
A more realistic generation has a greater chance of deceiving the user into trusting the model, even when the plan may be hallucinated.
Overall, hallucinated outputs may contain one or more of the core characteristics (noncompliant, undesired, irrelevant, and plausible) simultaneously, and our definition can be flexibly applied to any deployment scenario in mind by choosing metrics for each characteristic, respectively.
We show more examples of applying our definition to various tasks in Table~\ref{tab:examples}.

\subsection{Examples}
\label{sec:hal_examples}

\begin{table*}[t]
\centering
\caption{
\textbf{Examples of applying Definition~\ref{def:hal} to different tasks.}
Note that developers may choose to only define a subset of hallucination characteristics for their deployment depending on evaluation preferences.
The table is split into non-decision-making and decision-making applications.
}
\label{tab:examples}
\resizebox{\textwidth}{!}{
\begin{tabular}{m{1.2in}m{1.7in}m{1.7in}m{1.7in}m{1.7in}m{0cm}}
\toprule
\centering \multirow{2}{*}{\textbf{Problem Setting}} 
                             & \multicolumn{4}{c}{\textbf{Characteristic}}                                                               & \\
                             \cmidrule(lr){2-5}
                             & \textbf{Compliance}      
                                                  & \textbf{Desired Behavior}      
                                                                       & \textbf{Relevancy}       
                                                                                            & \textbf{Plausibility}   & \\
\midrule
Question-Answering           & \makecell[l]{Generations must align with\\database facts}
                                                  & \makecell[l]{Tone of answer should be\\informative}                  
                                                                       & \makecell[l]{Answers should not include\\references to unrelated topics}        
                                                                                            & \multirow{4}{*}{\makecell[l]{Generation is syntactically\\sound and believable}}
                                                                                                                        & \\
\vspace{4pt}Image Captioning & \vspace{4pt}\makecell[l]{Objects in description must\\appear in image}                   
                                                  & \vspace{4pt}\makecell[l]{Censor descriptions for\\inappropriate images}                 
                                                                       & \vspace{4pt}\makecell[l]{Descriptions should not be\\embellished with details that\\cannot be confirmed}
                                                                                            &
                                                                                                                        & \\
\midrule
Planning                     & \makecell[l]{Predicted sub-task must be\\feasible to solve}                   
                                                  & \makecell[l]{Plans should maximize\\expected return}                  
                                                                       & \multirow{4}{*}{\makecell[l]{Predicted sub-tasks and\\actions should not stray from\\the end goal with added steps}}
                                                                                            & \multirow{4}{*}{\makecell[l]{Generated plan is reasonable\\and seems to attempt to\\accomplish goal}}
                                                                                                                        & \\
\vspace{4pt}Control          & \vspace{4pt}\makecell[l]{Predicted action must be\\possible to perform}                   
                                                  & \vspace{4pt}\makecell[l]{Predict actions to complete\\plan efficiently}
                                                                       &          
                                                                                            &     
                                                                                                                        & \\
\bottomrule
\end{tabular}
}
\end{table*}

\paragraph{Driving Tasks}
As discussed in Section~\ref{sec:found_ad},~\citet{wen2023road} test GPT-4V on the autonomous driving task and identify failure modes.
Regardless of the weather and driving conditions, GPT-4V has difficulty detecting and identifying the traffic light state at an intersection, until the image has zoomed in on the light itself.
It also presents additional irrelevant (or completely false) details about other agents, when the prompt had no mention of them in the first place.
Furthermore, the model also has difficulty in describing temporal sequences (\ie~videos) and categorizing images by their direction within a panoramic view from the vehicle's perspective.
In their later work,~\citet{wen2024dilu} describe that hallucinations arise in these complex environments because of the high variability in driving scenarios.
Even after applying hallucination mitigation techniques like chain-of-thought reasoning, the model is not free of these undesired outputs.
A similar work evaluating the frequency at which LVLMs hallucinate in their descriptions of images, finds that these models' outputs may include non-existent objects, or additional irrelevant phrases (that may not even be possible to test for accuracy)~\cite{li2023evaluating}.
For example, in a picture of food on a table, an LVLM hallucinates a non-existent beverage, and predicts that the ``table is neatly arranged, showcasing the different food items in an appetizing manner.''
Although the classification error and irrelevant generation in this example are not critical, earlier works warn of possible failures with more severe, high societal impact (\eg~biases in models leading to marginalizing users)~\cite{bommasani2022opportunities}.

\paragraph{Code Generation}
\citet{chen2021evaluating} explore alignment failures of LLMs applied to code completion tasks.
The authors evaluate the likelihood of these models generating defective code given different input prompts, and discover that in-context learning using examples with buggy code has a higher chance of resulting in poor generations from the model on the actual task at hand.
The study also identifies similar model biases towards race, gender, religion, and other representations.
Furthermore, the authors find that their model, Codex, is able to generate code that could assist with developing insecure applications or malware, albeit in a limited manner.
These findings have been corroborated by other foundation model code generation works in the robotics domain.
For example,~\citet{wang2023voyager} describe that Voyager sometimes generates code with references to items that do not exist within MineDojo.
Similarly,~\citet{hu2024deploying} find that their model has the tendency to call functions with invalid objects or locations, pickup objects when it is already holding something, ask for help when no one is near, and other undesired behaviors.

\paragraph{Question-answering Domain}
Several works focus on identifying cases of hallucinations in QA tasks.
Although this application area is not the direct focus of this work, we present examples of hallucinations in this field as we can glean similar failure modes that could arise within decision-making systems.
Common hallucinations in QA result in incorrect answers to questions.
For example,~\citet{achiam2023gpt} find that GPT-4 ``hallucinates facts and makes reasoning errors.''
\citeauthor{achiam2023gpt} categorize these failures into closed-domain (given context, the model generates irrelevant information that was not in the context) and open-domain (the model outputs incorrect claims without any context) hallucinations.
After fine-tuning on more data with a hallucination mitigation objective, the model reduces its tendency to hallucinate, but still does not achieve perfect accuracy --- a similar trend encountered by~\citet{touvron2023llama}.
Another set of works identify hallucinations with contradictions among several sampled generations from an LLM, discussed further in Section~\ref{sec:det_analyze}~\cite{mundler2024self, zhang2023language}.
Intuitively, if a context passed into a model results in conflicting generations, the model must be hallucinating some part of the output.
Notice in this example, with relation to Definition~\ref{def:hal}, self-contradiction works test for compliance by checking \emph{consistency} among multiple (hallucinated) generations, rather than with respect to a ground-truth knowledge-base that usually exists in QA tasks.
As such, our definition can flexibly apply to different system setups by describing compliance, desired behavior, and relevancy respectively.

Additional examples of hallucinations encountered in image, video, and $3$D generation methods, and broader impacts faced by medical, legal, and finance industries are discussed in Appendix~\ref{sec:app_hal_examples} and~\ref{sec:app_broader_impacts}.

\subsection{Why do they happen?}
\label{sec:hal_why}

There are several speculations as to how hallucinations come about during deployment.
First and foremost, like any learning task, foundation models are sensitive to biases in training data~\cite{rawte2023survey}.
Once a model is trained on a given large dataset, some facts may become out-of-date or stale at any point in time~\cite{puthumanaillam2024moral}.
Furthermore, as the training set is embedded into a smaller encoding dimension, the knowledge within an L(V)LM's frozen parameters is lossy, and models cannot feasibly be fine-tuned every time there is new data~\cite{peng2023check, elaraby2023halo}.
\citet{zhang2023language} recommend changing algorithm parameters at runtime, such as, \emph{temperature} (spread of probability distribution of next token), \emph{top-$K$ sampling} (narrows the set of next tokens to be considered), and \emph{beam search} (choosing a set of possible beams, \ie~trajectories, of next tokens based on high conditional probabilities), but the process of tuning these parameters is expensive.

To combat out-of-date training data, some works provide models with an external knowledge-base of information to pull facts from, with the hope of increasing model accuracy.
Even with this up-to-date information,~\citet{zhang2023knowledge} pose that there may exist a misalignment between the true capabilities of a model, and what a user believes the model is capable of, leading to poor prompt engineering.
In fact, poor prompting is one of the most significant causes of hallucinations.
\citet{chen2021evaluating} find that poor quality prompts lead to poor quality generations, in the context of code completion.
This phenomenon is attributed to the reconstruction training objective of LLMs attempting to maximize the likelihood of next generated tokens, given context and past outputs~\cite{mialon2023augmented},~\ie
\[
\log p\left(s|x\right) = \sum_{i=1}^{N} \log p\left(\sigma_i|x,\sigma_{i-k}\ldots \sigma_{i-1}\right)
\]
where $x$ is a context input to the model, $s$ is an output sequence of $N$ tokens $\sigma_1\ldots\sigma_N$, and any generated token $\sigma_i$ is conditioned on $k$ previously generated tokens.
As the public datasets these models are trained on contain some fraction of undesirable generations (\eg~defective code), the models become biased to generate similar results under those inputs.
\citet{qiu2023latent} show that this limitation can actually be exploited to push foundation models to generate toxic sentences, or completely lie, by simply rewording the prompt.

While foundation models condition generated tokens on ground-truth text without hallucinations at train time, during inference, the model chooses future tokens conditioned on previously (possibly hallucinated) generated text.
As such,~\citet{chen2023hallucination} and~\citet{varshney2023stitch} state that generated outputs are more likely to contain hallucinations if prior tokens are hallucinated as well.
Furthermore,~\citet{li2023large} find that, even if prompt context provided to a foundation model is relevant, the model may choose to ignore the information and revert to its own (possibly outdated or biased) parameterized knowledge.

Overall, the hallucination detection task is highly complex with several possible sources of failures that need to be considered at test-time.
\citet{chen2024can} validate the complexity of the detection problem with studies identifying that human- and machine-based detectors have higher difficulty correctly classifying misinformation generated from LLMs than those written by other people.

\section{Detection and Mitigation Strategies}
\label{sec:detection}

Hallucination detection and mitigation methods can be classified into three types (white-, grey-, and black-box) depending on the available inputs to the algorithm.
Generally, given some context, a foundation model outputs a predicted sequence of tokens, the corresponding probabilities of each token, and embeddings of the generation from intermediate layers in the network.
White-box hallucination detection methods assume access to all three output types, grey-box require token probabilities, and black-box only need the predicted sequence of tokens.
Because not all foundation models provide access to their hidden states, or even the output probability distribution of tokens (\eg~the ChatGPT web interface), black-box algorithms are more flexible during testing.
In this section, we present existing detection and mitigation approaches clustered by input type. 
While several of these works show promise in QA and image captioning settings, many of them require further validation on decision-making tasks, and we will point out these methods as they come about.
Works in this section are summarized in Table~\ref{tab:summary}.
We also provide additional details about the evaluation setups of works deployed on custom datasets, simulators, or the real-world.
Certain methods that are less related to decision-making are described in Appendix~\ref{sec:app_detection}.
Additionally, frequently used metrics, datasets, and simulators are summarized in Appendix~\ref{sec:app_evaluation}.

\afterpage{
{\fontsize{8}{12}\selectfont
\begin{center}
\setlength\LTcapwidth{\textwidth}
\setlength\tabcolsep{0pt}
\begin{longtable}{m{1.7cm}m{0.8in}m{0.7in}m{0.7in}m{0.7in}m{0.8in}m{1.1in}m{0cm}}
\caption{
\textbf{A summary of hallucination detection \& mitigation methods discussed in Section~\ref{sec:detection}.}
Deployment scenarios are split into question-answering (QA), information retrieval (IR), image captioning (IC), image generation (IG), \& planning (P) tasks.
The method ID includes the subsection the method appears in the paper and the order in which it appears in the subsection.
Bolded method IDs are deployed to decision-making tasks specifically.
Custom datasets, custom simulators, \& real-world experiments for testing are abbreviated as CD, CS, \& RW, respectively.
}
\label{tab:summary}
\\
\toprule
\multicolumn{1}{c}{\multirow{2}{*}{\textbf{Modality}}} & \multicolumn{1}{c}{\multirow{2}{*}{\textbf{Method Type}}} & \multicolumn{1}{c}{\multirow{2}{*}{\textbf{Method ID}}} & \multicolumn{2}{c}{\textbf{Application}}                                         & \multicolumn{1}{c}{\multirow{2}{*}{\textbf{Deployment}}} & \multicolumn{1}{c}{\multirow{2}{*}{\textbf{\makecell[c]{Evaluation Setting}}}}  \\ 
                                                                                                                                                                             \cmidrule(lr){4-5}
                                                       &                                                           &                                                         & \multicolumn{1}{c}{\textbf{Detection}} & \multicolumn{1}{c}{\textbf{Mitigation}} &                                                          &                                                                   \\
\midrule
\endfirsthead
\multicolumn{7}{c}{{\bfseries \tablename\ \thetable{} -- continued from previous page}}                                                                                                                                                                                                                                                                                                        \\
\toprule
\multicolumn{1}{c}{\multirow{2}{*}{\textbf{Modality}}} & \multicolumn{1}{c}{\multirow{2}{*}{\textbf{Method Type}}} & \multicolumn{1}{c}{\multirow{2}{*}{\textbf{Method ID}}} & \multicolumn{2}{c}{\textbf{Application}}                                         & \multicolumn{1}{c}{\multirow{2}{*}{\textbf{Deployment}}} & \multicolumn{1}{c}{\multirow{2}{*}{\textbf{\makecell[c]{Evaluation Setting}}}}  \\ 
                                                                                                                                                                             \cmidrule(lr){4-5}
                                                       &                                                           &                                                         & \multicolumn{1}{c}{\textbf{Detection}} & \multicolumn{1}{c}{\textbf{Mitigation}} &                                                          &                                                                   \\
\midrule
\endhead
\multicolumn{7}{l}{{\textbf{Continued on next page}}}                                                                                                                                                                                                                                                                                                                                          \\
\endfoot
\endlastfoot
\multicolumn{1}{c}{\multirow{9}{*}{\STAB{\rotatebox[origin=c]{90}{White-box}}}}           
                          & \multicolumn{1}{c}{\multirow{4}{*}{\makecell[c]{Hidden\\States}}}                      & \centering \hyperref[itm:saplma]{4.1.1.1}                                & \centering $\bullet$                  &                                       & \centering QA                              & \multicolumn{1}{c}{CD}                                            \\
                          &                                                                                        & \centering \hyperref[itm:yao_adv]{4.1.1.2}                               & \centering $\bullet$                  &                                       & \centering QA                              & \multicolumn{1}{c}{CD}                                            \\
                                                                                                                                                                                                                                                                                                                           \cmidrule(lr){7-7}
                          &                                                                                        & \centering \multirow{2}{*}{\hyperref[itm:luna]{4.1.1.3}}                 & \centering \multirow{2}{*}{$\bullet$} &                                       & \centering \multirow{2}{*}{QA}             & \multicolumn{1}{c}{DecodingTrust~\cite{wang2023decoding}}         \\
                          &                                                                                        &                                                                          &                                       &                                       &                                            & \multicolumn{1}{c}{TruthfulQA~\cite{lin2022truthful}}             \\
                                                                                                                   \cmidrule(lr){3-7}                                                                                                                                                    
                          & \multicolumn{1}{c}{\multirow{2}{*}{\makecell[c]{Attention\\Weights}}}                  & \centering \multirow{2}{*}{\hyperref[itm:opera]{4.1.2.1}}                & \centering \multirow{2}{*}{$\bullet$} & \centering \multirow{2}{*}{$\bullet$} & \centering \multirow{2}{*}{IC}             & \multicolumn{1}{c}{MSCOCO~\cite{lin2014mscoco}}                   \\ 
                          &                                                                                        &                                                                          &                                       &                                       &                                            & \multicolumn{1}{c}{Visual Genome~\cite{krishna2017visual}}        \\
                                                                                                                   \cmidrule(lr){3-7}                                                                                                                                                    
                          & \multicolumn{1}{c}{\multirow{2}{*}{\makecell[c]{Honesty\\Alignment}}}                  & \centering \hyperref[itm:lin_hon]{4.1.3.1}                               & \centering $\bullet$                  &                                       & \centering QA                              & \multicolumn{1}{c}{CD}                                            \\
                          &                                                                                        & \centering \hyperref[itm:yang_hon]{4.1.3.2}                              & \centering $\bullet$                  & \centering $\bullet$                  & \centering QA                              & \multicolumn{1}{c}{TriviaQA~\cite{joshi2017triviaqa}}             \\
\midrule         
\multicolumn{1}{c}{\multirow{15}{*}{\STAB{\rotatebox[origin=c]{90}{Grey-box}}}}            
                          & \multicolumn{1}{c}{\multirow{6}{*}{\makecell[c]{Concept\\Probabilities}}}              & \centering \multirow{2}{*}{\hyperref[itm:varsh_prob]{4.2.1.1}}           & \centering \multirow{2}{*}{$\bullet$} & \centering \multirow{2}{*}{$\bullet$} & \centering \multirow{2}{*}{IR/QA}          & \multicolumn{1}{c}{HotpotQA~\cite{yang2018hotpotqa}}              \\ 
                          &                                                                                        &                                                                          &                                       &                                       &                                            & \multicolumn{1}{c}{CD}                                            \\
                                                                                                                                                                                                                                                                                                                           \cmidrule(lr){7-7}
                          &                                                                                        & \centering \hyperref[itm:lure]{4.2.1.2}                                  &                                       & \centering $\bullet$                  & \centering IC                              & \multicolumn{1}{c}{MSCOCO~\cite{lin2014mscoco}}                   \\
                                                                                                                                                                                                                                                                                                                           \cmidrule(lr){7-7}
                          &                                                                                        & \centering \multirow{3}{*}{\textbf{\hyperref[itm:saycanpay]{4.2.1.3}}}   &                                       & \centering \multirow{3}{*}{$\bullet$} & \centering \multirow{3}{*}{P}              & \multicolumn{1}{c}{Ravens~\cite{zeng2021transporter}}             \\ 
                          &                                                                                        &                                                                          &                                       &                                       &                                            & \multicolumn{1}{c}{BabyAI~\cite{chevalier2018babyai}}             \\
                          &                                                                                        &                                                                          &                                       &                                       &                                            & \multicolumn{1}{c}{VirtualHome~\cite{puig2018virtual}}            \\
                                                                                                                   \cmidrule(lr){3-7}         
                          & \multicolumn{1}{c}{\multirow{9}{*}{\makecell[c]{Conformal\\Prediction}}}               & \centering \multirow{3}{*}{\hyperref[itm:quach]{4.2.2.1}}                & \centering \multirow{3}{*}{$\bullet$} & \centering \multirow{3}{*}{$\bullet$} & \centering \multirow{3}{*}{IR/QA}          & \multicolumn{1}{c}{MIMIC-CXR~\cite{johnson2019mimic}}             \\ 
                          &                                                                                        &                                                                          &                                       &                                       &                                            & \multicolumn{1}{c}{CNN/DM~\cite{hermann2015teaching}}             \\ 
                          &                                                                                        &                                                                          &                                       &                                       &                                            & \multicolumn{1}{c}{TriviaQA~\cite{joshi2017triviaqa}}             \\
                                                                                                                                                                                                                                                                                                                           \cmidrule(lr){7-7}
                          &                                                                                        & \centering \hyperref[itm:kumar]{4.2.2.2}                                 & \centering $\bullet$                  & \centering $\bullet$                  & \centering QA                              & \multicolumn{1}{c}{MMLU~\cite{hendrycks2021measuring}}            \\
                                                                                                                                                                                                                                                                                                                           \cmidrule(lr){7-7}
                          &                                                                                        & \centering \multirow{2}{*}{\textbf{\hyperref[itm:knowno]{4.2.2.3}}}      & \centering \multirow{2}{*}{$\bullet$} & \centering \multirow{2}{*}{$\bullet$} & \centering \multirow{2}{*}{P}              & \multicolumn{1}{c}{TableSim~\cite{ren2023robots}}                 \\ 
                          &                                                                                        &                                                                          &                                       &                                       &                                            & \multicolumn{1}{c}{RW}                                            \\
                                                                                                                                                                                                                                                                                                                           \cmidrule(lr){7-7}
                          &                                                                                        & \centering \multirow{2}{*}{\textbf{\hyperref[itm:liang_intro]{4.2.2.4}}} & \centering \multirow{2}{*}{$\bullet$} & \centering \multirow{2}{*}{$\bullet$} & \centering \multirow{2}{*}{P}              & \multicolumn{1}{c}{TableSim~\cite{ren2023robots}}                 \\ 
                          &                                                                                        &                                                                          &                                       &                                       &                                            & \multicolumn{1}{c}{CD}                                            \\
                                                                                                                                                                                                                                                                                                                           \cmidrule(lr){7-7}
                          &                                                                                        & \centering \textbf{\hyperref[itm:heracles]{4.2.2.5}}                     & \centering $\bullet$                  & \centering $\bullet$                  & \centering P                               & \multicolumn{1}{c}{CS}                                            \\
\midrule         
\multicolumn{1}{c}{\multirow{9}{*}{\STAB{\rotatebox[origin=c]{90}{Black-box}}}}          
                          & \multicolumn{1}{c}{\multirow{9}{*}{\makecell[c]{Analyzing\\Samples}}}                  & \centering \hyperref[itm:selfcheck]{4.3.1.1}                             & \centering $\bullet$                  &                                       & \centering IR/QA                           & \multicolumn{1}{c}{CD}                                            \\
                          &                                                                                        & \centering \hyperref[itm:halo]{4.3.1.2}                                  & \centering $\bullet$                  & \centering $\bullet$                  & \centering IR/QA                           & \multicolumn{1}{c}{CD}                                            \\
                                                                                                                                                                                                                                                                                                                           \cmidrule(lr){7-7}
                          &                                                                                        & \centering \multirow{3}{*}{\hyperref[itm:debate]{4.3.1.3}}               & \centering \multirow{3}{*}{$\bullet$} & \centering \multirow{3}{*}{$\bullet$} & \centering \multirow{3}{*}{IR/QA}          & \multicolumn{1}{c}{GSM$8$K~\cite{cobbe2021training}}              \\ 
                          &                                                                                        &                                                                          &                                       &                                       &                                            & \multicolumn{1}{c}{MMLU~\cite{hendrycks2021measuring}}            \\
                          &                                                                                        &                                                                          &                                       &                                       &                                            & \multicolumn{1}{c}{CD}                                            \\
                                                                                                                                                                                                                                                                                                                           \cmidrule(lr){7-7}
                          &                                                                                        & \centering \multirow{3}{*}{\textbf{\hyperref[itm:clara]{4.3.1.4}}}       & \centering \multirow{3}{*}{$\bullet$} & \centering \multirow{3}{*}{$\bullet$} & \centering \multirow{3}{*}{P}              & \multicolumn{1}{c}{SaGC~\cite{park2024clara}}                     \\ 
                          &                                                                                        &                                                                          &                                       &                                       &                                            & \multicolumn{1}{c}{CS}                                            \\
                          &                                                                                        &                                                                          &                                       &                                       &                                            & \multicolumn{1}{c}{RW}                                            \\
                                                                                                                                                                                                                                                                                                                           \cmidrule(lr){7-7}
                          &                                                                                        & \centering \hyperref[itm:mundler]{4.3.1.5}                               & \centering $\bullet$                  & \centering $\bullet$                  & \centering IR/QA                           & \multicolumn{1}{c}{CD}                                            \\
                                                                                                                                                                                                                                                                                                                           \cmidrule(lr){7-7}
\multicolumn{1}{c}{\multirow{38}{*}{\STAB{\rotatebox[origin=c]{90}{Black-box}}}}          
                          & \multicolumn{1}{c}{\multirow{11}{*}{\makecell[c]{Analyzing\\Samples}}}                 & \centering \multirow{4}{*}{\hyperref[itm:chain]{4.3.1.6}}                & \centering \multirow{4}{*}{$\bullet$} & \centering \multirow{4}{*}{$\bullet$} & \centering \multirow{4}{*}{IR/QA}          & \multicolumn{1}{c}{Quest~\cite{malaviya2023quest}}                \\ 
                          &                                                                                        &                                                                          &                                       &                                       &                                            & \multicolumn{1}{c}{MultiSpanQA~\cite{li2022multispanqa}}          \\
                          &                                                                                        &                                                                          &                                       &                                       &                                            & \multicolumn{1}{c}{FActScore~\cite{min2023factscore}}             \\
                          &                                                                                        &                                                                          &                                       &                                       &                                            & \multicolumn{1}{c}{CD}                                            \\
                                                                                                                                                                                                                                                                                                                           \cmidrule(lr){7-7}
                          &                                                                                        & \centering \multirow{3}{*}{\hyperref[itm:pope]{4.3.1.7}}                 & \centering \multirow{3}{*}{$\bullet$} &                                       & \centering \multirow{3}{*}{IC/QA}          & \multicolumn{1}{c}{MSCOCO~\cite{lin2014mscoco}}                   \\
                          &                                                                                        &                                                                          &                                       &                                       &                                            & \multicolumn{1}{c}{A-OKVQA~\cite{schwenk2022aokvqa}}              \\
                          &                                                                                        &                                                                          &                                       &                                       &                                            & \multicolumn{1}{c}{GQA~\cite{hudson2019visual}}                   \\
                                                                                                                                                                                                                                                                                                                           \cmidrule(lr){7-7}
                          &                                                                                        & \centering \multirow{3}{*}{\hyperref[itm:verbal]{4.3.1.8}}               & \centering \multirow{3}{*}{$\bullet$} &                                       & \centering \multirow{3}{*}{QA}             & \multicolumn{1}{c}{BIG-Bench~\cite{srivastava2023beyond}}         \\ 
                          &                                                                                        &                                                                          &                                       &                                       &                                            & \multicolumn{1}{c}{GSM$8$K~\cite{cobbe2021training}}              \\
                          &                                                                                        &                                                                          &                                       &                                       &                                            & \multicolumn{1}{c}{MMLU~\cite{hendrycks2021measuring}}            \\
                                                                                                                                                                                                                                                                                                                           \cmidrule(lr){7-7}
                          &                                                                                        & \centering \hyperref[itm:interrogate]{4.3.1.9}                           & \centering $\bullet$                  &                                       & \centering QA                              & \multicolumn{1}{c}{CD}                                            \\
                                                                                                                   \cmidrule(lr){3-7}         
                          & \multicolumn{1}{c}{\multirow{5}{*}{\makecell[c]{Adversarial\\Prompting}}}              & \centering \hyperref[itm:flirt]{4.3.2.1}                                 & \centering $\bullet$                  &                                       & \centering IG                              & \multicolumn{1}{c}{CD}                                            \\
                                                                                                                                                                                                                                                                                                                           \cmidrule(lr){7-7}
                          &                                                                                        & \centering \multirow{2}{*}{\hyperref[itm:autodebug]{4.3.2.2}}            & \centering \multirow{2}{*}{$\bullet$} & \centering \multirow{2}{*}{$\bullet$} & \centering \multirow{2}{*}{IR/QA}          & \multicolumn{1}{c}{Natural Questions~\cite{kwiatkowski2019natural}} \\ 
                          &                                                                                        &                                                                          &                                       &                                       &                                            & \multicolumn{1}{c}{CD}                                            \\
                                                                                                                                                                                                                                                                                                                           \cmidrule(lr){7-7}
                          &                                                                                        & \centering \hyperref[itm:invite]{4.3.2.3}                                & \centering $\bullet$                  &                                       & \centering QA                              & \multicolumn{1}{c}{CD}                                            \\
                          &                                                                                        & \centering \hyperref[itm:hypoterm]{4.3.2.4}                              & \centering $\bullet$                  &                                       & \centering QA                              & \multicolumn{1}{c}{CD}                                            \\
                                                                                                                   \cmidrule(lr){3-7}         
                          & \multicolumn{1}{c}{\multirow{6}{*}{\makecell[c]{Proxy\\Model}}}                        & \centering \hyperref[itm:manakul_proxy]{4.3.3.1}                         & \centering $\bullet$                  &                                       & \centering IR/QA                           & \multicolumn{1}{c}{CD}                                            \\
                                                                                                                                                                                                                                                                                                                           \cmidrule(lr){7-7}
                          &                                                                                        & \centering \multirow{3}{*}{\hyperref[itm:reld]{4.3.3.2}}                 & \centering \multirow{3}{*}{$\bullet$} &                                       & \centering \multirow{3}{*}{IR/QA}          & \multicolumn{1}{c}{SQuAD~\cite{rajpurkar2016squad}}               \\ 
                          &                                                                                        &                                                                          &                                       &                                       &                                            & \multicolumn{1}{c}{HotpotQA~\cite{yang2018hotpotqa}}              \\
                          &                                                                                        &                                                                          &                                       &                                       &                                            & \multicolumn{1}{c}{TriviaQA~\cite{joshi2017triviaqa}}             \\
                                                                                                                                                                                                                                                                                                                           \cmidrule(lr){7-7}
                          &                                                                                        & \centering \hyperref[itm:pacchiardi]{4.3.3.3}                            & \centering $\bullet$                  &                                       & \centering IR/QA                           & \multicolumn{1}{c}{CD}                                            \\
                          &                                                                                        & \centering \hyperref[itm:soradetector]{4.3.3.4}                          & \centering $\bullet$                  &                                       & \centering IG                              & \multicolumn{1}{c}{CD}                                            \\
                                                                                                                   \cmidrule(lr){3-7}         
                          & \multicolumn{1}{c}{\multirow{11}{*}{\makecell[c]{Grounding\\Knowledge}}}               & \centering \multirow{3}{*}{\hyperref[itm:purr]{4.3.4.1}}                 &                                       & \centering \multirow{3}{*}{$\bullet$} & \centering \multirow{3}{*}{IR/QA}          & \multicolumn{1}{c}{Natural Questions~\cite{kwiatkowski2019natural}} \\ 
                          &                                                                                        &                                                                          &                                       &                                       &                                            & \multicolumn{1}{c}{StrategyQA~\cite{geva2021aristotle}}           \\ 
                          &                                                                                        &                                                                          &                                       &                                       &                                            & \multicolumn{1}{c}{QRreCC~\cite{anantha2021open}}                 \\
                                                                                                                                                                                                                                                                                                                           \cmidrule(lr){7-7}
                          &                                                                                        & \centering \multirow{3}{*}{\hyperref[itm:cok]{4.3.4.2}}                  & \centering \multirow{3}{*}{$\bullet$} & \centering \multirow{3}{*}{$\bullet$} & \centering \multirow{3}{*}{IR/QA}          & \multicolumn{1}{c}{LC-QuAD~\cite{trivedi2017lcquad}}              \\ 
                          &                                                                                        &                                                                          &                                       &                                       &                                            & \multicolumn{1}{c}{KQA-Pro~\cite{cao2022kqa}}                     \\
                          &                                                                                        &                                                                          &                                       &                                       &                                            & \multicolumn{1}{c}{ScienceQA~\cite{lu2022learn}}                  \\
                                                                                                                                                                                                                                                                                                                           \cmidrule(lr){7-7}
                          &                                                                                        & \centering \hyperref[itm:mixalign]{4.3.4.3}                              & \centering $\bullet$                  & \centering $\bullet$                  & \centering IR/QA                           & \multicolumn{1}{c}{FuzzyQA~\cite{zhang2023knowledge}}             \\
                          &                                                                                        & \centering \hyperref[itm:peng]{4.3.4.4}                                  & \centering $\bullet$                  & \centering $\bullet$                  & \centering IR/QA                           & \multicolumn{1}{c}{CD}                                            \\
                          &                                                                                        & \centering \textbf{\hyperref[itm:tips]{4.3.4.5}}                         &                                       & \centering $\bullet$                  & \centering P                               & \multicolumn{1}{c}{TextWorld~\cite{cote2019textworld}}            \\
                          &                                                                                        & \centering \hyperref[itm:factual_img_retrieval]{4.3.4.6}                 & \centering $\bullet$                  & \centering $\bullet$                  & \centering IG                              & \multicolumn{1}{c}{CD}                                            \\
                          &                                                                                        & \centering \hyperref[itm:sld]{4.3.4.7}                                   &                                       & \centering $\bullet$                  & \centering IG                              & \multicolumn{1}{c}{I$2$P~\cite{schramowski2023safe}}              \\
                                                                                                                   \cmidrule(lr){3-7}         
                          & \multicolumn{1}{c}{\multirow{2}{*}{\makecell[c]{Constraint\\Satisfaction}}}            & \centering \textbf{\hyperref[itm:jha]{4.3.5.1}}                          & \centering $\bullet$                  & \centering $\bullet$                  & \centering P                               & \multicolumn{1}{c}{CD}                                            \\
                          &                                                                                        & \centering \textbf{\hyperref[itm:roboeval]{4.3.5.2}}                     & \centering $\bullet$                  &                                       & \centering P                               & \multicolumn{1}{c}{RoboEval~\cite{hu2024deploying}}               \\
\bottomrule
\end{longtable}
\end{center}
}
\vspace{-15pt}
}

\subsection{White-box Methods}
\label{sec:det_white}

Methods in this section require access to internal weights of the model for hallucination detection.

\subsubsection{Hidden States}

Some approaches utilize intermediate embeddings at different network layers.

\begin{enumerate}

\item \label{itm:saplma}
\citet{azaria2023internal} empirically find that language models attempt to correct themselves after outputting an untruthful claim.
As such, they hypothesize that the internal states of a model must have some understanding of whether the output is correct.
Furthermore, the authors stray away from directly using the token probabilities, even though they have correlation with model accuracy~\cite{varshney2023stitch}, because the complete output sentence's probability is dependent on the length of the generation and appearance frequency of tokens.
\citeauthor{azaria2023internal} present SAPLMA, a simple classifier trained with supervised learning, that takes the activation values of a hidden layer of an LLM as input, and outputs the probability of the generated claim being true. 
The authors choose to collect a new dataset of simple statements with corresponding true/false answers, since popular datasets like FEVER~\cite{thorne2018fever} do not partition statements by topic, and some statements cannot be cleanly classified.
In total, their dataset contains $6$K sentences spanning six topics.
SAPLMA is shown to be able to identify untruthful outputs, even when trained on a held-out dataset on a completely different topic than evaluated on.

\item \label{itm:yao_adv}
\citet{yao2023llm} aim to test the resiliency of foundation models to varying prompts (a form of adversarial prompting further discussed in Section~\ref{sec:det_black_adv}).
They propose perturbing an input prompt with additional tokens so as to make an LLM under test produce a desired hallucination (\eg~modify the original query, ``Who won the $2020$ US election,'' to get the LLM to generate, ``Donald Trump was the victor,'' while its original response was correctly stated as, ``Joe Biden was the victor'').
As the search space of possible tokens to add/replace when developing the adversarial prompt is massive, the work uses a gradient-based token replacing strategy.
Specifically, they define an objective that attempts to find trigger tokens in the direction of the gradient that maximizes the likelihood of the model outputting the desired hallucination.
To evaluate their approach,~\citeauthor{yao2023llm} collect a dataset of truthful facts by feeding questions from Wikipedia~\cite{wikimedia2022wiki} into an LLM.
Then, certain subjects, objects, or predicates in the generated answers are replaced to manually create hallucinations.
With simple prompt modifications, the authors show that the white-box approach is able to induce the specified hallucinations.

\item \label{itm:luna}
The work from~\citet{song2023luna} is described in Appendix~\hyperref[itm:app_luna]{C.1.1.3}.

\end{enumerate}

\subsubsection{Attention Weights}

Attention weight matrices, which are prominent within transformer model architectures, signify the importance the model places on earlier tokens within a generation when predicting future tokens.

\begin{enumerate}

\item \label{itm:opera}
OPERA, proposed by~\citet{huang2024opera}, is a hallucination detection method for LVLMs that makes use of the model's internal attention weights.
When visualizing the attention matrix, the authors find that there exist peculiar column patterns that align with the beginning of a hallucinated phrase.
These \emph{aggregation patterns} usually occur on a non-substantial token like a period or quotation mark, but are deemed to have a large impact on the prediction of future tokens.
As such, this finding led~\citeauthor{huang2024opera} to modify the beam search algorithm~\cite{freitag2017beam} by applying a penalty term to beams wherever an aggregation pattern is detected, and roll back the search to before the pattern arises. 
Their method is shown to reduce hallucinations, and even eliminate possible repetitions in generations.

\end{enumerate}

\subsubsection{Honesty Alignment}

In addition to methods that require hidden states or attention matrices, we also include methods that fine-tune foundation models to better communicate their uncertainty to questions under white-box algorithms, as they require access to model weights for training.

\begin{enumerate}

\item \label{itm:lin_hon}
For example,~\citet{lin2022teaching} collect a calibration dataset of questions and answers from GPT-3 under $21$ types of arithmetic tasks (\eg~add/subtract and multiply/divide), and record how often each task is incorrectly answered.
They aim to fine-tune the LLM to also output its certainty that the prediction is correct.
Consequently,~\citeauthor{lin2022teaching} fine-tune the model with data pairs of a question and the empirical accuracy on the task that the question originates from in the calibration dataset, such that the model is expected to similarly output a probability of accuracy at test-time.
The authors show that the proposed verbalized probability in deployment does correlate with actual accuracy on the tasks.
Specifically,~\citeauthor{lin2022teaching} find that certain problems, like multiplication, are more challenging for GPT-3 to answer correctly.
By calibrating the model on a set of simpler questions (add/subtract), the model is able to generalize its verbal uncertainty to more challenging tasks with multiple possible answer choices.

\item \label{itm:yang_hon}
The work from~\citet{yang2023alignment} is described in Appendix~\hyperref[itm:app_yang_hon]{C.1.3.2}.

\end{enumerate}

\subsection{Grey-box Methods}
\label{sec:det_grey}

Grey-box approaches leverage the probability distributions of tokens output from the model.

\subsubsection{Concept Probabilities}
\label{sec:det_grey_prob}

\begin{enumerate}

\item \label{itm:varsh_prob}
Empirically,~\citet{varshney2023stitch} show that there is a negative correlation between hallucination rate and token probability (\ie~as a token's probability decreases within a sentence, the tendency to hallucinate increases).
Thus, the authors rely on token probabilities to estimate uncertainty of concepts within a generated claim, and they check for correctness by cross-referencing a knowledge-base.
Whenever a concept is found to be conflicting with a fact through verification questions, their method attempts to mitigate the error by prompting the LLM to replace the incorrect claim with the evidence.
The authors query GPT-3.5 for summaries about $150$ different topics sampled from popular Wikipedia subjects, including sports, music, and politics.
To ensure the accuracy of labels, the authors manually label the truthfulness of the first five sentences in each generated article, rather than relying on crowd-sourced labels.
Although effective in the QA setting,~\citeauthor{varshney2023stitch} concede that, in the event token probabilities are not available, some form of heuristic must be used to detect hallucination candidates.

\item \label{itm:lure}
\citet{zhou2024analyzing} show that external models can be developed to automatically \emph{clean} hallucinations.
The authors tackle the issue of object hallucinations that LVLMs experience when describing the content of images.
Through theoretical formulations, the authors show that LVLM responses tend to hallucinate in three settings: when described object classes appear frequently within a description, when a token output has low probability, and when an object appears closer to the end of the response.
As such, their model, LURE, is a fine-tuned LVLM trained on a denoising objective with a training dataset that is augmented to include objects that appear frequently within responses, and replacing objects with low token probabilities or appearing close to the end of the response, with a placeholder tag.
At inference time, tokens are augmented similarly to how they were changed to generate the training dataset, and the LURE LVLM is prompted to denoise hallucinations by filling in uncertain objects.

\item \label{itm:saycanpay}
SayCanPay, proposed by~\citet{hazra2024saycanpay}, builds off of the SayCan framework~\cite{ichter2023do} to improve the expected payoff of following a plan specified by a language model.
Within our hallucination definition, this goal translates to increasing the desirability of generations by improving the likelihood of the model achieving higher rewards.
The authors propose three different strategies for planning: Say, SayCan, and SayCanPay.
Say methods greedily choose next actions based only on token probabilities.
SayCan approaches also take the success rate of the chosen action into consideration.
Finally, SayCanPay additionally estimates the expected payoff from following the plan with some heuristic.
\citeauthor{hazra2024saycanpay} learn this Pay model with regression on an expert trajectory dataset.
Combining all three models together minimizes the likelihood that a generated plan contains conflicting infeasible action calls, while maximizing the efficiency of the task completion.

\end{enumerate}

\subsubsection{Conformal Prediction}

Another range of works estimate the uncertainty of a model output with conformal prediction so as to provide statistical guarantees on the likelihood of predictions being correct~\cite{shafer2008tutorial}.

\begin{enumerate}

\item \label{itm:quach}
\citet{quach2024conformal} propose conformal language modeling to build a set of possible candidate responses to a test prompt, while calibrating algorithm parameters on a held-out dataset of independent prompts and their corresponding admission functions, which check whether a model output meets the criteria of an input prompt.
In their algorithm, the authors calibrate thresholds for three separate scoring functions that test for generation quality, similarity with other responses, and model confidence using ``Learn then Test''~\cite{angelopoulos2022learn}.
At inference time, given scoring functions, calibrated thresholds, and an input prompt, the method samples outputs from the model and adds them to a prediction set if they meet quality and similarity thresholds, until the whole set is guaranteed to meet the user-defined confidence parameter.

\item \label{itm:kumar}
The work from~\citet{kumar2023conformal} is described in Appendix~\hyperref[itm:app_kumar]{C.2.2.2}.

\item \label{itm:knowno}
While the previous hallucination mitigation works presented using conformal prediction are solely applied to QA settings,~\citet{ren2023robots} are the first to apply conformal prediction of foundation models to robotic tasks.
The authors are motivated by a desire for language-conditioned robots to understand when they are uncertain about the next action to take, such that they can ask for help in those cases (while minimizing frequency of clarifications).
Because LLM generations with different length sequences inherently produce different complete sentence probabilities, the authors propose framing the control task as a multiple-choice problem, like~\citet{kumar2023conformal}.
Their approach, KnowNo, prompts an LLM to generate a possible set of next actions to take in multiple choice form. 
They first collect and hand-label a calibration dataset of pairs of held-out instructions and the probability of the model choosing the best action to take next.
At test-time, the model outputs a set of actions to take and KnowNo eliminates actions with token probabilities less than a calibrated certainty from a user.
If there are still multiple actions left in the prediction set after eliminating uncertain actions, the model queries a human for help choosing the next action.
\citeauthor{ren2023robots} show that KnowNo deviates from the user-defined error rate least often compared to methods that do not use conformal prediction and has the highest success-to-clarification ratio. 
Additionally, the authors deploy KnowNo to a real UR5 robot arm, where it is tasked with sorting foods on a table by order of user preference, and disposing of undesired objects conditioned on ambiguous instructions. 
However, several assumptions had to be made to produce the demonstrated results, including having access to next token probabilities, having resources to collect a large calibration dataset, presuming people will faithfully provide help when asked for it, and using ground-truth vision to fully ground the environmental objects with the text input to the model.

\item \label{itm:liang_intro}
\citet{liang2024introspective} extend the KnowNo methodology by incorporating an introspective planning step using a previously constructed knowledge-base of experiences, which tends to (1) enhance quality of generated plans, and (2) improve interpretability of decisions.
Specifically, introspective planning first constructs a knowledge-base containing training pairs of tasks, observations, and valid plans, which the LLM is prompted to generate explanations behind why they are reasonable.
Each experience is stored with a key as an embedding of the original instruction.
During inference, given a new test instruction, their method queries the database to find the key with the closest embedding to that of the new instruction.
This previous experience and reasoning is fed into the model to generate a set of candidate plans to follow.
Finally, the remainder of the algorithm follows the same process as KnowNo to calibrate and narrow down the prediction set to fall within a desired error rate.
\citeauthor{liang2024introspective} evaluate their method on the KnowNo dataset~\cite{ren2023robots} and an augmentation of the original dataset that considers more safety-critical tasks with ambiguous instructions (\eg~making sure not to place metal objects in a microwave when tasked with heating a bowl).

\item \label{itm:heracles}
\citet{wang2024conformal} aim to provide additional guarantees on completing the task provided within the natural language instruction.
To do so, the authors propose a novel task specification, LTL-NL, which combines linear-temporal-logic (LTL) descriptions with natural language from a user instruction, which the authors claim is easier to define than classical LTL specifications.
Given this specification, a symbolic task planner chooses a sub-task to complete next and an LLM generates plans for each sub-task, respectively.
Like~\citet{ren2023robots} and~\citet{liang2024introspective},~\citeauthor{wang2024conformal} apply conformal prediction to minimize the number possible actions to take next within some desired error rate.
However, rather than directly asking a user for assistance when there is high uncertainty in the next action to take (or when there are environmental constraints), their method, HERACLEs, samples a new sub-task to complete from the task planner. 
If on the other hand, the task planner is unable to provide a new sub-task, HERACLEs requests help from the user.
The authors deploy the model and baselines in a custom-made 3D mobile manipulator simulator that allows for evaluation of methods that use LTL specifications.
For example, the instruction, ``Deliver Apple to A'' is converted to LTL specifications and fed into HERACLEs to navigate the robot to (1) pick up the requested apple and (2) drop it off at location A.
With experimentation, the authors find that their method achieves higher task completion rate on missions requiring more sub-tasks, outperforming baseline planners that do not utilize LTL specifications.

\end{enumerate}

\subsection{Black-box Methods}
\label{sec:det_black_box}

Black-box algorithms only rely on the input prompts and output predictions from the model, without making assumptions on the availability of the hidden state, nor the token probabilities.

\subsubsection{Analyzing Samples from Model}
\label{sec:det_analyze}

As a result, several works of this type sample multiple sentences from an LLM, and measure the similarity of the information present in all samples.

\begin{enumerate}

\item \label{itm:selfcheck}
For example, SelfCheckGPT, proposed by~\citet{manakul2023self}, samples multiple responses (each of which may contain many sentences) from an LLM to a single query, and measures the consistency among the varied responses through a study with five different approaches.
SelfCheckGPT with BERTScore~\cite{zhang2020bertscore}, for example, computes a similarity score between two sentences from different sampled outputs.
Intuitively, a hallucination is detected when the similarity score for a sentence is low across all other samples.
Other consistency-checking methods the authors consider include using an automatic multiple-choice QA system conditioned on the responses, relying on a proxy-LLM which has access to token probabilities, training an external classifier to predict contradictions (coined SelfCheck-NLI), and directly prompting the LLM to evaluate whether any sentence can be supported by another sampled response.
Before evaluating the proposed methods, the authors find that there is no standardized hallucination detection dataset for question-answering.
Furthermore, the existing hallucination dataset from~\citet{liu2022token} is generated by manually replacing tokens in actual facts, which may not represent generations from real language models.
Thus,~\citeauthor{manakul2023self} choose to curate their own evaluation dataset by querying GPT-3 for articles on topics from the WikiBio dataset~\cite{lebret2016neural}.
They then manually label the accuracy of each sentence in each generated article for a total of $1908$ sentences across $238$ summaries. 
As expected,~\citeauthor{manakul2023self} find that sampling more responses from the model leads to better estimation of the validity of a claim, but is slower to compute.

\item \label{itm:halo}
The work from~\citet{elaraby2023halo} is described in Appendix~\hyperref[itm:app_halo]{C.3.1.2}.

\item \label{itm:debate}
Rather than analyzing the responses of a single language model,~\citet{du2023improving} take an ensemble approach.
Specifically, they propose pitting multiple instances of an LLM into a debate of the correct answer to a question.
In practice, several agents are given the same question and predict a response.
Over multiple iterations, all the responses of other agents are concatenated, and fed in as additional context to each model, and a new response is sampled.
\citeauthor{du2023improving} first identify whether the proposed debate method can improve the \emph{reasoning} of language models through three offline tasks with increasing difficulty in a custom benchmark: (1) simple arithmetic, (2) grade-school math~\cite{hendrycks2021measuring}, and (3) predicting the next best move to take in a game of chess.
They additionally evaluate the \emph{accuracy} of debate results in (1) generating biographies of famous computer scientists from Wikipedia, (2) general exam knowledge~\cite{hendrycks2021measuring}, and (3) confirming the validity of next moves in chess.
The authors show that a combination of their iterative ensemble approach with chain-of-thought reasoning mitigates individual hallucinations (as agents tend to converge on a single consensus) and increases QA accuracy. 
While only tested on offline datasets, the approach could be utilized within a simulation framework, like the one presented by~\citet{park2023generative}, where, for example, multiple language agents may debate about plans to make.
It is important to note that~\citeauthor{du2023improving} evaluate the accuracy of biography generations by querying ChatGPT for the consistency between a fact and generated response.
We hypothesize that, like~\citet{yu2015lsun}, this automatic labeling scheme will result in lower quality labels than manual annotation.

\item \label{itm:clara}
CLARA is a framework engineered by~\citet{park2024clara} that predicts when an instruction provided to a robotic system controlled by an LLM may be ambiguous or infeasible.
Intuitively, if a language model is uncertain about an instruction, it might output diverse (or conflicting) actions to other instructions that hold similar information.
Thus, CLARA samples several sets of concepts from the original prompt, randomly orders them to assemble multiple inputs to the model, and passes them as input to the language model under test.
The method computes the average similarity of pairs of output actions in an embedding space, over all outputs.
Next, to check for infeasibility of the original goal, the foundation model is provided the possible action space of the robot, environmental observation, and goal, and is prompted to output whether the desired task is practical.
In the event the model is uncertain from the multi-prompting step, but the goal is feasible, CLARA asks for clarification from the user with reasoning for why it is uncertain.
In addition to evaluating CLARA on SaGC, a custom dataset described in Appendix~\ref{sec:app_eval_code_gen}, the authors also put their model to the test on a tabletop manipulator robot in simulation and the real world.
While the method achieves a reasonable success rate on robotic pick-and-place tasks with real user instructions (where other discussed methods are primarily evaluated in QA settings), there are still failure cases where the model hallucinates during uncertainty reasoning and feasibility prediction.

\item \label{itm:mundler}
Another set of works explicitly identify contradictions among responses, instead of estimating similarity.
Naturally, detecting a self-contradiction is guaranteed to reveal an invalid claim.
\citet{mundler2024self} pose that removing detected conflicting information will increase the validity of a generated response.
As such, the authors suggest a solution that finds important concepts within a response to be evaluated, prompts the generation model to generate more information about each of the concepts, and uses a separate analyzer language model to evaluate the consistency of pairs of sentences on the same concept.
Any sentences that are found to be conflicting are revised by the analyzer model, before being output to the user.
The authors ask four language models to generate $360$ summaries for $30$ diverse topics from WikiBio~\cite{lebret2016neural} and Wikipedia~\cite{wikimedia2022wiki} and have three human annotators manually label any instances of self-contradiction, inaccurate sentences, and unverifiable statements, leading to high-quality labels.
With respect to Definition~\ref{def:hal}, the annotators are identifying cases of noncompliant and irrelevant generations to decide which statements are hallucinations.

\item \label{itm:chain}
Rather than relying on another language model to analyze the correctness of predictions,~\citet{dhuliawala2023chain} utilize chain-of-thought reasoning to prompt an LLM to generate possible verification questions about its original responses.
If there are conflicts in answers to the verification questions, the LLM is prompted to regenerate its output with updated context and conflicting reasoning.
One of the first evaluations the authors perform is on a custom set of $56$ questions with topics collected from Wikipedia, each of which have multiple correct answers for a total of around $600$ ground-truth entities.

\item \label{itm:pope}
The work from~\citet{li2023evaluating} is described in Appendix~\hyperref[itm:app_pope]{C.3.1.7}.

\item \label{itm:verbal}
The work from~\citet{xiong2024can} is described in Appendix~\hyperref[itm:app_verbal]{C.3.1.8}.

\item \label{itm:interrogate}
The work from~\citet{yehuda2024interrogatellm} is described in Appendix~\hyperref[itm:app_interrogate]{C.3.1.9}.

\end{enumerate}

\subsubsection{Adversarial Prompting}
\label{sec:det_black_adv}

Works specializing in adversarial prompting attempt to test the robustness of models to varying inputs that may coerce the model into producing out-of-distribution results.

\begin{enumerate}

\item \label{itm:flirt}
For example,~\citet{mehrabi2023flirt} apply adversarial prompting to text-to-image foundation models, like Stable Diffusion~\cite{schramowski2023safe}, to generate offensive images.
With respect to Definition~\ref{def:hal}, their framework, FLIRT, is essentially testing the tendency of foundation models to hallucinate undesired generations in deployment.
FLIRT uses an adversarial language model to predict a prompt to input to the image generator, scores the generated image for the presence of undesirable traits using an external classifier, re-prompts the adversary to produce a new instruction conditioned on the findings of the classifier, and repeatedly generates images until the adversary successfully prompts the test model to output an undesirable result.
\citeauthor{mehrabi2023flirt} define objective functions conditioned on the score output by external classifiers to maximize diversity of adversarial prompts and minimize toxicity so as to pass text filters that detect malicious inputs, while improving attack effectiveness.
The authors form a large set of prompts with varying levels of detail across different sexual, violent, hate,~\etc~contexts, inspired by~\citet{schramowski2023safe}.
Each prompt corresponds to one of three test splits that change the type of toxicity, level of detail, and phrasing in the prompt.
Tangentially,~\citeauthor{mehrabi2023flirt} evaluate FLIRT on text-to-text models by similarly collecting adversarial prompts with varying vulgarity.

\item \label{itm:autodebug}
Another work from~\citet{yu2023automatic} presents the AutoDebug framework for automatically sampling and updating several prompts for use in adversarial testing of the language model.
\citeauthor{yu2023automatic} argue that evaluating information-retrieval LLMs on popular datasets like Wikipedia provides an over-approximation on the accuracy of these models since they have been overfit on the same data, possibly leading to memorization.
Thus, the authors specifically explore adversarial testing under the case that the model predicts a correct response when provided relevant context, but generates an incorrect prediction when the evidence is modified.
They apply two different modification approaches: replacing tokens within the context to provide incorrect facts, and adding additional relevant facts to the prompt that may make it difficult to pick out the most important details.
The authors collect a new dataset by applying their adversarial generator to Natural Questions~\cite{kwiatkowski2019natural} and RealtimeQA~\cite{kasai2023realtime}, with additional human filtering to ensure plausible answers are collected.

\item \label{itm:invite}
The work from~\citet{ramakrishna2023invite} is described in Appendix~\hyperref[itm:app_invite]{C.3.2.3}.

\item \label{itm:hypoterm}
The work from~\citet{uluoglakci2024hypotermqa} is described in Appendix~\hyperref[itm:app_hypoterm]{C.3.2.4}.

\end{enumerate}

All in all, adversarial prompting is an effective method for identifying robustness of models to unseen inputs, which can be used to develop stronger input filters or fine-tune the model for decreased hallucination tendency.

\subsubsection{Proxy Model}
\label{sec:proxy_methods}

Certain black-box works rely on an external, proxy model to detect and mitigate hallucinations.

\begin{enumerate}

\item \label{itm:manakul_proxy}
One such method is used as a baseline within the SelfCheckGPT article~\cite{manakul2023self}.
As many language foundation models do not provide access to token probabilities, the authors use an open-source proxy LLM that does provide token probabilities as an estimate of the original output's probability.
They find that using proxy LLMs for probability estimation and hallucination detection successfully is highly variable.
The accuracy of detection is dependent on the complexity of the LLM itself, as well as the training data of the proxy LLM (\ie~models trained on independent datasets from the original LLM will have different generation patterns).
Refer to the description of Method ID~\hyperref[itm:selfcheck]{4.3.1.1} for a discussion on the custom dataset used for evaluation.

\item \label{itm:reld}
Within this section, we also include works that use an external trained classifier to detect hallucinations.
For example,~\citet{chen2023hallucination} curate a dataset of QA dialogue from LLM generated responses.
They apply a composition of metrics to assess quality of responses, including a self-assessment from the LLM comparing the ground-truth and predicted text, human-labeled, and machine metrics (\eg~BERTScore~\cite{zhang2020bertscore}, F1 score, BLEU~\cite{papineni2002bleu}, \etc).
Their hallucination discriminator, RelD, is trained on the dataset in multiple separate phases, each using a different objective: regression, multi-class classification, and finally binary classification.
Through experiments, they find that RelD closely aligns with human evaluators' original predictions.

\item \label{itm:pacchiardi}
The work from~\citet{pacchiardi2024how} is described in Appendix~\hyperref[itm:app_pacchiardi]{C.3.3.3}.

\item \label{itm:soradetector}
The work from~\citet{chu2024sora} is described in Appendix~\hyperref[itm:app_soradetector]{C.3.3.4}.

\end{enumerate}

\subsubsection{Grounding Knowledge}
\label{sec:grounding_know}

In knowledge grounding tasks, a language model is tasked with identifying evidence from an external knowledge-base that supports claims within a summary. 
Although seemingly irrelevant to decision-making scenarios, similar methods to ones discussed in this section may be applied in planning tasks to identify observations that are most relevant to predicting the next action, or to generate reasoning behind a specified plan.

\begin{enumerate}

\item \label{itm:purr}
PURR, proposed by~\citet{chen2023purr}, is a denoising agent, like LURE (discussed in Section~\ref{sec:det_grey_prob}), that is trained in an unsupervised fashion given evidence from online sources, a clean (correct) summary, and a noisy (hallucinated) summary.
The model learns to denoise the incorrect summary to the clean statement.
During deployment, given a possibly hallucinated claim, a question generation model queries online sources for evidence about the claim, and PURR generates a cleaned version of the original summary with said evidence.

\item \label{itm:cok}
The work from~\citet{li2024chain} is described in Appendix~\hyperref[itm:app_cok]{C.3.4.2}.

\item \label{itm:mixalign}
The work from~\citet{zhang2023knowledge} is described in Appendix~\hyperref[itm:app_mixalign]{C.3.4.3}.

\item \label{itm:peng}
\citet{peng2023check} aim to add plug-and-play modules to an LLM to make its outputs more accurate, since these large foundation models cannot feasibly be fine-tuned whenever there is new information.
Their work formulates the user conversation system as a Markov decision process (MDP) whose state space is an infinite set of dialogue states which encode the information stored in a memory bank, and whose discrete action space includes actions to call a knowledge consolidator to summarize evidence, to call an LLM prompt engine to generate responses, and to send its response to the user if it passes verification with a utility module.
The proposed LLM-Augmenter has a memory storing dialogue history, evidence from the consolidator, set of output responses from an LLM, and utility module results.
Its policy is trained in multiple phases with REINFORCE~\cite{williams1992simple} starting with bootstrapping from a rule-based policy designed from domain experts, then learning from simulators, and finally, from real users. 
\citeauthor{peng2023check} deploy LLM-Augmenter to two different information retrieval domains: news and customer service.
For the news application, the authors retrieve relevant news articles by crawling Reddit news forums for $1.3$K articles, following the DSTC7 Track $2$~\cite{yoshino2019dialog} approach.
Similarly, to emulate the required knowledge of a customer service chat bot, the authors use the data from DSTC11 Track $5$~\cite{kim2023task}, which holds $14.7$K examples of user reviews and frequently answered questions. 
The authors find that access to ground-truth knowledge drastically improves QA results, and feedback from the utility module and knowledge consolidator help to provide more accurate answers.

\item \label{itm:tips}
Evaluated in decision-making settings, Introspective Tips~\cite{chen2023introspective} provide concise, relevant information to a language planner to learn to solve problems more efficiently.
Intuitively, summaries that collect information over all past experiences may be long and contain unnecessary details.
In contrast, tips are compact information with high-level guidance that can be learned from one's own experiences, from other demonstrations, and from other tasks in a similar setting.
\citeauthor{chen2023introspective} show that providing low-level trajectories is less effective than tips on simulated planning tasks.
Additionally, with expert demonstrations, the LLM learns faster with fewer number of failed trials than with just past experience alone. 
However, one limitation identified in the study is that the LLM underperforms in unseen, low-difficulty missions where it has issues generating general tips for zero-shot testing.

\item \label{itm:factual_img_retrieval}
The work from~\citet{lim2024addressing} is described in Appendix~\hyperref[itm:app_factual_img_retrieval]{C.3.4.6}.

\item \label{itm:sld}
The work from~\citet{schramowski2023safe} is described in Appendix~\hyperref[itm:app_sld]{C.3.4.7}.

\end{enumerate}

\subsubsection{Constraint Satisfaction}

There is also additional work in creating black-box algorithms for ensuring decision plans generated by foundation models meet user-defined goal specifications and system constraints, like their grey-box counterpart developed by~\citet{wang2024conformal}.

\begin{enumerate}

\item \label{itm:jha}
Because these models under test provide their results in text form, it is natural to apply formal method approaches (\eg~satisfiability modulo theory, SMT, solvers) to verify the satisfaction of generated plans.
For example,~\citet{jha2023counterexample} prompt an LLM planner with a problem formulated with first order constraints to predict a set of actions to complete the task.
The output plan is input to an SMT solver to check for any infeasibilities in the program, and any counterexamples found are used to iteratively update the prompt and generate new plans.
This counterexample approach is much faster than relying on combinatorial search methods that find a plan from scratch.
However, the quality of generated plans and the number of iterations before a successful plan is generated are heavily dependent on the LLM generator itself, with similar reasons to the proxy-model used by~\citet{manakul2023self}.
In particular, the authors explore the capability and efficiency of state-of-the-art large language models to solve block-world planning tasks~\cite{gupta1992complexity}.
For every experiment, each LLM is fed the initial random setup of a finite number of blocks in a scene, and the desired goal setup in the form of first-order constraints.
Inoperable plans are detected by the Z3 SMT solver~\cite{moura2008z3}, who iteratively works with the LLM to approach a feasible solution using counterexamples.

\item \label{itm:roboeval}
Another work from~\citet{hu2024deploying} develops a RoboEval benchmark to test generated plans on real robots, in a black-box manner.
Like~\citet{wang2024conformal}, the authors introduce their own extension of LTL formulations, known as RTL, which specifies temporal logic at a higher, scenario-specific, level, while abstracting away constraints that are not dependent on available robot skills.
RTL and LTL-NL are easier to read and define than classic LTL methods.
RoboEval utilizes the provided RTL formulation of a problem, a simulator, and evaluator to systematically check whether the output meets requested goals.
Furthermore, to check for robustness of the model to varied instructions,~\citeauthor{hu2024deploying} hand-engineer paraphrased sentences within an offline dataset that should ideally result in the same task completion.
Primary causes of failures were found to be a result of generated code syntax errors, attempting to execute infeasible actions on the robot, and failing RTL checks.

\end{enumerate}

Like adversarial prompting approaches, testing generated plans on robots in diverse scenarios enable researchers to design more robust systems that hallucinate less frequently at test-time.

\section{Guidelines on Current Methodologies}
\label{sec:guidelines}

In section~\ref{sec:detection}, we present a taxonomy of hallucination detection and mitigation algorithms in various deployment settings. 
Combining our findings from our extensive review, we now present guidelines for choosing hallucination intervention algorithms and metrics for different environments.
As such, we hope these guidelines will enable developers to follow a standardized procedure to define hallucinations for their deployment context, design intervention algorithms, and evaluate efficacy before integrating hallucination-prone models into systems with humans at risk.
Even while the field of LVLMs evolves rapidly, we argue our guidelines are general enough to continue to assist researchers in the near future.
Additional considerations when using deep learning methods are discussed in Appendix~\ref{sec:app_beware}.

\subsection{Process of Choosing and Integrating Intervention Algorithms}

\begin{wrapfigure}[12]{r}{0.5\textwidth}
  \centering
  \includegraphics[width=0.9\linewidth]{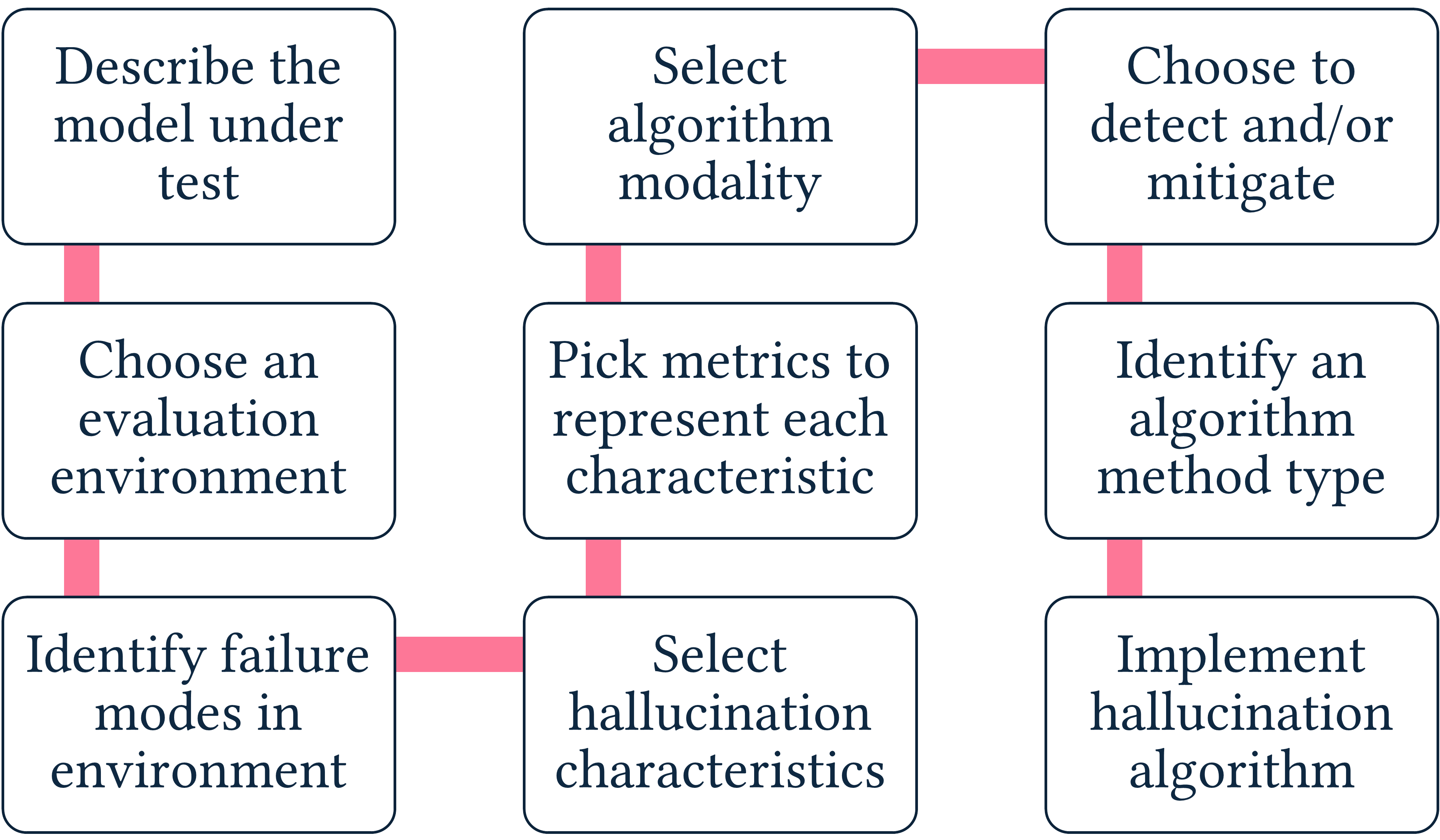} 
  \caption{\textbf{Design process of hallucination intervention methods.}}
  \label{fig:guide_flow}
\end{wrapfigure}

We split the design process of hallucination detection and mitigation methods into nine steps, shown in Figure~\ref{fig:guide_flow}.

\paragraph{Describe the Model Under Test}
The model under test is the specific L(V)LM that has the potential to hallucinate in a deployment environment.
Describing the model requires understanding its space of inputs and outputs, and whether designers have access to model weights and/or generation probability distributions.
Understanding the model under test is important for narrowing down possible hallucination intervention methods.

\paragraph{Choosing an Evaluation Setting}
Evaluation settings are datasets, simulators, or real-world environments where the model under test and hallucination intervention algorithm are being tested hand in hand.
Several possible evaluation settings are described in Appendix~\ref{sec:app_evaluation}, categorized by application area (\eg~image captioning, conversational QA, control, code generation).
Developers should choose an evaluation setting that most closely resembles the intended deployment area to minimize the sim-to-real gap between evaluation and final integration~\cite{he2024bridging, salvato2021crossing, zhao2020sim}. 
A too large of a sim-to-real gap results in a poor understanding of the true capabilities of the deployed model.
However, designers need to balance the tradeoffs between the cost of developing a new high-fidelity simulator or collecting representative offline data, and using off the shelf low-fidelity environments or datasets that may not truly cover the distribution of the intended deployment area.

\paragraph{Identifying Failure Modes in Evaluation Setting}
Failure modes are directly related to the characteristics of hallucinations that need to be detected or mitigated.
These modes are identified in one of three primary ways.
Firstly, designers can choose failure modes based on the hard constraints of the evaluation setting.
For example, in decision-making contexts, failure modes may represent a collision space that a robot should not enter, and in QA settings, failure modes could be defined by a set of ground-truth answers that a model generates conflicting results against.
Second, stakeholders impacted by model decisions will have personal preferences for how the model under test should act in deployment.
As such, the next set of failure modes are defined by acting outside of the range of desired behaviors.
Finally, as discussed in Section~\ref{sec:intro}, it is nearly impossible to account for all possible failure modes at inference time.
Thus, intervention algorithm designers should also query the L(V)LM under test with different inputs to identify other undesired and irrelevant generations.
Furthermore, many models are released alongside with a \emph{model card} with details of failures found during its design stage~\cite{gallifant2024peer, dalle32024system, meta2024model}, which developers can use to identify additional modes to consider.
This final sampling procedure will not cover all possible remaining failure modes, but will improve coverage rate.
Through this three step process, engineers gain an understanding of what behaviors should be considered as hallucinations in the particular evaluation setting.
Although generalized detection/mitigation methods are preferred, identifying these specific subsets of hallucinations will allow engineers to quantitatively evaluate the efficacy of intervention approaches prior to deployment.

\paragraph{Selecting Hallucination Characteristics}
Given that designers have selected a set of failure modes, or hallucination behaviors, they can now categorize each mode as one of three content characteristics defined in Section~\ref{sec:what_are_hallu}: compliance, desirability, and relevancy.
Examples of categorizing failure modes are shown in Table~\ref{tab:examples}.
It might be the case that one or more of the categories do not end up with any failure modes.
In this case, remaining unmatched characteristics are deemed irrelevant to the development of the hallucination intervention algorithm at this stage.
Finally, if stakeholders also care about cases that the L(V)LM is attempting to deceive them through syntactically plausible generations, the plausibility characteristic is also relevant to hallucination detection or mitigation. 

\paragraph{Picking Metrics per Characteristic}
Intervention algorithm designers can now choose metrics to measure the efficacy of model generations and hallucination detection/mitigation methods in meeting the defined requirements of each characteristic.
Metrics for each chosen characteristic will differ depending on the deployment context and outputs of the model under test.
As such, we provide examples of current metrics in literature categorized by characteristic and application area in Table~\ref{tab:metrics} from our review in Section~\ref{sec:detection}.

\begin{table*}[p]
\centering
\caption{
\textbf{Common datasets/simulators and metrics across settings and hallucination characteristics.}
Each metric is accompanied by references that use the metric as the specified characteristic. The same metric could be applied as different characteristics.
}
\label{tab:metrics}
\resizebox{\textwidth}{!}{
\begin{tabular}{m{0.5in}m{1.5in}m{1.6in}m{1.6in}m{1.4in}m{1.4in}m{0cm}}
\toprule
\multirow{2}{*}{\centering\textbf{Setting}} 
                & \centering\textbf{\multirow{2}{*}{\makecell[c]{Common\\Datasets/Simulators}}}
                             & \multicolumn{4}{c}{\textbf{Characteristic}}                                                      & \\
                             \cmidrule(lr){3-6}
                &            & \textbf{Compliance}      
                                                  & \textbf{Desirability}      
                                                                       & \textbf{Relevancy}       
                                                                                            & \textbf{Plausibility}   & \\
\midrule
\vspace{4pt}\makecell[{b{p{0.5in}}}]{QA\vspace{8pt}}                           
                & \vspace{4pt}\makecell[{b{p{1.5in}<{\raggedright}}}]
                  {
                  --- TriviaQA~\cite{joshi2017triviaqa},\\
                  --- GSM$8$K~\cite{cobbe2021training},\\
                  --- HotpotQA~\cite{yang2018hotpotqa},\\
                  --- MMLU~\cite{hendrycks2021measuring},\\
                  --- Natural Questions~\cite{kwiatkowski2019natural}
                  \vspace{8pt}
                  }
                             & \vspace{4pt}\makecell[{b{p{1.6in}<{\raggedright}}}]
                               {
                               --- Accuracy$^\dagger$\\\cite{azaria2023internal, kumar2023conformal, elaraby2023halo, du2023improving, mundler2024self, dhuliawala2023chain, manakul2023self, xiong2024can, yehuda2024interrogatellm, yu2023automatic, chen2023hallucination, pacchiardi2024how, chen2023purr, li2024chain, zhang2023knowledge, peng2023check},\\
                               --- Contradictions\\\cite{elaraby2023halo, du2023improving, mundler2024self, dhuliawala2023chain, yehuda2024interrogatellm, pacchiardi2024how},\\
                               --- Concept Probabilities\\\cite{varshney2023stitch, quach2024conformal, manakul2023self},\\
                               --- BERTScore~\cite{manakul2023self, ramakrishna2023invite, peng2023check},\\
                               --- BLEU~\cite{ramakrishna2023invite, li2024chain, peng2023check},\\
                               --- Adversarial Success Rate\\\cite{yao2023llm, yu2023automatic},\\
                               --- Coverage~\cite{kumar2023conformal, zhang2023knowledge},\\
                               --- METEOR~\cite{ramakrishna2023invite, peng2023check},\\
                               --- SummaC~\cite{elaraby2023halo},\\
                               --- FactScore~\cite{dhuliawala2023chain},\\
                               --- Cosine Similarity~\cite{yehuda2024interrogatellm},\\
                               --- Sentence-Bert~\cite{yu2023automatic},\\
                               --- AlignScore~\cite{ramakrishna2023invite}
                               \vspace{8pt}
                               }
                                                  & \vspace{4pt}\makecell[{b{p{1.6in}<{\raggedright}}}]
                                                    {
                                                    --- Calibration Error\\\cite{lin2022teaching, xiong2024can},\\
                                                    --- Succinctness~\cite{song2023luna, peng2023check},\\
                                                    --- Over-Conservativeness\\\cite{yang2023alignment},\\
                                                    --- \# of Clarifying Questions\\\cite{zhang2023knowledge},\\
                                                    --- Prudence~\cite{yang2023alignment},\\
                                                    --- Preservation~\cite{chen2023purr},\\
                                                    --- TOXIGEN Classifier~\cite{mehrabi2023flirt}
                                                    \vspace{8pt}
                                                    }
                                                                       & \vspace{4pt}\makecell[{b{p{1.4in}<{\raggedright}}}]
                                                                         {
                                                                         --- Perplexity~\cite{song2023luna, mundler2024self},\\
                                                                         --- Cross Encoder~\cite{chen2023purr}
                                                                         \vspace{8pt}
                                                                         }  
                                                                                            & \vspace{4pt}\makecell[{b{p{1.4in}<{\raggedright}}}]
                                                                                              {
                                                                                              --- ROUGE\\\cite{quach2024conformal, ramakrishna2023invite, peng2023check},\\
                                                                                              --- Semantic Preciseness\\\cite{song2023luna, chen2023hallucination}
                                                                                              \vspace{8pt}
                                                                                              }  
                                                                                                                                & \\
\hline
\vspace{8pt}\makecell[{b{p{0.5in}}}]{IC\vspace{8pt}}               
                & \vspace{8pt}\makecell[{b{p{1.5in}<{\raggedright}}}]
                              {
                              --- MSCOCO~\cite{lin2014mscoco},\\
                              --- GQA~\cite{hudson2019visual},\\
                              --- Visual Genome~\cite{krishna2017visual},\\
                              --- A-OKVQA~\cite{schwenk2022aokvqa}
                              \vspace{8pt}
                              }
                             & \vspace{8pt}\makecell[{b{p{1.6in}<{\raggedright}}}]
                                           {
                                           --- LVLM-Based Scoring\\\cite{huang2024opera, zhou2024analyzing, li2023evaluating},\\
                                           --- CHAIR~\cite{huang2024opera, zhou2024analyzing, li2023evaluating},\\
                                           --- POPE$^\dagger$~\cite{huang2024opera, zhou2024analyzing, li2023evaluating},\\
                                           --- Co-Occurance~\cite{zhou2024analyzing},\\
                                           --- Concept Probabilities\\\cite{zhou2024analyzing},\\
                                           --- BLEU~\cite{zhou2024analyzing},\\
                                           --- BertScore~\cite{zhou2024analyzing},\\
                                           --- CLIP Score~\cite{zhou2024analyzing},\\
                                           --- METEOR~\cite{zhou2024analyzing}
                                           \vspace{8pt}
                                           }  
                                                  & \vspace{8pt}\makecell[{b{p{1.6in}<{\raggedright}}}]
                                                                {
                                                                --- CIDER Human Alignment\\\cite{zhou2024analyzing},\\
                                                                --- Detailedness~\cite{huang2024opera}
                                                                \vspace{8pt}
                                                                }            
                                                                       & \vspace{8pt}\makecell[{b{p{1.4in}<{\raggedright}}}]
                                                                                     {
                                                                                     --- CHAIR~\cite{huang2024opera, zhou2024analyzing, li2023evaluating},\\
                                                                                     --- POPE$^\dagger$~\cite{huang2024opera, zhou2024analyzing, li2023evaluating}
                                                                                     \vspace{8pt}
                                                                                     }  
                                                                                            & \vspace{8pt}\makecell[{b{p{1.4in}<{\raggedright}}}]
                                                                                                          {
                                                                                                          --- Perplexity~\cite{huang2024opera},\\
                                                                                                          --- ROUGE~\cite{zhou2024analyzing},\\
                                                                                                          --- SPICE~\cite{zhou2024analyzing}
                                                                                                          \vspace{8pt}
                                                                                                          }  
                                                                                                                                & \\
\hline
\vspace{8pt}\makecell[{b{p{0.5in}}}]{IG\vspace{8pt}}               
                & \vspace{8pt}\makecell[{b{p{1.5in}<{\raggedright}}}]
                              {
                              --- MSCOCO~\cite{lin2014mscoco},\\
                              --- LSUN~\cite{yu2015lsun},\\
                              --- LAION-$400$M~\cite{schuhmann2021laion},\\
                              --- I$2$P~\cite{schramowski2023safe}
                              \vspace{8pt}
                              }
                             & \vspace{8pt}\makecell[{b{p{1.6in}<{\raggedright}}}]
                                           {
                                           --- Contradictions~\cite{chu2024sora, lim2024addressing},\\
                                           --- Accuracy$^\dagger$~\cite{chu2024sora},\\
                                           --- LVLM-Based Scoring~\cite{chu2024sora},\\
                                           --- CLIP Score~\cite{schramowski2023safe}
                                           \vspace{8pt}
                                           }  
                                                  & \vspace{8pt}\makecell[{b{p{1.6in}<{\raggedright}}}]
                                                                {
                                                                --- NudeNet Classifier\\\cite{mehrabi2023flirt, schramowski2023safe},\\
                                                                --- Q$16$ Classifier~\cite{mehrabi2023flirt, schramowski2023safe},\\
                                                                --- Diversity~\cite{mehrabi2023flirt},\\
                                                                --- Monetary Cost~\cite{chu2024sora},\\
                                                                --- Bias~\cite{schramowski2023safe},\\
                                                                --- Inappropriate Probability\\\cite{schramowski2023safe},\\
                                                                --- Expected Inappropriateness~\cite{schramowski2023safe},\\
                                                                --- COCO FID-$30$k Fidelity\\\cite{schramowski2023safe}
                                                                \vspace{8pt}
                                                                }  
                                                                       & \vspace{8pt}\makecell[{b{p{1.4in}<{\raggedright}}}]
                                                                                     {
                                                                                     --- LVLM-Based Scoring\\\cite{chu2024sora},\\
                                                                                     --- CLIP Score~\cite{schramowski2023safe}
                                                                                     \vspace{8pt}
                                                                                     }  
                                                                                            & \vspace{8pt}\makecell[{b{p{1.4in}<{\raggedright}}}]
                                                                                                          {
                                                                                                          --- Factual Fabrications\\\cite{lim2024addressing}
                                                                                                          \vspace{8pt}
                                                                                                          }  
                                                                                                                                & \\
\hline
\vspace{8pt}\makecell[{b{p{0.5in}}}]{P\vspace{8pt}}                
                & \vspace{8pt}\makecell[{b{p{1.5in}<{\raggedright}}}]
                              {
                              --- RoboEval~\cite{hu2024deploying},\\
                              --- SaGC~\cite{park2024clara},\\
                              --- Ravens~\cite{zeng2021transporter},\\
                              --- BabyAI~\cite{chevalier2018babyai},\\
                              --- TableSim~\cite{ren2023robots}
                              \vspace{8pt}
                              }
                             & \vspace{8pt}\makecell[{b{p{1.6in}<{\raggedright}}}]
                                           {
                                           --- Feasibility\\\cite{hazra2024saycanpay, ren2023robots, liang2024introspective, wang2024conformal, park2024clara, jha2023counterexample},\\
                                           --- Action Probabilities\\\cite{hazra2024saycanpay, ren2023robots, liang2024introspective},\\
                                           --- Plan/Action Accuracy$^\dagger$\\\cite{wang2024conformal, park2024clara, hu2024deploying},\\
                                           --- Action Variance~\cite{schramowski2023safe},\\
                                           --- Contradictions~\cite{jha2023counterexample}
                                           \vspace{8pt}
                                           }                
                                                  & \vspace{8pt}\makecell[{b{p{1.6in}<{\raggedright}}}]
                                                                {
                                                                --- Success Rate\\\cite{hazra2024saycanpay, ren2023robots, liang2024introspective, wang2024conformal, park2024clara, chen2023introspective, jha2023counterexample, hu2024deploying},\\
                                                                --- Generalizability\\\cite{hazra2024saycanpay, liang2024introspective, park2024clara, chen2023introspective},\\
                                                                --- Clarification/Unsure Rate\\\cite{ren2023robots, liang2024introspective, wang2024conformal, park2024clara},\\
                                                                --- \# of Attempts~\cite{chen2023introspective, jha2023counterexample, hu2024deploying},\\
                                                                --- Calibration Error\\\cite{ren2023robots, liang2024introspective},\\
                                                                --- Coverage~\cite{ren2023robots, liang2024introspective},\\
                                                                --- Estimated Payoff~\cite{hazra2024saycanpay, chen2023introspective},\\
                                                                --- Action Set Size~\cite{ren2023robots, liang2024introspective},\\
                                                                --- Plan Readability~\cite{wang2024conformal},\\
                                                                --- Plan Length~\cite{hazra2024saycanpay},\\
                                                                --- Unsafe Action Rate~\cite{liang2024introspective},\\
                                                                --- Monetary Cost~\cite{liang2024introspective},\\
                                                                --- Inference Speed~\cite{wang2024conformal}
                                                                \vspace{8pt}
                                                                }                
                                                                       & \vspace{8pt}\makecell[{b{p{1.4in}<{\raggedright}}}]
                                                                                     {
                                                                                     --- Action Probabilities\\\cite{hazra2024saycanpay, ren2023robots, liang2024introspective}
                                                                                     \vspace{8pt}
                                                                                     }                
                                                                                            & \vspace{8pt}\makecell[{b{p{1.4in}<{\raggedright}}}]
                                                                                                          {
                                                                                                          --- Non-Compliance Rate\\\cite{liang2024introspective},\\
                                                                                                          --- Preciseness~\cite{liang2024introspective}
                                                                                                          \vspace{8pt}
                                                                                                          }                
                                                                                                                                & \\
\bottomrule
\end{tabular}
}
\begin{tablenotes}
\small
\item $^\dagger$Encompasses machine metrics like (balanced) accuracy, precision, recall, F1, AUC, exact match~\cite{chen2017reading}, and pass@1~\cite{chen2021evaluating}.
\end{tablenotes}
\end{table*}

\paragraph{Select Intervention Algorithm Modality}
By this point, designers have performed the prerequisites of describing the available information from the model under test, its limitations, and defining relevant hallucination characteristics and metrics.
The remaining steps are actually developing the intervention algorithm. 
To do so, engineers need to choose which modality the proposed algorithm will fall under (\ie~white-, grey-, or black-box).
As described in Section~\ref{sec:detection}, possible modalities depends on the availability of access to necessary information from the model under test.
It is also important to consider at this stage the desired flexibility of the intervention algorithm (\ie~the ease of integrating the algorithm to different test models and evaluation settings).
For example, black-box methods are easier to deploy to proprietary models where access to weights and token probabilities is limited. 
However, if the designer only needs to intervene in the hallucination tendency of a particular open-source language model, white-box intervention algorithms can be tuned to the limitations of that specific model.

\paragraph{Intervention Type}
Next engineers can choose whether the intervention algorithm should detect and/or mitigate hallucinations.
Again, this choice depends on the intended deployment area of the model under test and the needs of the stakeholders in the model design process.
As seen in Table~\ref{tab:summary}, we find several existing algorithms enable both detection and mitigation of hallucinations.
Detection of hallucinations enables reacting to failures, while pure mitigation relies on the efficacy of the intervention algorithm alone to proactively filter hallucinations prior to final generation.
As such, we recommend safety-critical scenarios be deployed with detection and mitigation algorithms to simultaneously reduce chances of failures and inform impacted parties of potential hallucinations to increase transparency.

\begin{table}[t]
  \begin{center}
    \caption{\textbf{Benefits and limitations of each intervention algorithm type.}}
    \label{tab:pros_cons}
    \resizebox{0.77\textwidth}{!}{
    \begin{tabular}{lll} 
    \toprule
    \textbf{Method Type}      & \textbf{Pros} & \textbf{Cons} \\
      \midrule
      Hidden States\vspace{2pt}
                              & \makecell[{t{p{2.3in}<{\raggedright}}}]
                                {
                                --- tuned for specific model\\
                                --- hidden states hold useful embeddings\\
                                --- no need to re-embed text output
                                \vspace{2pt}
                                }
                                              & \makecell[{t{p{2.3in}<{\raggedright}}}]
                                                {
                                                --- reduced model transfer flexibility\\
                                                --- not applicable for proprietary models
                                                \vspace{2pt}
                                                } \\
      Attention Weights\vspace{2pt}       
                              & \makecell[{t{p{2.3in}<{\raggedright}}}]
                                {
                                --- tuned for specific model\\
                                --- attention weights hold useful info
                                \vspace{2pt}
                                }
                                              & \makecell[{t{p{2.3in}<{\raggedright}}}]{
                                                --- reduced model transfer flexibility\\
                                                --- not applicable for proprietary models
                                                \vspace{2pt}
                                                } \\
      Honesty Alignment       
                              & \makecell[{t{p{2.3in}<{\raggedright}}}]
                                {
                                --- directly fine-tunes model under test\\
                                --- empirically generalizes to new data
                                }
                                              & \makecell[{t{p{2.3in}<{\raggedright}}}]
                                                {
                                                --- tuned model still susceptible\\
                                                --- test efficacy impacted by data quality
                                                } \\
      \midrule
      Concept Probabilities\vspace{2pt}
                              & \makecell[{t{p{2.3in}<{\raggedright}}}]
                                {
                                --- generally model agnostic\\
                                --- intuitive\\
                                --- tried in broadest set of deployments
                                \vspace{2pt}
                                }
                                              & \makecell[{t{p{2.3in}<{\raggedright}}}]
                                                {
                                                --- requires access to token probabilities\\
                                                --- correlation may not necessarily hold
                                                \vspace{2pt}
                                                } \\
      Conformal Prediction
                              & \makecell[{t{p{2.3in}<{\raggedright}}}]
                                {
                                --- theoretical guarantees\\
                                --- applicable to multi-step planning\\
                                --- provides model uncertainty metric
                                }
                                              & \makecell[{t{p{2.3in}<{\raggedright}}}]
                                                {
                                                --- requires access to token probabilities\\
                                                --- requires collecting calibration dataset\\
                                                --- relies on human intervention
                                                } \\
      \midrule
      Analyzing Samples\vspace{2pt}
                              & \makecell[{t{p{2.3in}<{\raggedright}}}]
                                {
                                --- applicable to proprietary models\\
                                --- intuitive\\
                                --- removing conflicts reduces failures
                                \vspace{2pt}
                                }
                                              & \makecell[{t{p{2.3in}<{\raggedright}}}]
                                                {
                                                --- efficiency impacted by \# of samples\\
                                                --- affected by compliance metric choice
                                                \vspace{2pt}
                                                } \\
      Adversarial Prompting\vspace{2pt}
                              & \makecell[{t{p{2.3in}<{\raggedright}}}]
                                {
                                --- applicable to proprietary models\\
                                --- reveals undesired behaviors\\
                                --- can lead to better filters\\
                                --- red-teaming often occurs pre-release
                                \vspace{2pt}
                                }
                                              & \makecell[{t{p{2.3in}<{\raggedright}}}]
                                                {
                                                --- requires covering large input space\\
                                                --- generally does not mitigate\\
                                                --- requires hallucination classifier
                                                \vspace{2pt}
                                                } \\
      Proxy Model\vspace{2pt}
                              & \makecell[{t{p{2.3in}<{\raggedright}}}]
                                {
                                --- applicable to proprietary models\\
                                --- simple classifier could work well
                                \vspace{2pt}
                                }
                                              & \makecell[{t{p{2.3in}<{\raggedright}}}]
                                                {
                                                --- proxy has mismatched distribution\\
                                                --- dependent on proxy complexity\\
                                                --- LVLM proxy could hallucinate
                                                \vspace{2pt}
                                                } \\
      Grounding Knowledge\vspace{2pt}
                              & \makecell[{t{p{2.3in}<{\raggedright}}}]
                                {
                                --- applicable to proprietary models\\
                                --- provides evidence for responses\\
                                --- aligns model knowledge with users'\\
                                --- can help to learn policies faster
                                \vspace{2pt}
                                }
                                              & \makecell[{t{p{2.3in}<{\raggedright}}}]
                                                {
                                                --- requires ground-truth database\\
                                                --- could reference stale knowledge\\
                                                --- still fails in low-data regimes
                                                \vspace{2pt}
                                                } \\
      Constraint Satisfaction
                              & \makecell[{t{p{2.3in}<{\raggedright}}}]
                                {
                                --- applicable to proprietary models\\
                                --- theoretical guarantees\\
                                --- SMT solvers find failure cases quickly
                                }
                                              & \makecell[{t{p{2.3in}<{\raggedright}}}]
                                                {
                                                --- requires precise constraint definitions\\
                                                --- specification may be hard to parse
                                                } \\
      \bottomrule
    \end{tabular}
    }
  \end{center}
\end{table}

\paragraph{Identifying an Intervention Sub-Type}
Now that the deployment setting, modality, and intervention type have been identified, engineers can choose a method type from Table~\ref{tab:summary} that falls within those constraints.
This step will require some experimentation across method types and specific algorithms to identify the most effective intervention approach using the previously chosen metrics.
We list pros and cons of each method type in Table~\ref{tab:pros_cons} to assist researchers with narrowing down the scope of their algorithm search.

\paragraph{Implementing the Hallucination Intervention Algorithm}
Finally, designers can implement and integrate a chosen algorithm into the specific deployment setting and perform additional tests to measure its efficacy in detecting/mitigating hallucinations from the model under test.

\section{Future Directions}
\label{sec:future}

Here, we discuss some possible future directions in hallucination detection and mitigation techniques for foundation models to improve deployments to decision-making tasks.

\paragraph{Evaluating Methods on Decision-Making Tasks}
Most hallucination detection approaches are currently tested in offline QA settings for information retrieval or knowledge alignment, as seen in Table~\ref{tab:summary}.
As foundation models are increasingly used for more complex tasks, researchers should make an effort to adapt and evaluate earlier detection/mitigation approaches that were applied to QA problems.
Although dissimilar in practice from QA settings, planning and control problems may be formulated such that earlier mitigation methods can be evaluated on decision-making tasks.
For example, as discussed in Section~\ref{sec:found_ad},~\citet{chen2023driving} treat the autonomous driving task as a QA problem, which could be naturally extended to test other QA hallucination detection methods in the same setting.
This evaluation may lead to greater understanding of the general limitations of these models, as we draw parallels across diverse deployments.

\paragraph{Development of More Black-box Approaches}
White- and grey-box detection methods may not generally be applicable in situations where the internal state or token probabilities are unavailable from the language model.
Thus, we predict black-box approaches will take precedence in the near future, as state-of-the-art LVLMs like GPT-4V already prohibit access to probability outputs.
However, current black-box methods are limited with simplistic sampling techniques to gauge uncertainty, and proxy models may not be representative of the true state of the model under test.
Works like FLIRT showcase the promise of black-box adversarial prompting approaches in generating undesirable results~\cite{mehrabi2023flirt}.
We argue developing more aggressive black-box adversarial generative models, which explicitly optimize for producing inputs that may perturb the system outputs, is key to identifying the limits of a foundation model's knowledge. 

\paragraph{Pushing Models' Generalization Capabilities}
Currently, foundation models are primarily deployed to decision-making tasks that likely have some relation to its training set.
For example, although complex, tasks like multi-agent communication, autonomous driving, and code generation will be present in training datasets.
On the other hand, dynamic environments like robot crowd navigation require identifying nuances in pedestrian behaviors which the model may not have explicitly seen during training.
Thus, when models are deployed in decision-making contexts and encounter a previously unseen scenario, or they are utilized in a completely different setting from their training data, it is necessary to consider their generalization capabilities.
As discussed in Section~\ref{sec:found_ad}, existing examples of foundation models applied in autonomous driving and robotics utilize external tools to retrieve sensor data or memories before planning.
\citet{mialon2023augmented} refer to such models that extract useful information from databases as augmented language models.
As such, the authors argue that the combined efforts of using external tools and internal reasoning is critical to the generalizability of language models to broader tasks --- also explored by~\citet{chen2023introspective},~\citet{park2023generative}, and~\citet{wang2023voyager}.
From another perspective,~\citet{tong2024cambrian} approach the development of generalized multi-modal large language models from a vision-centric focus.
In particular, the authors find that training LVLMs with heavy consideration on the design of vision encoders and the respective connector between the vision and language models drastically improves the capabilities of architectures deployed in vision tasks (\eg~image captioning, QA, depth ordering).
This line of thinking can bolster the performance of LVLMs deployed in autonomous driving decision-making, where current methods have failed~\cite{wen2023road}.
Before integrating models into real-world applications, we argue that designers should thoroughly explore their generalization limitations to find directions for future growth and to maximize transparency of model capabilities.

\paragraph{Testing Multi-modal Models}
With the explosion of LVLMs, which allow for explicit grounding of natural language and vision modalities, further exploration should be performed in evaluating their effectiveness in decision-making systems.
\citet{wen2023road} take a step in the right direction towards testing black-box LVLMs in offline driving scenarios, but there is still work to be done in deploying these models in online settings.
This direction can shed light on the long-standing debate of whether modular or end-to-end systems should be preferred in a particular deployment setting.
In fact, while our work has focused on LVLMs, there exist other families of multi-modal foundation models for the audio~\cite{paab2023foundation} and $3$D generation~\cite{chen2024meshxl} spaces, which similarly hallucinate~\cite{wang2024hallod, sahoo2024comprehensive} and should be evaluated before deployment.

\paragraph{Summary} 
We provide a glimpse into the progress of research for evaluating hallucinations of foundation models for decision-making problems. 
First, we identify existing use cases of foundation models in decision-making applications (\eg~autonomous driving, robotics) and find several works make note of undesired hallucinated generations in practice.
By referencing works that encounter hallucinations across diverse domains, we provide a flexible definition for hallucinations that researchers can leverage, regardless of their deployment scenario.
Then, we give a taxonomy of existing hallucination detection and mitigation approaches for decision-making, question-answering,~\etc, alongside a list of commonly used metrics, datasets, and simulators for evaluation.
We find that existing methods range in varying assumptions of inputs and evaluation settings, and believe there is room for growth in general, black-box hallucination detection algorithms for foundation models.
Finally, we present generalized guidelines to assist engineers with selecting hallucination intervention algorithms across varied deployment contexts, and suggest future research directions.

\appendix

\section{Foundation Models Making Decisions}
\label{sec:app_foundation_models}

\subsection{Robotics}
\label{sec:app_robotics}

\begin{figure*}[h]
  \centering
  \includegraphics[width=0.9\linewidth]{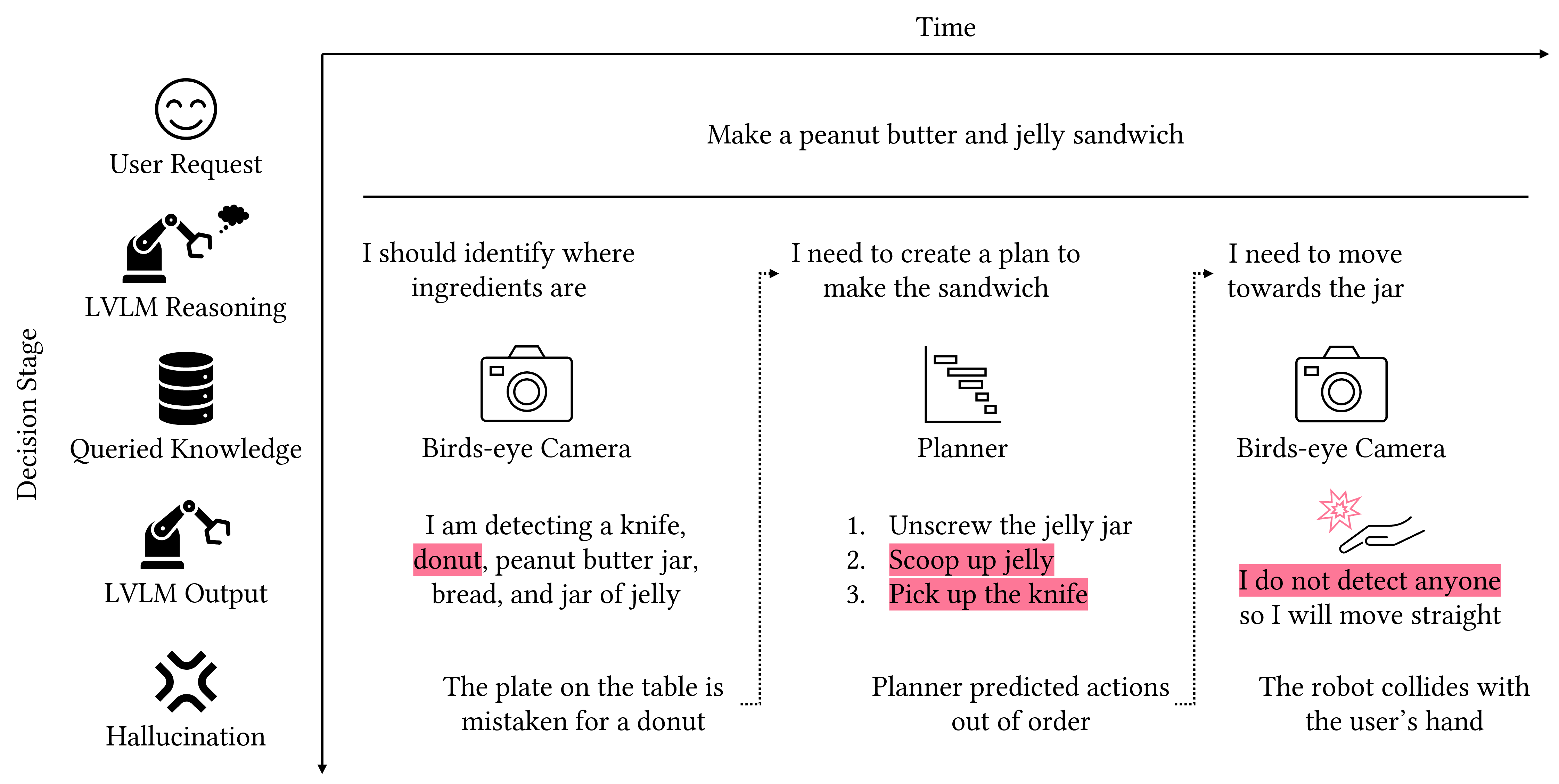}
  \caption{\textbf{Example deployment of an LVLM foundation model in a robotics setting.} Hallucinations are highlighted pink. Here, a robot tasked with assembling a sandwich initially identifies an object incorrectly. Then, the model comes up with an infeasible plan. Finally, when attempting to perform one of the actions, the robot collides with the human as it did not perceive any danger.}
  \label{fig:app_hall_robot_example}
\end{figure*}

\subsection{Other Areas}
\label{sec:app_other_areas}

There are also other works that apply foundation models for decision-making outside of the robotics and autonomous vehicle domains.
For example, ReAct from~\citet{yao2023react} identifies that a key limitation of chain-of-thought reasoning~\cite{wei2022chain} is that the model does not update its context or action based on observations from an environment.
As such, chain-of-thought reasoning relies purely on the internal reasoning of the foundation model itself to predict actions to take, missing a crucial step in grounding its actions with their effects on the environment.
Given a prompt, ReAct iterates between an internal reasoning step and acting in the environment to build up context relevant to the task.
\citeauthor{yao2023react} showcase the promise of the method in a QA setting where the LLM can take actions to query information from an external knowledge base, as well as an interactive text-based game, ALFWorld~\cite{shridhar2021alfworld}.
\citet{chen2023introspective} admit that ReAct is a powerful tool for dynamic reasoning and grounding, but is limited by the fact that the updated context from the Act step is only helpful for the particular task the model is currently deployed for.
They propose Introspective Tips to allow an LLM to reason about its past successes and failures in a world to generate general tips that will be helpful across diverse instruction-following tasks.
Specifically, tips are generated from the past experience of the model from a similar set of tasks, from expert demonstrations, and from several games that differ from the target task.
By summarizing these experiences into more concise tips,~\citeauthor{chen2023introspective} show that Introspective Tips outperform other methods in ALFWorld with both few- and zero-shot contexts.

\citet{park2023generative} and~\citet{wang2023voyager} apply foundation models in more complex environments to push models to their limits to simulate realistic human behaviors and test lifelong learning.
\citeauthor{park2023generative} propose generative agents that produce believable, human-like interactions and decisions within a small town sandbox environment.
They develop a module for individual agents in the simulation to store and retrieve memories, reflect about past and current experiences, and interact with other agents. 
Their generative agents use similar methods to ReAct and Introspective Tips to act based on a memory of experiences, but also interact and build relationships with other agents through dialogue.
The authors show that the agents are able to effectively spread information, recall what has been said to others and stay consistent in future dialogue interactions, and coordinate events together.
Sometimes, however, agents are found to hallucinate and \emph{embellish} their responses with irrelevant details that may be attributed to the training dataset of outside, real-world knowledge.
Voyager, from~\citeauthor{wang2023voyager}, deploys GPT-4 to the MineDojo environment~\cite{fan2022minedojo} to test its in-context lifelong learning capabilities.
The architecture prompts GPT-4 to generate next high-level tasks to complete, given the agent's current state and results of past tasks --- a form of automatic curriculum generation. 
Voyager then identifies what intermediate general skills would be required to complete the task, and the LLM is used to fill in a skill library with helpful low-level skills in the form of programs that call functions that are available to the simulator.
GPT-4 is prompted to generate skills that are generalizable to multiple tasks, so that the skill generation step does not have to be called for every task if the skill is already stored in the library.
\citeauthor{wang2023voyager} show that Voyager continuously learns to explore the diverse tech tree available within MineDojo while building and leveraging skills.
Even so, they find that the LLM hallucinates when generating tasks to tackle and when writing the code to execute for a particular skill, discussed further in Section~\ref{sec:hal_examples}.

\citet{kwon2023reward} explore the use of LLMs to act as a proxy for a hand-tuned reward function in RL tasks.
This application is particularly motivated by decision-making tasks that are difficult to specify with a reward function, but can be explained textually with preferences of how a policy should generally act. 
Specifically, the LLM evaluator first undergoes in-context learning with examples of how it should decide the reward in several cases of the task that the agent will be deployed to. 
Then, during RL training, the LLM is provided a prompt with the trajectory of the agent within the episode, the resulting state from the simulator, and the original task objective from the user, and is asked to generate a binary reward for the agent (1 if success, 0 else).
The binary reward is added to the experience replay, and the agent can be updated using any RL algorithm.
\citeauthor{kwon2023reward} find that a baseline in their work that predicts rewards with no in-context learning especially hallucinates with incoherent reasoning.
There has also been work by~\citet{suri2024large} on evaluating whether large language models elicit heuristics and biases when making decisions, like humans.
Specifically, while it is commonly believed in literature that biases like anchoring, representativeness, availability, framing, and endowment are brought about by cognitive processes,~\citeauthor{suri2024large} pose that if foundation models generate similar responses, these effects may be partly caused by language, since language models inherently have no cognitive capabilities.
Through four user studies comparing the responses of humans and ChatGPT to the same questions intending to bring about the decision biases, the authors find that both the model and participants showcased similar effects of partiality. 
As such, the authors posit that these results imply that decision heuristics in people may actually be influenced by linguistic syntax.
We further argue in Appendix~\ref{sec:app_broader_impacts} that language models deployed in decision-making contexts where they may impact humans' lives should aim to minimize these biases for fairness.

\section{Hallucinations}
\label{sec:app_hallucinations}

\subsection{Examples}
\label{sec:app_hal_examples}

\paragraph{Image and Video Generation}
Likewise, image and video generation models are not safe from hallucinations.
Recent generative models like DALL·E~\cite{ramesh2022hierarchical,betker2023improving}, Stable Diffusion~\cite{rombach2022high}, and SORA~\cite{openai2024video} have shown great promise in producing high quality frames given text input, but they can also hallucinate noncompliant, undesirable, and irrelevant features.
The DALL·E system cards~\cite{mishkin2022risks,dalle32024system} provide plenty of examples of such characteristics with biased, stereotypical generations, deepfakes of public figures, and violent and racy imagery.
\citet{betker2023improving}, who train DALL·E 3 to generate images given synthetic captions from a captioning model, find that poor, hallucinated training captions result in noncompliant image generations at test time that ignore important caption details.
Furthermore, generations may also include nonfactual contents that misguide users~\cite{lim2024addressing}.
Even with the development of safety filters that attempt to detect hallucinations before displaying the generation to the user,~\citet{rando2022redteaming} find that simple prompt engineering and red-teaming approaches can still lead to vulgar content, bypassing the filter.
As such, recent efforts have shifted from solely filtering inappropriate content at test time, and instead providing guidance to the model to assist in generating safer features~\cite{schramowski2023safe}.
However, instances of stereotypes are still prevalent in generations.
In a positive light,~\citet{huang2024visual} propose exploiting the hallucination tendency of image generation models to curate a benchmark to evaluate image captioning methods.
While the results of image generation models should minimize hallucinations in static frames, video generation models also need to produce temporally consistent and plausible clips across a sequence of frames.
Sora Detector, proposed by~\citet{chu2024sora}, attempts to detect static hallucinations (\eg~color distortion, deformations, unrealistic depth of field) and dynamic hallucinations (\eg~unnatural overlapping of objects, implausible motions, temporally illogical generations) within generated videos conditioned on text. 

\paragraph{$3$D Modeling and Generation}
Another field that is quickly gaining the attention of researchers is $3$D generation, where foundation models are generating high fidelity representations of objects or scenes conditioned on images or text~\cite{xia2023survey,liu2024comprehensive,bai2024progress,li2023generative}.
These foundation models span a wide range of downstream applications including human avatar creation~\cite{xu2023seeavatar}, medical tomography understanding~\cite{blankemeier2024merlin}, $3$D object generation~\cite{chen2024meshxl,han2025vfusion}, environment modeling~\cite{yang2024holodeck}, and more.
Although billions of labeled pairs of images and text exist across many datasets, as discussed in Appendix~\ref{sec:app_offline_datasets}, $3$D modality labels have only recently begun to approach a similar scale~\cite{yang20243d}, limiting the number of training pairs for $3$D foundation models.
As such, early $3$D generative foundation models conditioned on text relied on embeddings from text-image similarity models like CLIP~\cite{radford2021learning} to serve as a guide for generating accurate scenes~\cite{sanghi2022clip,jain2022zero}, or used pre-trained text-to-image diffusion models to provide feedback for training $3$D neural radiance field models~\cite{poole2023dreamfusion}.
Instead of optimizing individual low-level object representations directly, some works assume access to a collection of $3$D assets, which a standard LLM like GPT-$4$ can query to complete the floor plan of a $3$D scene, bypassing the need for low-level text-$3$D datasets~\cite{yang2024holodeck}.
~\citet{yang20243d} are some of the latest researchers to tackle the data scarcity problem of text-labeled $3$D scenes by introducing their own procedure for collecting text-scene labels across $40$K rooms.
When evaluating models with their new $3$D-POPE metric, the authors find that several $3$D LLMs hallucinate the presence of non-existent objects in the scene, similarly to other LVLMs tested on images.
~\citet{wang2024hallod} find that $3$D generative foundation models hallucinate spatial inconsistencies when rendering a generated scene from different perspectives.
Due to the scarcity of $3$D datasets, the authors propose Hallo$3$D: a three-part hallucination detection and mitigation approach leveraging LVLMs as advisors.
Specifically, Hallo$3$D samples multiple rendered views of a $3$D object with a pre-trained diffusion model conditioned on text.
Each rendering is passed through an LVLM which identifies abnormalities, and the predicted statements are fed into the diffusion model as negative contexts to update the original renderings through a \emph{prompt-enhanced reconsistency} step.
The authors use a cross-attention mechanism to improve the consistency of frames across different view points.

\subsection{Broader Impacts}
\label{sec:app_broader_impacts}

Thus far, we have primarily formulated a unified definition for hallucinations and provided examples of where they have come up in the research community.
However, as this new field advances rapidly every day, numerous industries have begun utilizing the technology in their respective applications.
As such, it is even more imperative that designers develop robust hallucination detection and mitigation methods before deploying models into areas with real humans at risk.
Three critical industries where LVLMs have shown great promise are the medical, legal, and finance sectors~\cite{li2023language, clusmann2023future, lai2023large}.
For example,~\citet{li2023language} and~\citet{zhao2024revolutionizing} explain that LLMs are currently used in finance for portfolio management, fraud detection, credit scoring, text summarization for forecasting, and customer-facing chatbots.
In fact, Bloomberg has already designed its own LLM, BloombergGPT, trained on a custom curated dataset for investor sentiment analysis given news transcripts, numerical reasoning QA, and named entity recognition~\cite{wu2023bloomberggpt}.
Similarly, LVLMs are being applied to medical tasks for patient education through QA, assisting physicians when writing reports and examining test results, and used in academia for taking medical exams --- even going so far as accomplishing passing results~\cite{omiye2024large}. 
Likewise,~\citet{lai2023large} showcase recent examples of use cases of LLMs in law, including summarizing legal documents, generating drafts of legal documents, acting as a legal consult, and making decisions given case facts.

For all of their promising results, foundation models have even greater impacts in these critical applications.
Specifically, designers need to consider problems of data privacy, training with out-of-date data, potential inconsistencies between generations and references, model bias, the ethics of using AI for a particular problem setting, and transparency to stakeholders and impacted parties in the decision-making pipeline.
In the finance sector,~\citet{kang2024deficiency} identify common failure modes of various open-source and proprietary language models in tasks for recognizing financial abbreviations and stock symbols, providing explanations for financial terms, and querying a stock price without access to an external database.
Each model is shown to have defects at inference time without using retrieval augmented generation (RAG)~\cite{lewis2020retrieval} methods (similar methods are discussed in Section~\ref{sec:grounding_know}), especially when the training data is out of date.
As such, companies like BlackRock, Inc. are directly applying RAG to existing LLMs for financial QA~\cite{sarmah2024towards}.
In the context of legal applications,~\citet{dahl2024large} provide a taxonomy of the complexities of different problems a language model could be tasked with, each with increasing risk of generating undesired hallucinations.
Notably, the frequency of hallucinations are found to increase with task complexity and varies with the specified court, jurisdiction, case prominence, and year. 
In fact, LLM hallucinations have already made their mark in a real court case, where a New York attorney utilized ChatGPT to generate a brief, which in turn referenced non-existent cases.
He states the model told him its generations were accurate, underscoring the importance of developing transparent models~\cite{weiser2023here}.
\citet{ali2024large} evaluate the efficacy of foundation models for automatically generating billing codes given medical procedure descriptions.
As the tested models currently perform miserably on the given task, the authors suggest further research before deploying the models in an automated medical context.

In an even more generalized medical setting, multiple LLMs are queried to answer novel questions, where even RAG results in generations that are either inconsistent with sources, or the cited text cannot be found in the source~\cite{wu2024well}. 
In fact, the AI search engine company Perplexity AI, which uses RAG to generate summaries, came under fire during the 2024 United States presidential election for releasing an election hub that generated hallucinated results~\cite{constantino2024ai}.
One human tester found that, while the company's product centralized information into one source, the model's generations wrote in varied tones for different candidates, resulting in biased generations.
Similarly, LLMs have been found to generate fake citations to support claims queried by users.
For example, an education official from the state of Alaska used a generative model to collect citations for a proposed policy~\cite{stremple2024false}.
The burden was then left to the policymaker to replace citations to nonexistent sources, leading journalists to question the lack of policy on using generative AI to write proposals which impact the public. 
The promise of LLMs have also led many educators to directly rely on these models to produce lesson plans, leading to hallucinations when left unchecked~\cite{barack2024using}.
One audio transcription tool leveraged in medical domains has been found to add irrelevant details to generations, and erase the ground-truth audio for security~\cite{swain2024patients}.
In an era where foundation models are being deployed in new, critical areas, we argue ground-truth data should be kept for later model evaluation and redundancies.
Overall, it is of the utmost importance to ensure LVLM generations are free of hallucinations in critical decision-making applications.
While these models are still far from reaching this milestone, it is even more important to be transparent about model capabilities to users to minimize misalignment of expectations.

\section{Detection and Mitigation Strategies}
\label{sec:app_detection}

\subsection{White-box Methods}
\label{sec:app_det_white}

\subsubsection{Hidden States}

\begin{enumerate}
\setcounter{enumi}{2}

\item \label{itm:app_luna}
LUNA, introduced by~\citet{song2023luna}, is a general framework that measures the trustworthiness of an LLM output containing four stages of evaluation: model construction, semantic binding, quality metrics, and practical application.
The abstract model construction phase attempts to profile the LLM using its hidden states with either a discrete time Markov chain (DTMC) or a hidden Markov model (HMM) architecture.
For example, when fitting a DTMC model, the authors encode the hidden states of the language model into a lower dimensional space, cluster them into abstract discrete states, and learn a transition function between said states.
Semantic binding is used alongside quality metrics to identify the states and transitions that are trustworthy, and which ones are undesired.
Finally, at inference time, as the model generates output tokens to a given prompt, the intermediate network layer embeddings are iteratively passed through the profiling model to identify when undesired transitions occur.
The authors evaluate their framework's capability of detecting hallucinations within QA datasets.

\end{enumerate}

\setcounter{subsubsection}{2}
\subsubsection{Honesty Alignment}

\begin{enumerate}
\setcounter{enumi}{1}

\item \label{itm:app_yang_hon}
\citet{yang2023alignment} take the method one step further by also training the model to refuse to answer questions with high uncertainty. 

\end{enumerate}

\subsection{Grey-box Methods}
\label{sec:app_det_grey}

\setcounter{subsubsection}{1}
\subsubsection{Conformal Prediction}

\begin{enumerate}
\setcounter{enumi}{1}

\item \label{itm:app_kumar}
\citet{kumar2023conformal} similarly apply conformal prediction to LLMs, but for answering multiple choice questions. 
Specifically, the method first collects a calibration dataset of prompts and the normalized token probabilities of the correct token (\ie~A, B, C, or D) being chosen from the model.
Then, during deployment, given a user-defined error rate and a prompt, their algorithm chooses the multiple choice answers with token probabilities that fall within the calibrated score on the held-out dataset.  

\end{enumerate}

\subsection{Black-box Methods}
\label{sec:app_det_black_box}

\subsubsection{Analyzing Samples from Model}
\label{sec:app_det_analyze}

\begin{enumerate}
\setcounter{enumi}{1}

\item \label{itm:app_halo}
A concurrent work from~\citet{elaraby2023halo} rather computes the \emph{entailment} among responses at the sentence-level.
Consistency metrics check whether responses contradict one another while entailment metrics identify if the responses imply one another.
The nuanced difference between SelfCheck-NLI and their method, HaloCheck, is that~\citeauthor{elaraby2023halo} use the SummaC~\cite{laban2022summac} entailment estimation method, placing equal weightage among all sentences, in all responses, to compute a more balanced prediction score. 
The authors evaluate HaloCheck and hallucination mitigation techniques on a domain-specific, custom-curated dataset, with facts about the US National Basketball Association (NBA).
Specifically,~\citeauthor{elaraby2023halo} first prompt GPT-4 for questions on topics within the NBA domain and manually filter out low-quality generations, resulting in $151$ questions.
An LLM is then queried for $5$ responses to each question, and each response is manually annotated for consistency and accuracy of generations.
HaloCheck is also shown to be more efficient at predicting scores.
Other commonly used metrics within the language community for similarity estimation are discussed in Appendix~\ref{sec:app_eval_metrics_lang}.

\setcounter{enumi}{6}
\item \label{itm:app_pope}
In the problem space of image captioning,~\citet{li2023evaluating} estimate the accuracy of a (possibly hallucinating) LVLM when describing an image with a text caption.
Early metrics, like CHAIR~\cite{rohrbach2018object}, fail to provide stable estimates of accuracy when different captions with similar semantic grounding lead to varying scores.
To tackle this stability problem, the authors propose POPE, which curates binary questions about whether an object exists within an image scene.
Questions to which the foundation model provides conflicting responses describe objects that the model may be hallucinating.
These metrics are further described in Appendix~\ref{sec:app_eval_metrics_obj}.

\item \label{itm:app_verbal}
Yet another recent work by~\citet{xiong2024can} asks LLMs in a zero-shot manner to verbally include their uncertainty in the generated output.
This desired behavior is elicited through additional prompt engineering (\ie~add a phrase to the prompt like ``Please provide your confidence level as a percentage'').
Unlike~\citet{lin2022teaching}, the authors do not further fine-tune the model to a calibration dataset of uncertainties. 
To combat over-confidence in output scores,~\citeauthor{xiong2024can} utilize chain-of-thought reasoning, predict the confidence score of each sub-claim, and combine them over the whole response to compute the final belief. 
The authors find that a hybrid approach, merging verbalized uncertainty with self-contradiction detection, outperforms the individual components alone on expected calibration error, comparing predicted confidence and actual model accuracy.

\item \label{itm:app_interrogate}
\citet{yehuda2024interrogatellm} present InterrogateLLM, a sampling-based approach that attempts to reconstruct an original query given a possibly hallucinated LLM response, and measures the similarity of the two queries.
Inspired by human studies~\cite{brewer1999beliefs}, the authors specifically hypothesize that responses that generate queries that differ greatly from the original prompt point to possible hallucinations.
However, they also confess that, like human studies, there is the possibility of false positive hallucination detections due to the stochastic nature of language models.
InterrogateLLM follows a simple procedure: form a prompt with few-shot examples containing queries and answers followed lastly by the actual query, generate a (hallucinated) response from a language model under test, reverse the original prompt examples to provide answers and their queries with the generated response appended last, sample generated queries from any language model (not necessarily the one under test), and measure the cosine similarity of vector embeddings between the original query and generated queries.
Using public datasets covering a range of trivia knowledge~\cite{kaggle2017movies,kaggle2017books,kaggle2017gci}, the authors curate a dataset of QA pairs.
Most importantly, the authors find that their backward few-shot method is critical to decreasing the false-negative detection rate of hallucinations seen in SelfCheckGPT~\cite{manakul2023self}, which only compares generated responses from the original forward pass.
Furthermore, using an ensemble of models for the backward generation and more iterations of query sample generation lead to higher detection accuracy, at the cost of efficiency.
We expect this backward-query generation approach to hallucination detection can also be applied to planning tasks, where hallucinated plans will reverse-generate task descriptions that do not match well with the original goal.

\end{enumerate}

\subsubsection{Adversarial Prompting}
\label{sec:app_det_black_adv}

\begin{enumerate}
\setcounter{enumi}{2}

\item \label{itm:app_invite}
Motivated by similar findings that language models are overfitting to QA datasets,~\citet{ramakrishna2023invite} present a novel framework for generating invalid questions to test the frequency of hallucinations output by new LLMs.
The authors collect an augmented version of the DBpedia dataset~\cite{lehmann2015dbpedia} replaced with invalid questions to evaluate the hallucination rate of language models, and note that any deployment dataset could have been used instead.
Specifically,~\citeauthor{ramakrishna2023invite} manually create a list of $24$ question templates with tags for subjects and objects that can be filled in.
A set of $100$ invalid questions using the templates is generated by sampling subjects and objects from disjoint facts in the original dataset, and ensuring the questions do not have valid answers in DBpedia.
Additional invalid questions are produced by replacing dates within questions from TriviaQA~\cite{joshi2017triviaqa} with ones that do not exist.
By passing each test question through various open-source and proprietary LLMs, the authors manually validate that each model has a tendency to hallucinate.
Unfortunately,~\citeauthor{ramakrishna2023invite} find that using automated evaluation metrics (discussed in Appendix~\ref{sec:app_eval_metrics}) do not correlate well with human annotations on their generated dataset, leaving room for future growth in aligned automatic evaluation metrics.

\item \label{itm:app_hypoterm}
A more recent adversarial method from~\citet{uluoglakci2024hypotermqa} attempts to automatically generate hypothetical, invalid questions that language models should reject answering.
For example, the hypothetical question, ``What are the differences between Platypus LLM and Wolf LLM?'' should be rejected since Wolf LLM is nonexistent, even though Platypus~\cite{lee2023platypus} is a real family of language models.
Intuitively, if a model provides a plausible answer to the invalid question, the authors claim that either the fabricated term is not be in the model's training set or the model has a higher tendency to hallucinate overall.
As such, the authors propose a new framework to generate hypothetical questions, with which they create the HypoTermQA dataset.
In particular,~\citeauthor{uluoglakci2024hypotermqa} first query GPT-3.5 for $20$ popular topics and then $50$ hypothetical terms per topic, resulting in $790$ filtered fake phrases.
To generate a diverse set of outputs, the model is set with a high temperature parameter.
The authors argue that questions produced with only hypothetical terms will be easier to distinguish by language models, and thus, generate a set of similar terms with real meanings, in an attempt to deceive models.
Similar, valid phrases were generated using three distinct approaches: querying GPT-3.5 directly, retrieving titles from Wikipedia with similar vector embeddings to the hypothetical terms, and taking the titles of Wikipedia passages whose definition embeddings are similar to those of the hypothetical terms.
Finally, GPT-3.5 is instructed to generate plausible questions with filtered hypothetical and valid terms to produce a total of $19.5$K hypothetical and valid questions.
Rather than relying on human annotation for identifying hallucinated responses,~\citeauthor{uluoglakci2024hypotermqa} additionally propose HypoTermQA Score --- the ratio of valid answers to the total number of hypothetical questions, automatically labeled by a proxy evaluator LLM agent (like methods in Section~\ref{sec:proxy_methods}).
In evaluation, the authors find that both open-source and proprietary models have over $90\%$ frequency of generating invalid responses.
However, this frequency differed depending on the chosen evalautor model due to biases.
Thus, proxy evaluator agents are also tested by comparing against manual annotation.
Overall, the proposed dataset generation and automatic hallucination detection method show great promise in evaluating the factual hallucination tendency of language models.
But, additional work should be done to handle biases in evaluator models, evaluate other characteristics of hallucinations, and consider other deployments outside QA.

\end{enumerate}

\subsubsection{Proxy Model}
\label{sec:app_proxy_methods}

\begin{enumerate}
\setcounter{enumi}{2}

\item \label{itm:app_pacchiardi}
Similarly,~\citet{pacchiardi2024how} develop a black-box lie detector for LLMs.
In their case, the authors hypothesize that models that output a lie will produce different behaviors in future responses, like~\citet{azaria2023internal}.
As such, at inference time,~\citeauthor{pacchiardi2024how} prompt the LLM with several binary questions (that may be completely unrelated to the original response) and collect yes/no answers.
All the responses are concatenated into a single embedding that is input to the logistic regression model to predict the likelihood that the response was untruthful.
To evaluate their lie detector, the authors assemble over $20$K questions from existing data sources on topics including general trivia~\cite{vrandevcic2014wikidata,welbl2017crowdsourcing,meng2024locating}, basic arithmetic~\cite{patel2021nlp}, common sense reasoning~\cite{talmor2021commonsenseqa}, text translation~\cite{tiedemann2012parallel}, and self-awareness~\cite{perez2023discovering}.
They additionally synthetically generate questions with unknowable answers as a control split (\eg~``What day is it?''). 
The authors find that the simple detector is mostly task- and model-agnostic once trained on a single dataset.

\item \label{itm:app_soradetector}
\citet{chu2024sora} primarily rely on the LVLM GPT-4 to detect hallucinations within AI-generated videos.
Video hallucinations are classified as static hallucinations, which occur in individual frames, or dynamic hallucinations, which occur across multiple frames.
Here, the authors employ the generalized capabilities of GPT-4 to detect objects within keyframes of a video, summarize the video from the keyframes (which might be inconsistent with the original video generation prompt due to hallucinations), synthesize a temporal knowledge graph representing the changing relations among detected objects, detect inconsistencies along the knowledge graph, aggregate a hallucination score, and describe the detected hallucinations.
Sora Detector is evaluated on a custom dataset collected by the authors, T2VHaluBench (described in Appendix~\ref{sec:app_eval_img_gen}), and outperforms video hallucination detection ablation approaches that ignore knowledge graph construction, showcasing the usefulness of the representation.
As such the structured approach shows promise in identifying undesired effects in generated videos, but the authors do not provide discussion on the possible hallucinations generated by GPT-4 when detecting and describing failures.

\end{enumerate}

\subsubsection{Grounding Knowledge}
\label{sec:app_grounding_know}

\begin{enumerate}
\setcounter{enumi}{1}

\item \label{itm:app_cok}
Some knowledge grounding approaches prompt LLMs to generate code to directly query information from databases.
\citet{li2024chain} are motivated by the limitations of existing knowledge-based hallucination mitigation methods; namely that (1) they utilize a fixed knowledge source for all questions, (2) generating retrieval questions with LLMs that interface with a database is not effective because they may not be trained on the particular programming language of the database, and (3) there is no correction capability that handles error propagation between knowledge modules.
Consequently, the authors propose augmenting LLMs with heterogeneous knowledge sources to assist with summary generation.
Specifically, in the event that the model is found to be uncertain about its generated statement through self-contradiction, their framework, chain-of-knowledge (CoK), chooses subsets of knowledge-bases that may be helpful for answering the original question. 
Assuming each database has its own query generator, CoK queries for evidence, and corrects rationales between different sources iteratively.
Compared to chain-of-thought reasoning, CoK consistently produces more accurate answers with its iterative corrections.

\item \label{itm:app_mixalign}
Another source of potential conflict that leads to hallucinations, is misalignment between a model's capabilities and the user's beliefs about what it can do.
\citet{zhang2023knowledge} tackle this knowledge alignment problem and categorize alignment failures into four types:
\begin{itemize}[noitemsep]
  \item Semantic --- an ambiguous term maps to multiple items in a database
  \item Contextual --- the user failing to explicitly provide constraints
  \item Structural --- user provides constraints that are not feasible in the database
  \item Logical --- complex questions that require multiple queries
\end{itemize}
\noindent
Their proposed MixAlign framework interacts with the user to get clarification when the LLM is uncertain about its mapping from the user query to the database.
With the original query, knowledge-base evidence, and user clarifications, the LLM formats its final answer to the user.

\setcounter{enumi}{5}
\item \label{itm:app_factual_img_retrieval}
\citet{lim2024addressing} tackle the issue of image generation models producing results that are factually noncompliant with the original prompt, by utilizing an external database to address inconsistencies. 
In particular, the authors confront three forms of image hallucinations: factual inconsistencies, outdated knowledge, and factual fabrications (unlikely generations). 
Given a prompt and an original (hallucinated) generation from a generator, the authors manually collect a ground-truth image describing the prompt using Google. 
An LVLM is employed to identify the differences between the retrieved and generated images, resulting in an instruction for how to edit the original generation to become factual.
Finally, the original generation and editing instruction is passed through the generator to produce an updated result.
Qualitatively, the pipeline produces more factually compliant and plausible generations.
However, a human needs to manually choose the ground-truth image, the prompt correction generation model can hallucinate improper instructions, and the authors do not compare their method with other image hallucination mitigation methods.

\item \label{itm:app_sld}
\citet{schramowski2023safe} present Safe Latent Diffusion (SLD), a method to provide additional guidance to image generation diffusion models at inference time to reduce the tendency of inappropriate generations.
Intuitively, diffusion models iteratively remove noise from a generation until a plausible result is produced. 
Along the way however, inappropriate features like nudity or violence can be denoised, as seen in the training dataset.
SLD updates the original prompt guidance encoding by computing the latent encoding of unsafe guidance, and shifting the original encoding away from the unsafe encoding given set hyperparameters.
The resulting shifted ``safe'' guidance is far from unsafe embeddings, but still close enough to the original prompt guidance, to produce generations that attempt to follow the original prompt without inappropriate content.
This augmented guidance is only provided after some warm-up steps of denoising to allow for the model to generate results close to the original prompt.
The authors provide four sets of hyperparameters to configure the aggressiveness of the content filter.
Using a model that detects inappropriate content,~\citeauthor{schramowski2023safe} show that SLD has a lower probability of generating undesired features, and a user study supports that the safe generations are still preferred the same as (or more than) the unaugmented results.
It is important to note that, while SLD combats undesired generations, there are still instances of inappropriate features present due to their representation in the training dataset.
Furthermore, the proposed algorithm can be misused by shifting the augmented guidance toward the unsafe vector, producing more inappropriate features.

\end{enumerate}

\section{Metrics and Evaluation Platforms}
\label{sec:app_evaluation}

We now present common metrics, datasets, and simulation platforms leveraged when developing and evaluating the hallucination detection algorithms introduced in Section~\ref{sec:detection}.

\subsection{Metrics}
\label{sec:app_eval_metrics}

Here, we list established metrics used for computing language similarity and accuracy of generated image descriptions.

\subsubsection{Language Similarity}
\label{sec:app_eval_metrics_lang}

\paragraph{BERTScore~\cite{zhang2020bertscore}} 
Given a pair of responses, BERTScore computes the BERT~\cite{devlin2019bert} embeddings of the sentences and calculates their cosine similarity.

\paragraph{BARTScore~\cite{yuan2021bartscore}} 
Using a pre-trained BART model~\cite{lewis2020bart}, which provides access to generated token probabilities, BARTScore sums over the log probability of each token generated while conditioning on context and previously output tokens.
Essentially, BARTScore attempts to predict the quality of a generated text using BART as a proxy model.

\paragraph{SummaC~\cite{laban2022summac}}
SummaC is a class of natural language inference models that predict entailment, contradiction, and neutral scores between pairs of sentences among a document and its summary. 
Each score is collected into a separate matrix split by metric type.
The authors propose two approaches, SummaC\textsubscript{ZS} and SummaC\textsubscript{Conv}, for aggregating scores of each sentence in the summary with respect to each sentence in the document.

\paragraph{GPTScore~\cite{fu2023gptscore}}
Like BARTScore, GPTScore relies on a pre-trained language model with access to token probabilities to estimate quality of outputs, but uses the GPT series of LLMs.

\paragraph{AlignScore~\cite{zha2023alignscore}}
The creators of AlignScore pose that two pieces of text are \emph{aligned} when all information present in one text exists in the other, and the texts do not contradict one another. 
Consequently, they train a classification model on labeled data with three types of labels: a binary classification of aligned or not, a multi-class prediction including a neutral label in addition to the binary classification labels, and a continuous score for a regression task.
The AlignScore metric computes a weighted score across all three prediction heads at test-time.

\paragraph{Semantic Uncertainty~\cite{kuhn2023semantic}}
One common method of measuring uncertainty of a model's many generations is computing its entropy over all generated token probabilities. 
However, in cases where multiple sentences have the same semantic meaning but output different entropies, the aggregated measurement is not representative of the true uncertainty of the model.
\citeauthor{kuhn2023semantic} tackle this problem by clustering sentences into semantic classes and summing entropies of sentences from the same class together.

\subsubsection{Image Captioning}
\label{sec:app_eval_metrics_obj}

\paragraph{CHAIR~\cite{rohrbach2018object}}
CHAIR\textsubscript{$i$}, used for measuring accuracy of descriptions of images, is the ratio of the number of hallucinated objects to all the objects mentioned in the description.
To identify the hallucinated objects within the description, the authors assume access to ground-truth object classes in the image.

\paragraph{POPE~\cite{li2023evaluating}}
\citeauthor{li2023evaluating} recognize that different instructions prompting for a description of an image may lead to different responses from the model with the same semantic meaning.
In this case, CHAIR gives different scores to both descriptions although they are alike.
Instead, their proposed metric, POPE, asks binary questions about the existence of in-domain and out-of-domain objects in the image, which leads to more a more stable metric across different outputs.

\subsection{Offline Datasets}
\label{sec:app_offline_datasets}

In this section, we present relevant offline datasets used for evaluating the performance of hallucination detection and mitigation techniques in driving, robotic, and QA tasks.

\subsubsection{Driving}
\label{sec:app_offline_data_driving}

\paragraph{BDD-X~\cite{kim2018textual}}
BDD-X is a multi-modal driving dataset consisting of $20$K samples (\ie~video clips), each consisting of eight images with vehicle control actions and text annotations describing the scene and justifying actions.

\paragraph{DriveGPT4~\cite{xu2024drivegpt4}}
\citeauthor{xu2024drivegpt4} augment BDD-X into a QA dataset consisting of questions that ask about the current action of the vehicle, reasoning behind the action, and predicting future control signals. 
To incorporate other questions a user might ask about the vehicle, surroundings, and other miscellaneous queries, they prompt ChatGPT to generate further questions.
In total, the DriveGPT4 dataset contains $56$K samples.

\paragraph{nuScenes~\cite{caeser2020nuscenes}}
The nuScenes dataset contains $1$K driving videos, each running for $20$ seconds, collected from roads in Boston and Singapore. 
Each frame includes six different RGB camera views, GPS, annotated $3$D bounding boxes of various object classes, and semantically labeled rader, lidar, and map representations.

\paragraph{NuScenes-QA~\cite{qian2024nuscenesqa}}
Like DriveGPT4, NuScenes-QA is a visual QA dataset, but built on top of nuScenes.
It includes five different types of questions including checking the existence of objects, counting instances, detecting the object being referred to, identifying the action state of an object, and comparing two objects.
Overall, the dataset holds $450$K QA pairs across $34$K scenes in nuScenes.

\paragraph{Talk2Car~\cite{deruyttere2019talk2car}}
Talk2Car is an earlier extension of the nuScenes dataset which aims to ignite further research into developing systems that bridge the gap between passengers and an autonomous vehicle through natural language.
Annotators provided approximately $12$K text commands over $850$ videos within the nuScenes training split which refer to an object in the scene.

\paragraph{Refer-KITTI~\cite{wu2023referring}}
While Talk2Car is a pioneering work for object referral in real driving scenes through natural language, each annotated instruction only refers to one object.
As such,~\citeauthor{wu2023referring} propose a new task definition, referring multi-object tracking (RMOT), which attempts to predict all objects that are referred to within a natural language input.
They augment the KITTI driving dataset~\cite{geiger2012kitti} with labeled $2$D bounding boxes around objects that are referenced within a text prompt for $6.5$K images.

\paragraph{NuPrompt~\cite{wu2023language}}
NuPrompt is another RMOT-based benchmark, but applied to nuScenes and with $3$D bounding box labels.
It includes $35$K languages prompts, with most prompts referring to anywhere between one and ten objects.

\paragraph{DRAMA~\cite{malla2023drama}}
\citeauthor{malla2023drama} argue that, while several datasets exist for anomaly detection or identification on roads, there is a gap in explaining the reason for categorizing an object as being \emph{risky}, \ie~objects the model should pay attention to, like crosswalks, pedestrians, and traffic lights.
As such, DRAMA is a benchmark tackling identification of risky objects in a driving scene conditioned on natural language.
\citet{ding2023hilmd} extend DRAMA to further include suggestions on actions the ego vehicle can take to minimize risk, but the dataset is not public at this time.

\paragraph{NuInstruct~\cite{ding2024holistic}}
NuInstruct addresses two common limitations in existing driving datasets: they cover a limited subset of necessary tasks while driving (\eg~evaluating perception while ignoring planning), and disregard temporal and multi-view representations.
Built on top of NuScenes, the dataset provides $91$K samples of multi-view sequences with corresponding QA pairs spanning $17$ subtasks within perception, prediction, planning, and risk detection.

\paragraph{DriveLM~\cite{sima2023drivelm}}
The authors of DriveLM curate a similar comprehensive dataset from nuScenes and the CARLA driving simulator~\cite{dosovitskiy2017carla} with open-ended and factual questions about importance rankings of nearby vehicles, planning actions, detecting lanes, and more.

\paragraph{Driving with LLMs~\cite{chen2023driving}}
\citeauthor{chen2023driving} collect a text-based QA dataset from a proprietary driving simulator, generated from ChatGPT with ground-truth observations (\eg~relative locations of detected vehicles, ego vehicle control actions,~\etc) from the simulator.

\subsubsection{Code Generation and Robotics}
\label{sec:app_eval_code_gen}

\paragraph{HumanEval~\cite{chen2021evaluating}}
HumanEval is a set of $164$ handwritten programs, each with a function definition, docstring, program body, and unit tests.
The authors find there is great promise in using LLMs for code generation, but output quality is limited by length of context and buggy examples.

\paragraph{RoboCodeGen~\cite{liang2023code}}
\citeauthor{liang2023code} build a new code generation benchmark specifically for robot tasks with $37$ functions focused on spatial reasoning, geometric reasoning, and controls.

\paragraph{Language-Table~\cite{lynch2023interactive}}
The Language-Table dataset contains $594$K trajectories manually annotated with $198$K unique instructions across simulated and real-world manipulator robots.
The multi-modal dataset consists of video sequences, corresponding actions at each time step, and language instructions describing the policy of the robot in hindsight.

\paragraph{SaGC~\cite{park2024clara}}
The authors of the CLARA method developed a dataset to identify language goals from a user that are certain, ambiguous, and infeasible. 
Collected from three different types of robots (cooking, cleaning, and massage), SaGC is annotated with a floor-plan, descriptions of objects and people in view, a text goal, and a label of uncertainty.

\subsubsection{Question-answering}

\paragraph{HotPotQA~\cite{yang2018hotpotqa}}
HotPotQA is a question-answering benchmark with $113$K multi-hop questions (\ie~requiring multiple steps of reasoning to reach answer) collected from Wikipedia.
The dataset includes both questions that require finding relevant phrases from context paragraphs, and comparing two entities.

\paragraph{FEVER~\cite{thorne2018fever}}
In contrast to HotPotQA, the developers of FEVER attempt to answer the question of whether a fact is supported by a knowledge-base.
The database contains $185$K claims with annotated labels deciding if each claim is supported, refuted, or indeterminable from Wikipedia articles.

\paragraph{Natural Questions~\cite{kwiatkowski2019natural}}
Natural Questions is yet another QA dataset with sources from Wikipedia.
The authors release $307$K training and $7$K test samples of real (anonymized) queries into the Google search engine paired with a Wikipedia page and a long and short answer annotated by a person based on said article.

\paragraph{StrategyQA~\cite{geva2021aristotle}}
Like HotPotQA, StrategyQA aims to develop a dataset of implicit multi-hop questions, but includes a greater variety categories of questions, and with less category imbalance.
Furthermore, most of the questions in the dataset require three or more steps of decomposition and referencing to accurately solve.

\paragraph{QreCC~\cite{anantha2021open}}
Separate from the information retrieval task described in benchmarks above,~\citeauthor{anantha2021open} develop a dataset, QreCC, for conversational QA.
They focus on reading comprehension, passage retrieval, and question rewriting tasks, with a total of $13.7$K dialogues paired with $81$K questions.

\paragraph{HA-DPO~\cite{zhao2024hallucinations}}
\citeauthor{zhao2024hallucinations} present a multi-model visual QA dataset of images, hallucinated descriptions, and non-hallucinated samples from the VG dataset~\cite{krishna2017visual}.

\subsubsection{Image and Video Generation}
\label{sec:app_eval_img_gen}

\paragraph{AVA~\cite{murray2012ava}}
\citeauthor{murray2012ava} tackle curating a dataset to evaluate the aesthetic value of varying quality images. 
AVA contains $250$K images with aesthetic ratings, textual semantic class labels, and pictographic style labels.
The majority of images contain at least one semantic label, with $150$K containing at least two labels.

\paragraph{MSCOCO~\cite{lin2014mscoco}}
The MSCOCO dataset contains $328$K images with $2.5$M human-labeled segmentations of common objects.
Each image is accompanied by five text captions.
In contrast to earlier datasets like ImageNet~\cite{deng2009imagenet} and SUN~\cite{xiao2010sun}, MSCOCO provides more instance labels per class and more non-iconic images containing multiple objects per scene.

\paragraph{LSUN~\cite{yu2015lsun}}
As vision models continue to evolve and essentially \emph{solve} existing benchmarks, new datasets need to be curated to provide additional challenges and evaluate generalization performance.
Thus,~\citeauthor{yu2015lsun} propose a human-in-the-loop data labeling scheme to gather a dataset with one million images for ten scene and $20$ object classes, by iterating between manual labeling, overfitting a classifier, and automatic labeling.
Note that the overall precision of labels is slightly lower than pure manual labeling approaches.

\paragraph{LAION-$400$M~\cite{schuhmann2021laion}}
More recently, companies are collecting proprietary datasets to train large vision models, outperforming open-sourced models trained on smaller scale datasets.
To narrow the gap in training data for generative models,~\citeauthor{schuhmann2021laion} present an image dataset with $400$M images with corresponding metadata, CLIP~\cite{radford2021learning} embeddings, and web crawling resources.
Unfortunately, works have found that the training split of LAION-$400$M contains undesired content like racy imagery, slurs, stereotypes, and other biases, which should be filtered out~\cite{saharia2024photorealistic}.

\paragraph{I2P~\cite{schramowski2023safe}}
The Inappropriate Image Prompts dataset attempts to measure the likelihood of image generation models to output undesired, inappropriate content given a prompt.
In particular, the dataset contains $4.5$K prompts taken from real users of Stable Diffusion, which have been filtered to contain inappropriate details like hate speech, violent imagery, and criminal behavior.

\paragraph{DrawBench~\cite{saharia2024photorealistic}}
The DrawBench benchmark contains $200$ prompts across $11$ categories to evaluate the quality of generated images on a spectrum of properties (\eg~colors, quantities, generated text).
\citeauthor{saharia2024photorealistic} use their benchmark to present $8$ samples from two different models to human raters, who choose which model they prefer qualitatively. 

\paragraph{T2VHaluBench~\cite{chu2024sora}}
The developers of the Sora Detector curate a benchmark dataset for text to video hallucination detection.
In total, the dataset contains $59$, $8$-second-long videos --- $53$ of which contain consistency, static, or dynamic hallucinations generated by the Runway-Gen-2~\cite{runway2023gen2} and SORA~\cite{openai2024video} generative models.

\subsection{Simulation Platforms}

Finally, we introduce common online simulators used to test hallucination detection methods for decision-making tasks.

\subsubsection{Driving}

\paragraph{HighwayEnv~\cite{leurent2018highway}}
\citeauthor{leurent2018highway} presents a $2$D car simulator, with driving scenarios ranging from a passing on a multi-lane highway, merging into a highway, merging and exiting from a roundabout, parking, and more.
An ego vehicle can be controlled with discrete (\eg~merge left, merge right, faster, \etc) or continuous (\eg~providing an explicit acceleration command) actions.

\paragraph{SUMO~\cite{lopez2018sumo}}
Geared towards microscopic traffic simulation, SUMO allows researchers to design road networks, track traffic flow metrics, and control individual vehicles.

\paragraph{CARLA~\cite{dosovitskiy2017carla}}
CARLA is a $3$D driving simulator built on top of Unreal Engine.
Existing works benchmark their methods on CARLA for perception, planning, control, and QA tasks for its realism. 
There is also capability to perform co-simulation with SUMO and CARLA simultaneously~\cite{wegener2008traci}.

\subsubsection{Robotics}

\paragraph{Ravens~\cite{zeng2021transporter}}
Ravens is a $3$D manipulator robot (UR5e) simulator built with PyBullet~\cite{coumans2016pybullet} with tasks like block insertion, towers of hanoi, aligning boxes, assembling kits,~\etc
~Each simulated task features a manipulator robot with a suction gripper sitting on a table workspace, with three camera views.

\paragraph{ALFWorld~\cite{shridhar2021alfworld}}
Building on top of the TextWorld simulator, discussed in Appendix~\ref{sec:app_sim_other}, ALFWorld aligns perception from the $3$D robot simulation benchmark, ALFRED~\cite{shridhar2020alfred}, with text-based, discrete actions like ``MoveAhead,'' ``RotateLeft,'' and ``Open.''

\paragraph{ProgPrompt~\cite{sing2023progprompt}}
ProgPrompt is a benchmark of high-fidelity $3$D data collected from a virtual home robot.
It includes three environments, each with $115$ object instances.
These simulations are further used to create a dataset of $70$ household robot tasks with a ground-truth set of actions to achieve each goal.

\paragraph{RoboEval~\cite{hu2024deploying}}
RoboEval is a general platform for checking the correctness of code generated for a robot task.
It relies on a simulator, evaluator, and a set of defined tasks to perform evaluations on a simulated robot.
While ProgPrompt captures more realistic scenarios in its high-fidelity $3$D simulator, RoboEval is tuned towards verifying code efficiently.

\paragraph{KnowNo TableSim~\cite{ren2023robots}}
More recently, the developers of KnowNo also provide a tabletop simulator based on PyBullet, like~\citet{zeng2021transporter}, for robot manipulation of blocks and bowls.
Provided instructions vary in ambiguity by attribute, number, and spatial reasoning. 

\subsubsection{Other Simulators}
\label{sec:app_sim_other}

\paragraph{TextWorld~\cite{cote2019textworld}}
TextWorld is a suite of text-based games that can be either hand-engineered or procedurally generated, where an agent directly receives text-based observations from an abstract world, and acts with natural language actions to complete a task.

\paragraph{BabyAI~\cite{chevalier2018babyai}}
\citeauthor{chevalier2018babyai} present a $2$D top-down, grid-based simulator of instruction-following tasks with varying difficulty.
Some tasks include simple navigation to a single goal, picking and placing objects with ambiguous references, and instructions that implicitly require multi-step reasoning to complete.
The simulator provides a partial observation of the space near the agent at every timestep.

\paragraph{MineDojo~\cite{fan2022minedojo}}
The developers of MineDojo attempt to create a benchmark to test the continual learning of agents in an open-world setting.
They build an interface on top of Minecraft, a video game, to enable testing with diverse open-ended tasks, and provide access to an external knowledge-base of existing Minecraft tutorials and wiki discussions.
MineDojo includes several thousands of tasks that are more complex than earlier works (and require multi-step reasoning).
As such, task completion is judged with a learned LVLM, which acts like a human evaluator.

\paragraph{Smallville~\cite{park2023generative}}
\citeauthor{park2023generative} present a multi-agent conversational simulator where agents are controlled by language models.
Users may set up agents with a defined backstory and provide instructions when desired.
Each agent has access to a memory of past experiences, and generates natural language actions to go to certain areas, communicate with others, complete chores, and more. 

\section{Guidelines on Current Methodologies}
\label{sec:app_guidelines}

\subsection{Beware of Erroneous Hallucination Predictions}
\label{sec:app_beware}

We additionally point out risks of relying on hallucination intervention algorithms and metrics that use deep learning methods at inference time.
Generally speaking, deep neural networks are black boxes with few statistical guarantees and lack interpretability~\cite{aggarwal2023neural, dobson2023reading, benitez1997artificial, taheri2021statistical, poggio2020theoretical, buskulic2024convergence}.
As such, these models may incorrectly detect hallucinations during deployment --- effectively hallucinating themselves.
In particular, engineers should take precautions when utilizing detection algorithms under method types like hidden states, attention weights, concept probabilities, analyzing samples, and proxy model, where neural networks are used frequently.
Learned proxy models or classifiers are particularly prone to incorrect predictions because they are trained on a different data distribution from the model under test~\cite{quinonero2008dataset, baek2024agreement}. 
Similarly, decreased model complexity may result in poor predictions~\cite{stoffi2022simple, hu2021model, lee2020neural}.
As such, several works we have listed in Table~\ref{tab:metrics} use LVLM-based evaluators (\eg~GPT-4) to increase evaluation model complexity and cover a broader data distribution.
However, as we have found throughout this work, LVLMs should not be overly relied upon for consistently accurate estimates.
Instead, learned hallucination detection models that further output a calibrated confidence score (relaying their uncertainty of a prediction) can assist designers with choosing how to utilize the evaluated model's decision.
Additionally, classic machine metrics like accuracy, precision, recall, false-positive rate (FPR),~\etc~provide engineers with an understanding of the effectiveness of hallucination detection methods prior to deployment.
The learned intervention algorithm should be tuned to balance true-positive rate (TPR) and FPR, such that it does not miss critical hallucinations, nor act overly conservative (predicting hallucinations too frequently).
We also suggest caution when using learning-based similarity metrics like BERTScore, SummaC, AlignScore, CLIP Score, and others listed in Table~\ref{tab:metrics} for similar reasons. 
Specifically, while these metrics have been shown to reasonably reflect the semantic similarity of varied terms, there are still cases of poor alignment~\cite{hanna2021fine, hessel2021clipscore}.
Even adversarial models used in adversarial prompting approaches are not safe from learned biases, which could lead to a poor understanding of the uncertainty of the model under test.
For example, an adversarial agent tasked with prompting an autonomous driving LVLM could provide incorrect sensor readings, obviously resulting in poor decisions from the LVLM under test.
In this case, we have not learned any additional information on the reliability of the predicted action.
Thus, we argue that deep-learning-based adversarial prompters should be grounded in accurate data to truly understand the uncertainty of model predictions.
Overall, engineers should take care when using deep learning methods for hallucination intervention --- or as metrics during evaluation --- because of their tendency to act unpredictably with out-of-domain data, and limited theoretical guarantees.

\newpage


\bibliographystyle{ACM-Reference-Format}
\bibliography{root}


\begin{thebibliography}{260}


\ifx \showCODEN    \undefined \def \showCODEN     #1{\unskip}     \fi
\ifx \showISBNx    \undefined \def \showISBNx     #1{\unskip}     \fi
\ifx \showISBNxiii \undefined \def \showISBNxiii  #1{\unskip}     \fi
\ifx \showISSN     \undefined \def \showISSN      #1{\unskip}     \fi
\ifx \showLCCN     \undefined \def \showLCCN      #1{\unskip}     \fi
\ifx \shownote     \undefined \def \shownote      #1{#1}          \fi
\ifx \showarticletitle \undefined \def \showarticletitle #1{#1}   \fi
\ifx \showURL      \undefined \def \showURL       {\relax}        \fi
\providecommand\bibfield[2]{#2}
\providecommand\bibinfo[2]{#2}
\providecommand\natexlab[1]{#1}
\providecommand\showeprint[2][]{arXiv:#2}

\bibitem[Achiam et~al\mbox{.}(2023)]%
        {achiam2023gpt}
\bibfield{author}{\bibinfo{person}{Josh Achiam}, \bibinfo{person}{Steven Adler}, \bibinfo{person}{Sandhini Agarwal}, \bibinfo{person}{Lama Ahmad}, \bibinfo{person}{Ilge Akkaya}, \bibinfo{person}{Florencia~Leoni Aleman}, \bibinfo{person}{Diogo Almeida}, \bibinfo{person}{Janko Altenschmidt}, \bibinfo{person}{Sam Altman}, \bibinfo{person}{Shyamal Anadkat}, {et~al\mbox{.}}} \bibinfo{year}{2023}\natexlab{}.
\newblock \showarticletitle{{GPT}-4 {Technical} {Report}}.
\newblock \bibinfo{journal}{\emph{arXiv preprint arXiv:2303.08774}} (\bibinfo{year}{2023}).
\newblock


\bibitem[Aggarwal(2023)]%
        {aggarwal2023neural}
\bibfield{author}{\bibinfo{person}{Charu~C. Aggarwal}.} \bibinfo{year}{2023}\natexlab{}.
\newblock \bibinfo{booktitle}{\emph{{Neural Networks and Deep Learning}}}.
\newblock \bibinfo{publisher}{Springer International Publishing, Cham}.
\newblock


\bibitem[AI(2023)]%
        {runway2023gen2}
\bibfield{author}{\bibinfo{person}{Runway AI}.} \bibinfo{year}{2023}\natexlab{}.
\newblock \showarticletitle{{Gen-2: Generate novel videos with text, images or video clips}}.
\newblock \bibinfo{journal}{\emph{Runway blog}} (\bibinfo{year}{2023}).
\newblock
\urldef\tempurl%
\url{https://runwayml.com/research/gen-2}
\showURL{%
\tempurl}


\bibitem[Ali et~al\mbox{.}(2024)]%
        {ali2024large}
\bibfield{author}{\bibinfo{person}{Soroush Ali}, \bibinfo{person}{Glicksberg~Benjamin S.}, \bibinfo{person}{Zimlichman Eyal}, \bibinfo{person}{Barash Yiftach}, \bibinfo{person}{Freeman Robert}, \bibinfo{person}{Charney~Alexander W.}, \bibinfo{person}{Nadkarni~Girish N}, {and} \bibinfo{person}{Klang Eyal}.} \bibinfo{year}{2024}\natexlab{}.
\newblock \showarticletitle{{Large Language Models Are Poor Medical Coders --- Benchmarking of Medical Code Querying}}.
\newblock \bibinfo{journal}{\emph{NEJM AI}} \bibinfo{volume}{1}, \bibinfo{number}{5} (\bibinfo{date}{25 Apr} \bibinfo{year}{2024}), \bibinfo{numpages}{13}~pages.
\newblock


\bibitem[Anantha et~al\mbox{.}(2021)]%
        {anantha2021open}
\bibfield{author}{\bibinfo{person}{Raviteja Anantha}, \bibinfo{person}{Svitlana Vakulenko}, \bibinfo{person}{Zhucheng Tu}, \bibinfo{person}{Shayne Longpre}, \bibinfo{person}{Stephen Pulman}, {and} \bibinfo{person}{Srinivas Chappidi}.} \bibinfo{year}{2021}\natexlab{}.
\newblock \showarticletitle{{Open-Domain Question Answering Goes Conversational via Question Rewriting}}. In \bibinfo{booktitle}{\emph{Proceedings of the 2021 Conference of the North American Chapter of the Association for Computational Linguistics: Human Language Technologies}} (Virtual). \bibinfo{publisher}{Association for Computational Linguistics}, \bibinfo{pages}{520--534}.
\newblock


\bibitem[Angelopoulos et~al\mbox{.}(2022)]%
        {angelopoulos2022learn}
\bibfield{author}{\bibinfo{person}{Anastasios~N. Angelopoulos}, \bibinfo{person}{Stephen Bates}, \bibinfo{person}{Emmanuel~J. Candès}, \bibinfo{person}{Michael~I. Jordan}, {and} \bibinfo{person}{Lihua Lei}.} \bibinfo{year}{2022}\natexlab{}.
\newblock \showarticletitle{{Learn then Test: Calibrating Predictive Algorithms to Achieve Risk Control}}.
\newblock \bibinfo{journal}{\emph{arXiv preprint arXiv:2110.01052}} (\bibinfo{year}{2022}).
\newblock


\bibitem[Azaria and Mitchell(2023)]%
        {azaria2023internal}
\bibfield{author}{\bibinfo{person}{Amos Azaria} {and} \bibinfo{person}{Tom Mitchell}.} \bibinfo{year}{2023}\natexlab{}.
\newblock \showarticletitle{{The Internal State of an {LLM} Knows When It{'}s Lying}}. In \bibinfo{booktitle}{\emph{Findings of the 2023 Conference on Empirical Methods in Natural Language Processing}} (Singapore). \bibinfo{publisher}{Association for Computational Linguistics}, \bibinfo{pages}{967--976}.
\newblock


\bibitem[Baek et~al\mbox{.}(2022)]%
        {baek2024agreement}
\bibfield{author}{\bibinfo{person}{Christina Baek}, \bibinfo{person}{Yiding Jiang}, \bibinfo{person}{Aditi Raghunathan}, {and} \bibinfo{person}{J.~Zico Kolter}.} \bibinfo{year}{2022}\natexlab{}.
\newblock \showarticletitle{{Agreement-on-the-line: Predicting the Performance of Neural Networks under Distribution Shift}}. In \bibinfo{booktitle}{\emph{Proceedings of the 2022 Conference on Neural Information Processing Systems}} (New Orleans, LA, USA). \bibinfo{publisher}{Curran Associates, Inc.}, \bibinfo{pages}{19274--19289}.
\newblock


\bibitem[Bai and Li(2024)]%
        {bai2024progress}
\bibfield{author}{\bibinfo{person}{Song Bai} {and} \bibinfo{person}{Jie Li}.} \bibinfo{year}{2024}\natexlab{}.
\newblock \showarticletitle{{Progress and Prospects in 3D Generative AI: A Technical Overview including 3D human}}.
\newblock \bibinfo{journal}{\emph{arXiv preprint arXiv:2401.02620}} (\bibinfo{year}{2024}).
\newblock


\bibitem[Bai et~al\mbox{.}(2024)]%
        {bai2024hallucination}
\bibfield{author}{\bibinfo{person}{Zechen Bai}, \bibinfo{person}{Pichao Wang}, \bibinfo{person}{Tianjun Xiao}, \bibinfo{person}{Tong He}, \bibinfo{person}{Zongbo Han}, \bibinfo{person}{Zheng Zhang}, {and} \bibinfo{person}{Mike~Zheng Shou}.} \bibinfo{year}{2024}\natexlab{}.
\newblock \showarticletitle{{Hallucination of Multimodal Large Language Models: A Survey}}.
\newblock \bibinfo{journal}{\emph{arXiv preprint arXiv:2404.18930}} (\bibinfo{year}{2024}).
\newblock


\bibitem[Banik(2017)]%
        {kaggle2017movies}
\bibfield{author}{\bibinfo{person}{Rounak Banik}.} \bibinfo{year}{2017}\natexlab{}.
\newblock \showarticletitle{{The Movies Dataset}}.
\newblock \bibinfo{journal}{\emph{Kaggle}} (\bibinfo{year}{2017}).
\newblock
\urldef\tempurl%
\url{https://www.kaggle.com/datasets/rounakbanik/the-movies-dataset}
\showURL{%
\tempurl}


\bibitem[Barack(2024)]%
        {barack2024using}
\bibfield{author}{\bibinfo{person}{Lauren Barack}.} \bibinfo{year}{2024}\natexlab{}.
\newblock \showarticletitle{{Using AI in lesson planning? Beware hallucinations}}.
\newblock \bibinfo{journal}{\emph{K-12 Dive}} (\bibinfo{year}{2024}).
\newblock
\urldef\tempurl%
\url{https://www.k12dive.com/news/using-ai-lesson-planning-beware-hallucinations/726660/}
\showURL{%
\tempurl}


\bibitem[Bargagli~Stoffi et~al\mbox{.}(2022)]%
        {stoffi2022simple}
\bibfield{author}{\bibinfo{person}{Falco~J Bargagli~Stoffi}, \bibinfo{person}{Gustavo Cevolani}, {and} \bibinfo{person}{Giorgio Gnecco}.} \bibinfo{year}{2022}\natexlab{}.
\newblock \showarticletitle{{Simple Models in Complex Worlds: Occam's Razor and Statistical Learning Theory}}.
\newblock \bibinfo{journal}{\emph{Minds and Machines}} \bibinfo{volume}{32}, \bibinfo{number}{1} (\bibinfo{date}{March} \bibinfo{year}{2022}), \bibinfo{pages}{13--42}.
\newblock


\bibitem[Benitez et~al\mbox{.}(1997)]%
        {benitez1997artificial}
\bibfield{author}{\bibinfo{person}{J.M. Benitez}, \bibinfo{person}{J.L. Castro}, {and} \bibinfo{person}{I. Requena}.} \bibinfo{year}{1997}\natexlab{}.
\newblock \showarticletitle{{Are Artificial Neural Networks Black Boxes?}}
\newblock \bibinfo{journal}{\emph{IEEE Transactions on Neural Networks}} \bibinfo{volume}{8}, \bibinfo{number}{5} (\bibinfo{year}{1997}), \bibinfo{pages}{1156--1164}.
\newblock


\bibitem[Betker et~al\mbox{.}(2023)]%
        {betker2023improving}
\bibfield{author}{\bibinfo{person}{James Betker}, \bibinfo{person}{Gabriel Goh}, \bibinfo{person}{Li Jing}, \bibinfo{person}{Tim Brooks}, \bibinfo{person}{Jianfeng Wang}, \bibinfo{person}{Linjie Li}, \bibinfo{person}{Long Ouyang}, \bibinfo{person}{Juntang Zhuang}, \bibinfo{person}{Joyce Lee}, \bibinfo{person}{Yufei Guo}, {et~al\mbox{.}}} \bibinfo{year}{2023}\natexlab{}.
\newblock \showarticletitle{{Improving Image Generation with Better Captions}}.
\newblock \bibinfo{journal}{\emph{OpenAI blog}} (\bibinfo{year}{2023}).
\newblock
\urldef\tempurl%
\url{https://cdn.openai.com/papers/dall-e-3.pdf}
\showURL{%
\tempurl}


\bibitem[Blankemeier et~al\mbox{.}(2024)]%
        {blankemeier2024merlin}
\bibfield{author}{\bibinfo{person}{Louis Blankemeier}, \bibinfo{person}{Joseph~Paul Cohen}, \bibinfo{person}{Ashwin Kumar}, \bibinfo{person}{Dave~Van Veen}, \bibinfo{person}{Syed Jamal~Safdar Gardezi}, \bibinfo{person}{Magdalini Paschali}, \bibinfo{person}{Zhihong Chen}, \bibinfo{person}{Jean-Benoit Delbrouck}, \bibinfo{person}{Eduardo Reis}, \bibinfo{person}{Cesar Truyts}, {et~al\mbox{.}}} \bibinfo{year}{2024}\natexlab{}.
\newblock \showarticletitle{{Merlin: A Vision Language Foundation Model for 3D Computed Tomography}}.
\newblock \bibinfo{journal}{\emph{arXiv preprint arXiv:2406.06512}} (\bibinfo{year}{2024}).
\newblock


\bibitem[Bommasani et~al\mbox{.}(2022)]%
        {bommasani2022opportunities}
\bibfield{author}{\bibinfo{person}{Rishi Bommasani}, \bibinfo{person}{Drew~A. Hudson}, \bibinfo{person}{Ehsan Adeli}, \bibinfo{person}{Russ Altman}, \bibinfo{person}{Simran Arora}, \bibinfo{person}{Sydney von Arx}, \bibinfo{person}{Michael~S. Bernstein}, \bibinfo{person}{Jeannette Bohg}, \bibinfo{person}{Antoine Bosselut}, \bibinfo{person}{Emma Brunskill}, {et~al\mbox{.}}} \bibinfo{year}{2022}\natexlab{}.
\newblock \showarticletitle{{On the Opportunities and Risks of Foundation Models}}.
\newblock \bibinfo{journal}{\emph{arXiv preprint arXiv:2108.07258}} (\bibinfo{year}{2022}).
\newblock


\bibitem[Borase et~al\mbox{.}(2021)]%
        {borase2021review}
\bibfield{author}{\bibinfo{person}{Rakesh~P Borase}, \bibinfo{person}{DK Maghade}, \bibinfo{person}{SY Sondkar}, {and} \bibinfo{person}{SN Pawar}.} \bibinfo{year}{2021}\natexlab{}.
\newblock \showarticletitle{{A review of PID control, tuning methods and applications}}.
\newblock \bibinfo{journal}{\emph{International Journal of Dynamics and Control}}  \bibinfo{volume}{9} (\bibinfo{year}{2021}), \bibinfo{pages}{818--827}.
\newblock


\bibitem[Brewer et~al\mbox{.}(1999)]%
        {brewer1999beliefs}
\bibfield{author}{\bibinfo{person}{Neil Brewer}, \bibinfo{person}{Rob Potter}, \bibinfo{person}{Ronald~P. Fisher}, \bibinfo{person}{Nigel Bond}, {and} \bibinfo{person}{Mary~A. Luszcz}.} \bibinfo{year}{1999}\natexlab{}.
\newblock \showarticletitle{{Beliefs and Data on the Relationship Between Consistency and Accuracy of Eyewitness Testimony}}.
\newblock \bibinfo{journal}{\emph{Applied Cognitive Psychology}} \bibinfo{volume}{13}, \bibinfo{number}{4} (\bibinfo{year}{1999}), \bibinfo{pages}{297--313}.
\newblock


\bibitem[Brown et~al\mbox{.}(2020)]%
        {brown2020language}
\bibfield{author}{\bibinfo{person}{Tom Brown}, \bibinfo{person}{Benjamin Mann}, \bibinfo{person}{Nick Ryder}, \bibinfo{person}{Melanie Subbiah}, \bibinfo{person}{Jared~D Kaplan}, \bibinfo{person}{Prafulla Dhariwal}, \bibinfo{person}{Arvind Neelakantan}, \bibinfo{person}{Pranav Shyam}, \bibinfo{person}{Girish Sastry}, \bibinfo{person}{Amanda Askell}, {et~al\mbox{.}}} \bibinfo{year}{2020}\natexlab{}.
\newblock \showarticletitle{{Language Models are Few-Shot Learners}}. In \bibinfo{booktitle}{\emph{Proceedings of the 2020 Conference on Neural Information Processing Systems}} (Virtual). \bibinfo{publisher}{Curran Associates, Inc.}, \bibinfo{pages}{1877--1901}.
\newblock


\bibitem[Buskulic et~al\mbox{.}(2024)]%
        {buskulic2024convergence}
\bibfield{author}{\bibinfo{person}{Nathan Buskulic}, \bibinfo{person}{Jalal Fadili}, {and} \bibinfo{person}{Yvain Qu{\'e}au}.} \bibinfo{year}{2024}\natexlab{}.
\newblock \showarticletitle{{Convergence and Recovery Guarantees of Unsupervised Neural Networks for Inverse Problems}}.
\newblock \bibinfo{journal}{\emph{Journal of Mathematical Imaging and Vision}} \bibinfo{volume}{66}, \bibinfo{number}{4} (\bibinfo{date}{Aug.} \bibinfo{year}{2024}), \bibinfo{pages}{584--605}.
\newblock


\bibitem[Caesar et~al\mbox{.}(2020)]%
        {caeser2020nuscenes}
\bibfield{author}{\bibinfo{person}{Holger Caesar}, \bibinfo{person}{Varun Bankiti}, \bibinfo{person}{Alex~H. Lang}, \bibinfo{person}{Sourabh Vora}, \bibinfo{person}{Venice~Erin Liong}, \bibinfo{person}{Qiang Xu}, \bibinfo{person}{Anush Krishnan}, \bibinfo{person}{Yu Pan}, \bibinfo{person}{Giancarlo Baldan}, {and} \bibinfo{person}{Oscar Beijbom}.} \bibinfo{year}{2020}\natexlab{}.
\newblock \showarticletitle{{nuScenes: A Multimodal Dataset for Autonomous Driving}}. In \bibinfo{booktitle}{\emph{Proceedings of the 2020 IEEE/CVF Conference on Computer Vision and Pattern Recognition (CVPR)}} (Virtual). \bibinfo{publisher}{Institute of Electrical and Electronics Engineers}, \bibinfo{pages}{11618--11628}.
\newblock


\bibitem[Cao et~al\mbox{.}(2022)]%
        {cao2022kqa}
\bibfield{author}{\bibinfo{person}{Shulin Cao}, \bibinfo{person}{Jiaxin Shi}, \bibinfo{person}{Liangming Pan}, \bibinfo{person}{Lunyiu Nie}, \bibinfo{person}{Yutong Xiang}, \bibinfo{person}{Lei Hou}, \bibinfo{person}{Juanzi Li}, \bibinfo{person}{Bin He}, {and} \bibinfo{person}{Hanwang Zhang}.} \bibinfo{year}{2022}\natexlab{}.
\newblock \showarticletitle{{KQA Pro: A Dataset with Explicit Compositional Programs for Complex Question Answering over Knowledge Base}}. In \bibinfo{booktitle}{\emph{Proceedings of the 60th Annual Meeting of the Association for Computational Linguistics}} (Dublin, Ireland). \bibinfo{publisher}{Association for Computational Linguistics}, \bibinfo{pages}{6101--6119}.
\newblock


\bibitem[Caron et~al\mbox{.}(2021)]%
        {caron2021emerging}
\bibfield{author}{\bibinfo{person}{Mathilde Caron}, \bibinfo{person}{Hugo Touvron}, \bibinfo{person}{Ishan Misra}, \bibinfo{person}{Hervé Jegou}, \bibinfo{person}{Julien Mairal}, \bibinfo{person}{Piotr Bojanowski}, {and} \bibinfo{person}{Armand Joulin}.} \bibinfo{year}{2021}\natexlab{}.
\newblock \showarticletitle{{Emerging Properties in Self-Supervised Vision Transformers}}. In \bibinfo{booktitle}{\emph{Proceedings of the 2021 IEEE/CVF Conference on International Conference on Computer Vision (ICCV)}} (Virtual). \bibinfo{publisher}{Institute of Electrical and Electronics Engineers}, \bibinfo{pages}{9630--9640}.
\newblock


\bibitem[Chakraborty et~al\mbox{.}(2023)]%
        {chakraborty2023structural}
\bibfield{author}{\bibinfo{person}{Neeloy Chakraborty}, \bibinfo{person}{Aamir Hasan}, \bibinfo{person}{Shuijing Liu}, \bibinfo{person}{Tianchen Ji}, \bibinfo{person}{Weihang Liang}, \bibinfo{person}{D.~Livingston McPherson}, {and} \bibinfo{person}{Katherine Driggs-Campbell}.} \bibinfo{year}{2023}\natexlab{}.
\newblock \showarticletitle{{Structural Attention-based Recurrent Variational Autoencoder for Highway Vehicle Anomaly Detection}}. In \bibinfo{booktitle}{\emph{Proceedings of the 2023 International Conference on Autonomous Agents and Multiagent Systems}} (London, United Kingdom). \bibinfo{publisher}{International Foundation for Autonomous Agents and Multiagent Systems}, \bibinfo{address}{Richland, SC, USA}, \bibinfo{pages}{1125–1134}.
\newblock
\showISBNx{9781450394321}


\bibitem[Chen et~al\mbox{.}(2023b)]%
        {chen2023purr}
\bibfield{author}{\bibinfo{person}{Anthony Chen}, \bibinfo{person}{Panupong Pasupat}, \bibinfo{person}{Sameer Singh}, \bibinfo{person}{Hongrae Lee}, {and} \bibinfo{person}{Kelvin Guu}.} \bibinfo{year}{2023}\natexlab{b}.
\newblock \showarticletitle{{PURR: Efficiently Editing Language Model Hallucinations by Denoising Language Model Corruptions}}.
\newblock \bibinfo{journal}{\emph{arXiv preprint arXiv:2305.14908}} (\bibinfo{year}{2023}).
\newblock


\bibitem[Chen and Shu(2024)]%
        {chen2024can}
\bibfield{author}{\bibinfo{person}{Canyu Chen} {and} \bibinfo{person}{Kai Shu}.} \bibinfo{year}{2024}\natexlab{}.
\newblock \showarticletitle{{Can {LLM}-Generated Misinformation Be Detected?}}. In \bibinfo{booktitle}{\emph{Proceedings of the 12th International Conference on Learning Representations}} (Vienna, Austria).
\newblock


\bibitem[Chen et~al\mbox{.}(2017)]%
        {chen2017reading}
\bibfield{author}{\bibinfo{person}{Danqi Chen}, \bibinfo{person}{Adam Fisch}, \bibinfo{person}{Jason Weston}, {and} \bibinfo{person}{Antoine Bordes}.} \bibinfo{year}{2017}\natexlab{}.
\newblock \showarticletitle{{Reading {W}ikipedia to Answer Open-Domain Questions}}. In \bibinfo{booktitle}{\emph{Proceedings of the 55th Annual Meeting of the Association for Computational Linguistics}} (Vancouver, Canada). \bibinfo{publisher}{Association for Computational Linguistics}, \bibinfo{pages}{1870--1879}.
\newblock


\bibitem[Chen et~al\mbox{.}(2024b)]%
        {chen2023driving}
\bibfield{author}{\bibinfo{person}{Long Chen}, \bibinfo{person}{Oleg Sinavski}, \bibinfo{person}{Jan Hünermann}, \bibinfo{person}{Alice Karnsund}, \bibinfo{person}{Andrew~James Willmott}, \bibinfo{person}{Danny Birch}, \bibinfo{person}{Daniel Maund}, {and} \bibinfo{person}{Jamie Shotton}.} \bibinfo{year}{2024}\natexlab{b}.
\newblock \showarticletitle{{Driving with LLMs: Fusing Object-Level Vector Modality for Explainable Autonomous Driving}}. In \bibinfo{booktitle}{\emph{Proceedings of the 2024 IEEE International Conference on Robotics and Automation (ICRA)}} (Yokohama, Japan). \bibinfo{publisher}{Institute of Electrical and Electronics Engineers}, \bibinfo{pages}{14093--14100}.
\newblock


\bibitem[Chen et~al\mbox{.}(2023c)]%
        {chen2023introspective}
\bibfield{author}{\bibinfo{person}{Liting Chen}, \bibinfo{person}{Lu Wang}, \bibinfo{person}{Hang Dong}, \bibinfo{person}{Yali Du}, \bibinfo{person}{Jie Yan}, \bibinfo{person}{Fangkai Yang}, \bibinfo{person}{Shuang Li}, \bibinfo{person}{Pu Zhao}, \bibinfo{person}{Si Qin}, \bibinfo{person}{Saravan Rajmohan}, {et~al\mbox{.}}} \bibinfo{year}{2023}\natexlab{c}.
\newblock \showarticletitle{{Introspective Tips: Large Language Model for In-Context Decision Making}}.
\newblock \bibinfo{journal}{\emph{arXiv preprint arXiv:2305.11598}} (\bibinfo{year}{2023}).
\newblock


\bibitem[Chen et~al\mbox{.}(2021)]%
        {chen2021evaluating}
\bibfield{author}{\bibinfo{person}{Mark Chen}, \bibinfo{person}{Jerry Tworek}, \bibinfo{person}{Heewoo Jun}, \bibinfo{person}{Qiming Yuan}, \bibinfo{person}{Henrique~Ponde de Oliveira~Pinto}, \bibinfo{person}{Jared Kaplan}, \bibinfo{person}{Harri Edwards}, \bibinfo{person}{Yuri Burda}, \bibinfo{person}{Nicholas Joseph}, \bibinfo{person}{Greg Brockman}, {et~al\mbox{.}}} \bibinfo{year}{2021}\natexlab{}.
\newblock \showarticletitle{{Evaluating Large Language Models Trained on Code}}.
\newblock \bibinfo{journal}{\emph{arXiv preprint arXiv:2107.03374}} (\bibinfo{year}{2021}).
\newblock


\bibitem[Chen et~al\mbox{.}(2024a)]%
        {chen2024meshxl}
\bibfield{author}{\bibinfo{person}{Sijin Chen}, \bibinfo{person}{Xin Chen}, \bibinfo{person}{Anqi Pang}, \bibinfo{person}{Xianfang Zeng}, \bibinfo{person}{Wei Cheng}, \bibinfo{person}{Yijun Fu}, \bibinfo{person}{Fukun Yin}, \bibinfo{person}{Zhibin Wang}, \bibinfo{person}{Jingyi Yu}, \bibinfo{person}{Gang Yu}, {et~al\mbox{.}}} \bibinfo{year}{2024}\natexlab{a}.
\newblock \showarticletitle{{Mesh{XL}: Neural Coordinate Field for Generative 3D Foundation Models}}. In \bibinfo{booktitle}{\emph{Proceedings of the 2024 Conference on Neural Information Processing Systems}}.
\newblock


\bibitem[Chen et~al\mbox{.}(2023a)]%
        {chen2023hallucination}
\bibfield{author}{\bibinfo{person}{Yuyan Chen}, \bibinfo{person}{Qiang Fu}, \bibinfo{person}{Yichen Yuan}, \bibinfo{person}{Zhihao Wen}, \bibinfo{person}{Ge Fan}, \bibinfo{person}{Dayiheng Liu}, \bibinfo{person}{Dongmei Zhang}, \bibinfo{person}{Zhixu Li}, {and} \bibinfo{person}{Yanghua Xiao}.} \bibinfo{year}{2023}\natexlab{a}.
\newblock \showarticletitle{{Hallucination Detection: Robustly Discerning Reliable Answers in Large Language Models}}. In \bibinfo{booktitle}{\emph{Proceedings of the 32nd ACM International Conference on Information and Knowledge Management}} (Birmingham, United Kingdom). \bibinfo{publisher}{Association for Computing Machinery}, \bibinfo{address}{New York, NY, USA}, \bibinfo{pages}{245–255}.
\newblock
\showISBNx{9798400701245}


\bibitem[Chevalier-Boisvert et~al\mbox{.}(2019)]%
        {chevalier2018babyai}
\bibfield{author}{\bibinfo{person}{Maxime Chevalier-Boisvert}, \bibinfo{person}{Dzmitry Bahdanau}, \bibinfo{person}{Salem Lahlou}, \bibinfo{person}{Lucas Willems}, \bibinfo{person}{Chitwan Saharia}, \bibinfo{person}{Thien~Huu Nguyen}, {and} \bibinfo{person}{Yoshua Bengio}.} \bibinfo{year}{2019}\natexlab{}.
\newblock \showarticletitle{{Baby{AI}: First Steps Towards Grounded Language Learning With a Human In the Loop}}. In \bibinfo{booktitle}{\emph{Proceedings of the 7th International Conference on Learning Representations}} (New Orleans, LA, USA).
\newblock


\bibitem[Chowdhery et~al\mbox{.}(2023)]%
        {chowdhery2023palm}
\bibfield{author}{\bibinfo{person}{Aakanksha Chowdhery}, \bibinfo{person}{Sharan Narang}, \bibinfo{person}{Jacob Devlin}, \bibinfo{person}{Maarten Bosma}, \bibinfo{person}{Gaurav Mishra}, \bibinfo{person}{Adam Roberts}, \bibinfo{person}{Paul Barham}, \bibinfo{person}{Hyung~Won Chung}, \bibinfo{person}{Charles Sutton}, \bibinfo{person}{Sebastian Gehrmann}, {et~al\mbox{.}}} \bibinfo{year}{2023}\natexlab{}.
\newblock \showarticletitle{{PaLM: Scaling Language Modeling with Pathways}}.
\newblock \bibinfo{journal}{\emph{Journal of Machine Learning Research}} \bibinfo{volume}{24}, \bibinfo{number}{240} (\bibinfo{year}{2023}).
\newblock


\bibitem[Chu et~al\mbox{.}(2024)]%
        {chu2024sora}
\bibfield{author}{\bibinfo{person}{Zhixuan Chu}, \bibinfo{person}{Lei Zhang}, \bibinfo{person}{Yichen Sun}, \bibinfo{person}{Siqiao Xue}, \bibinfo{person}{Zhibo Wang}, \bibinfo{person}{Zhan Qin}, {and} \bibinfo{person}{Kui Ren}.} \bibinfo{year}{2024}\natexlab{}.
\newblock \showarticletitle{{Sora Detector: A Unified Hallucination Detection for Large Text-to-Video Models}}.
\newblock \bibinfo{journal}{\emph{arXiv preprint arXiv:2405.04180}} (\bibinfo{year}{2024}).
\newblock


\bibitem[Clusmann et~al\mbox{.}(2023)]%
        {clusmann2023future}
\bibfield{author}{\bibinfo{person}{Jan Clusmann}, \bibinfo{person}{Fiona~R. Kolbinger}, \bibinfo{person}{Hannah~Sophie Muti}, \bibinfo{person}{Zunamys~I. Carrero}, \bibinfo{person}{Jan-Niklas Eckardt}, \bibinfo{person}{Narmin~Ghaffari Laleh}, \bibinfo{person}{Chiara Maria~Lavinia L{\"o}ffler}, \bibinfo{person}{Sophie-Caroline Schwarzkopf}, \bibinfo{person}{Michaela Unger}, \bibinfo{person}{Gregory~P. Veldhuizen}, {et~al\mbox{.}}} \bibinfo{year}{2023}\natexlab{}.
\newblock \showarticletitle{{The future landscape of large language models in medicine}}.
\newblock \bibinfo{journal}{\emph{Communications Medicine}} \bibinfo{volume}{3}, \bibinfo{number}{1} (\bibinfo{date}{10 Oct} \bibinfo{year}{2023}), \bibinfo{pages}{141}.
\newblock
\showISSN{2730-664X}


\bibitem[Cobbe et~al\mbox{.}(2021)]%
        {cobbe2021training}
\bibfield{author}{\bibinfo{person}{Karl Cobbe}, \bibinfo{person}{Vineet Kosaraju}, \bibinfo{person}{Mohammad Bavarian}, \bibinfo{person}{Mark Chen}, \bibinfo{person}{Heewoo Jun}, \bibinfo{person}{Lukasz Kaiser}, \bibinfo{person}{Matthias Plappert}, \bibinfo{person}{Jerry Tworek}, \bibinfo{person}{Jacob Hilton}, \bibinfo{person}{Reiichiro Nakano}, {et~al\mbox{.}}} \bibinfo{year}{2021}\natexlab{}.
\newblock \showarticletitle{{Training Verifiers to Solve Math Word Problems}}.
\newblock \bibinfo{journal}{\emph{arXiv preprint arXiv:2110.14168}} (\bibinfo{year}{2021}).
\newblock


\bibitem[Constantino(2024)]%
        {constantino2024ai}
\bibfield{author}{\bibinfo{person}{Tor Constantino}.} \bibinfo{year}{2024}\natexlab{}.
\newblock \showarticletitle{{AI Experts Test Perplexity's New Election Hub}}.
\newblock \bibinfo{journal}{\emph{Forbes}} (\bibinfo{year}{2024}).
\newblock
\urldef\tempurl%
\url{https://www.forbes.com/sites/torconstantino/2024/11/04/perplexitys-new-election-hub-triggers-reactions-from-ai-experts/}
\showURL{%
\tempurl}


\bibitem[C{\^o}t{\'e} et~al\mbox{.}(2019)]%
        {cote2019textworld}
\bibfield{author}{\bibinfo{person}{Marc-Alexandre C{\^o}t{\'e}}, \bibinfo{person}{{\'A}kos K{\'a}d{\'a}r}, \bibinfo{person}{Xingdi Yuan}, \bibinfo{person}{Ben Kybartas}, \bibinfo{person}{Tavian Barnes}, \bibinfo{person}{Emery Fine}, \bibinfo{person}{James Moore}, \bibinfo{person}{Matthew Hausknecht}, \bibinfo{person}{Layla El~Asri}, {et~al\mbox{.}}} \bibinfo{year}{2019}\natexlab{}.
\newblock \showarticletitle{{TextWorld: A Learning Environment for Text-Based Games}}. In \bibinfo{booktitle}{\emph{Proceedings of the 7th Computer Games Workshop at the 27th International Conference on Artificial Intelligence}} (Stockholm, Sweden). \bibinfo{publisher}{Springer International Publishing}, \bibinfo{pages}{41--75}.
\newblock
\showISBNx{978-3-030-24337-1}


\bibitem[Coumans and Bai(2021)]%
        {coumans2016pybullet}
\bibfield{author}{\bibinfo{person}{Erwin Coumans} {and} \bibinfo{person}{Yunfei Bai}.} \bibinfo{year}{2016--2021}\natexlab{}.
\newblock \bibinfo{title}{{PyBullet, a Python module for physics simulation for games, robotics and machine learning}}.
\newblock \bibinfo{howpublished}{\url{http://pybullet.org}}.
\newblock


\bibitem[Cui et~al\mbox{.}(2024)]%
        {cui2024survey}
\bibfield{author}{\bibinfo{person}{Can Cui}, \bibinfo{person}{Yunsheng Ma}, \bibinfo{person}{Xu Cao}, \bibinfo{person}{Wenqian Ye}, \bibinfo{person}{Yang Zhou}, \bibinfo{person}{Kaizhao Liang}, \bibinfo{person}{Jintai Chen}, \bibinfo{person}{Juanwu Lu}, \bibinfo{person}{Zichong Yang}, \bibinfo{person}{Kuei-Da Liao}, {et~al\mbox{.}}} \bibinfo{year}{2024}\natexlab{}.
\newblock \showarticletitle{{A Survey on Multimodal Large Language Models for Autonomous Driving}}. In \bibinfo{booktitle}{\emph{Proceedings of the 1st Workshop on Large Language and Vision Models for Autonomous Driving at the 2024 IEEE/CVF Winter Conference on Applications of Computer Vision (WACV)}} (Waikoloa, HI, USA). \bibinfo{publisher}{Institute of Electrical and Electronics Engineers}, \bibinfo{pages}{958--979}.
\newblock


\bibitem[Dahl et~al\mbox{.}(2024)]%
        {dahl2024large}
\bibfield{author}{\bibinfo{person}{Matthew Dahl}, \bibinfo{person}{Varun Magesh}, \bibinfo{person}{Mirac Suzgun}, {and} \bibinfo{person}{Daniel~E Ho}.} \bibinfo{year}{2024}\natexlab{}.
\newblock \showarticletitle{{Large Legal Fictions: Profiling Legal Hallucinations in Large Language Models}}.
\newblock \bibinfo{journal}{\emph{Journal of Legal Analysis}} \bibinfo{volume}{16}, \bibinfo{number}{1} (\bibinfo{date}{06} \bibinfo{year}{2024}), \bibinfo{pages}{64--93}.
\newblock
\showISSN{2161-7201}


\bibitem[Dai et~al\mbox{.}(2023)]%
        {dai2023plausible}
\bibfield{author}{\bibinfo{person}{Wenliang Dai}, \bibinfo{person}{Zihan Liu}, \bibinfo{person}{Ziwei Ji}, \bibinfo{person}{Dan Su}, {and} \bibinfo{person}{Pascale Fung}.} \bibinfo{year}{2023}\natexlab{}.
\newblock \showarticletitle{{Plausible May Not Be Faithful: Probing Object Hallucination in Vision-Language Pre-training}}. In \bibinfo{booktitle}{\emph{Proceedings of the 17th Conference of the European Chapter of the Association for Computational Linguistics}}. \bibinfo{publisher}{Association for Computational Linguistics}, \bibinfo{address}{Dubrovnik, Croatia}, \bibinfo{pages}{2136--2148}.
\newblock


\bibitem[de~Moura and Bj{\o}rner(2008)]%
        {moura2008z3}
\bibfield{author}{\bibinfo{person}{Leonardo de Moura} {and} \bibinfo{person}{Nikolaj Bj{\o}rner}.} \bibinfo{year}{2008}\natexlab{}.
\newblock \showarticletitle{{Z3: An Efficient SMT Solver}}. In \bibinfo{booktitle}{\emph{Proceedings of the 14th International Conference on Tools and Algorithms for the Construction and Analysis of Systems}} (Budapest, Hungary). \bibinfo{publisher}{Springer Berlin Heidelberg}, \bibinfo{pages}{337--340}.
\newblock
\showISBNx{978-3-540-78800-3}


\bibitem[Deng et~al\mbox{.}(2009)]%
        {deng2009imagenet}
\bibfield{author}{\bibinfo{person}{Jia Deng}, \bibinfo{person}{Wei Dong}, \bibinfo{person}{Richard Socher}, \bibinfo{person}{Li-Jia Li}, \bibinfo{person}{Kai Li}, {and} \bibinfo{person}{Li Fei-Fei}.} \bibinfo{year}{2009}\natexlab{}.
\newblock \showarticletitle{{ImageNet: A Large-Scale Hierarchical Image Database}}. In \bibinfo{booktitle}{\emph{Proceedings of the 2009 IEEE Conference on Computer Vision and Pattern Recognition (CVPR)}} (Miami, FL, USA). \bibinfo{publisher}{Institute of Electrical and Electronics Engineers}, \bibinfo{pages}{248--255}.
\newblock


\bibitem[Deruyttere et~al\mbox{.}(2019)]%
        {deruyttere2019talk2car}
\bibfield{author}{\bibinfo{person}{Thierry Deruyttere}, \bibinfo{person}{Simon Vandenhende}, \bibinfo{person}{Dusan Grujicic}, \bibinfo{person}{Luc Van~Gool}, {and} \bibinfo{person}{Marie-Francine Moens}.} \bibinfo{year}{2019}\natexlab{}.
\newblock \showarticletitle{{{T}alk2{C}ar: Taking Control of Your Self-Driving Car}}. In \bibinfo{booktitle}{\emph{Proceedings of the 2019 Conference on Empirical Methods in Natural Language Processing and the 9th International Joint Conference on Natural Language Processing (EMNLP-IJCNLP)}} (Hong Kong, China). \bibinfo{publisher}{Association for Computational Linguistics}, \bibinfo{pages}{2088--2098}.
\newblock


\bibitem[Devlin et~al\mbox{.}(2019)]%
        {devlin2019bert}
\bibfield{author}{\bibinfo{person}{Jacob Devlin}, \bibinfo{person}{Ming-Wei Chang}, \bibinfo{person}{Kenton Lee}, {and} \bibinfo{person}{Kristina Toutanova}.} \bibinfo{year}{2019}\natexlab{}.
\newblock \showarticletitle{{{BERT}: Pre-training of Deep Bidirectional Transformers for Language Understanding}}. In \bibinfo{booktitle}{\emph{Proceedings of the 2019 Conference of the North {A}merican Chapter of the Association for Computational Linguistics: Human Language Technologies}} (Minneapolis, MN, USA). \bibinfo{publisher}{Association for Computational Linguistics}, \bibinfo{pages}{4171--4186}.
\newblock


\bibitem[Dhuliawala et~al\mbox{.}(2024)]%
        {dhuliawala2023chain}
\bibfield{author}{\bibinfo{person}{Shehzaad Dhuliawala}, \bibinfo{person}{Mojtaba Komeili}, \bibinfo{person}{Jing Xu}, \bibinfo{person}{Roberta Raileanu}, \bibinfo{person}{Xian Li}, \bibinfo{person}{Asli Celikyilmaz}, {and} \bibinfo{person}{Jason~E Weston}.} \bibinfo{year}{2024}\natexlab{}.
\newblock \showarticletitle{{Chain-of-Verification Reduces Hallucination in Large Language Models}}. In \bibinfo{booktitle}{\emph{Proceedings of the Workshop on Reliable and Responsible Foundation Models at the 12th International Conference on Learning Representations}} (Vienna, Austria).
\newblock


\bibitem[Ding et~al\mbox{.}(2024)]%
        {ding2024holistic}
\bibfield{author}{\bibinfo{person}{Xinpeng Ding}, \bibinfo{person}{Jianhua Han}, \bibinfo{person}{Hang Xu}, \bibinfo{person}{Xiaodan Liang}, \bibinfo{person}{Wei Zhang}, {and} \bibinfo{person}{Xiaomeng Li}.} \bibinfo{year}{2024}\natexlab{}.
\newblock \showarticletitle{{Holistic Autonomous Driving Understanding by Bird's-Eye-View Injected Multi-Modal Large Models}}. In \bibinfo{booktitle}{\emph{Proceedings of the 2024 IEEE/CVF Conference on Computer Vision and Pattern Recognition (CVPR)}} (Seattle, WA, USA). \bibinfo{publisher}{Institute of Electrical and Electronics Engineers}, \bibinfo{pages}{13668--13677}.
\newblock


\bibitem[Ding et~al\mbox{.}(2023)]%
        {ding2023hilmd}
\bibfield{author}{\bibinfo{person}{Xinpeng Ding}, \bibinfo{person}{Jianhua Han}, \bibinfo{person}{Hang Xu}, \bibinfo{person}{Wei Zhang}, {and} \bibinfo{person}{Xiaomeng Li}.} \bibinfo{year}{2023}\natexlab{}.
\newblock \showarticletitle{{HiLM-D: Towards High-Resolution Understanding in Multimodal Large Language Models for Autonomous Driving}}.
\newblock \bibinfo{journal}{\emph{arXiv preprint arXiv:2309.05186}} (\bibinfo{year}{2023}).
\newblock


\bibitem[Dobson(2023)]%
        {dobson2023reading}
\bibfield{author}{\bibinfo{person}{James~E Dobson}.} \bibinfo{year}{2023}\natexlab{}.
\newblock \showarticletitle{{On reading and interpreting black box deep neural networks}}.
\newblock \bibinfo{journal}{\emph{International Journal of Digital Humanities}} \bibinfo{volume}{5}, \bibinfo{number}{2} (\bibinfo{date}{Nov.} \bibinfo{year}{2023}), \bibinfo{pages}{431--449}.
\newblock


\bibitem[Dong et~al\mbox{.}(2023)]%
        {dong2023survey}
\bibfield{author}{\bibinfo{person}{Qingxiu Dong}, \bibinfo{person}{Lei Li}, \bibinfo{person}{Damai Dai}, \bibinfo{person}{Ce Zheng}, \bibinfo{person}{Zhiyong Wu}, \bibinfo{person}{Baobao Chang}, \bibinfo{person}{Xu Sun}, \bibinfo{person}{Jingjing Xu}, \bibinfo{person}{Lei Li}, {and} \bibinfo{person}{Zhifang Sui}.} \bibinfo{year}{2023}\natexlab{}.
\newblock \showarticletitle{{A Survey on In-context Learning}}.
\newblock \bibinfo{journal}{\emph{arXiv preprint arXiv:2301.00234}} (\bibinfo{year}{2023}).
\newblock


\bibitem[Dosovitskiy et~al\mbox{.}(2017)]%
        {dosovitskiy2017carla}
\bibfield{author}{\bibinfo{person}{Alexey Dosovitskiy}, \bibinfo{person}{German Ros}, \bibinfo{person}{Felipe Codevilla}, \bibinfo{person}{Antonio Lopez}, {and} \bibinfo{person}{Vladlen Koltun}.} \bibinfo{year}{2017}\natexlab{}.
\newblock \showarticletitle{{{CARLA}: {An} Open Urban Driving Simulator}}. In \bibinfo{booktitle}{\emph{Proceedings of the 1st Conference on Robot Learning}} (Mountain View, CA, USA) \emph{(\bibinfo{series}{Proceedings of Machine Learning Research}, Vol.~\bibinfo{volume}{78})}. \bibinfo{publisher}{PMLR}, \bibinfo{pages}{1--16}.
\newblock


\bibitem[Driess et~al\mbox{.}(2023)]%
        {driess2023palme}
\bibfield{author}{\bibinfo{person}{Danny Driess}, \bibinfo{person}{Fei Xia}, \bibinfo{person}{Mehdi S.~M. Sajjadi}, \bibinfo{person}{Corey Lynch}, \bibinfo{person}{Aakanksha Chowdhery}, \bibinfo{person}{Brian Ichter}, \bibinfo{person}{Ayzaan Wahid}, \bibinfo{person}{Jonathan Tompson}, \bibinfo{person}{Quan Vuong}, \bibinfo{person}{Tianhe Yu}, {et~al\mbox{.}}} \bibinfo{year}{2023}\natexlab{}.
\newblock \showarticletitle{{{P}a{LM}-E: An Embodied Multimodal Language Model}}. In \bibinfo{booktitle}{\emph{Proceedings of the 40th International Conference on Machine Learning}} (Honolulu, HI, USA) \emph{(\bibinfo{series}{Proceedings of Machine Learning Research}, Vol.~\bibinfo{volume}{202})}. \bibinfo{publisher}{PMLR}, \bibinfo{pages}{8469--8488}.
\newblock


\bibitem[Du et~al\mbox{.}(2024)]%
        {du2023improving}
\bibfield{author}{\bibinfo{person}{Yilun Du}, \bibinfo{person}{Shuang Li}, \bibinfo{person}{Antonio Torralba}, \bibinfo{person}{Joshua~B. Tenenbaum}, {and} \bibinfo{person}{Igor Mordatch}.} \bibinfo{year}{2024}\natexlab{}.
\newblock \showarticletitle{{Improving Factuality and Reasoning in Language Models through Multiagent Debate}}. In \bibinfo{booktitle}{\emph{Proceedings of the 41st International Conference on Machine Learning}} (Vienna, Austria) \emph{(\bibinfo{series}{Proceedings of Machine Learning Research}, Vol.~\bibinfo{volume}{235})}. \bibinfo{publisher}{PMLR}, \bibinfo{pages}{11733--11763}.
\newblock


\bibitem[Ekenberg(2000)]%
        {ekenberg2000logic}
\bibfield{author}{\bibinfo{person}{L Ekenberg}.} \bibinfo{year}{2000}\natexlab{}.
\newblock \showarticletitle{{The logic of conflicts between decision making agents}}.
\newblock \bibinfo{journal}{\emph{Journal of Logic and Computation}} \bibinfo{volume}{10}, \bibinfo{number}{4} (\bibinfo{date}{08} \bibinfo{year}{2000}), \bibinfo{pages}{583--602}.
\newblock
\showISSN{0955-792X}


\bibitem[Elaraby et~al\mbox{.}(2023)]%
        {elaraby2023halo}
\bibfield{author}{\bibinfo{person}{Mohamed Elaraby}, \bibinfo{person}{Mengyin Lu}, \bibinfo{person}{Jacob Dunn}, \bibinfo{person}{Xueying Zhang}, \bibinfo{person}{Yu Wang}, \bibinfo{person}{Shizhu Liu}, \bibinfo{person}{Pingchuan Tian}, \bibinfo{person}{Yuping Wang}, {and} \bibinfo{person}{Yuxuan Wang}.} \bibinfo{year}{2023}\natexlab{}.
\newblock \showarticletitle{{Halo: Estimation and Reduction of Hallucinations in Open-Source Weak Large Language Models}}.
\newblock \bibinfo{journal}{\emph{arXiv preprint arXiv:2308.11764}} (\bibinfo{year}{2023}).
\newblock


\bibitem[Elgiriyewithana(2023)]%
        {kaggle2017gci}
\bibfield{author}{\bibinfo{person}{Nidula Elgiriyewithana}.} \bibinfo{year}{2023}\natexlab{}.
\newblock \showarticletitle{{Global Country Information Dataset 2023}}.
\newblock \bibinfo{journal}{\emph{Kaggle}} (\bibinfo{year}{2023}).
\newblock
\urldef\tempurl%
\url{https://www.kaggle.com/datasets/nelgiriyewithana/countries-of-the-world-2023}
\showURL{%
\tempurl}


\bibitem[Fan et~al\mbox{.}(2022)]%
        {fan2022minedojo}
\bibfield{author}{\bibinfo{person}{Linxi Fan}, \bibinfo{person}{Guanzhi Wang}, \bibinfo{person}{Yunfan Jiang}, \bibinfo{person}{Ajay Mandlekar}, \bibinfo{person}{Yuncong Yang}, \bibinfo{person}{Haoyi Zhu}, \bibinfo{person}{Andrew Tang}, \bibinfo{person}{De-An Huang}, \bibinfo{person}{Yuke Zhu}, {and} \bibinfo{person}{Anima Anandkumar}.} \bibinfo{year}{2022}\natexlab{}.
\newblock \showarticletitle{{MineDojo: Building Open-Ended Embodied Agents with Internet-Scale Knowledge}}. In \bibinfo{booktitle}{\emph{Proceedings of the 2022 Conference on Neural Information Processing Systems}} (New Orleans, LA, USA). \bibinfo{publisher}{Curran Associates, Inc.}, \bibinfo{pages}{18343--18362}.
\newblock


\bibitem[Formentin et~al\mbox{.}(2013)]%
        {formentin2013model}
\bibfield{author}{\bibinfo{person}{Simone Formentin}, \bibinfo{person}{Klaske van Heusden}, {and} \bibinfo{person}{Alireza Karimi}.} \bibinfo{year}{2013}\natexlab{}.
\newblock \showarticletitle{{Model-based and data-driven model-reference control: A comparative analysis}}. In \bibinfo{booktitle}{\emph{Proceedings of the 2013 European Control Conference (ECC)}} (Zurich, Switzerland). \bibinfo{publisher}{Institute of Electrical and Electronics Engineers}, \bibinfo{pages}{1410--1415}.
\newblock


\bibitem[Foundation({[n.\,d.]})]%
        {wikimedia2022wiki}
\bibfield{author}{\bibinfo{person}{Wikimedia Foundation}.} \bibinfo{year}{[n.\,d.]}\natexlab{}.
\newblock \bibinfo{booktitle}{\emph{Wikimedia Downloads}}.
\newblock
\urldef\tempurl%
\url{https://dumps.wikimedia.org}
\showURL{%
\tempurl}


\bibitem[Freitag and Al-Onaizan(2017)]%
        {freitag2017beam}
\bibfield{author}{\bibinfo{person}{Markus Freitag} {and} \bibinfo{person}{Yaser Al-Onaizan}.} \bibinfo{year}{2017}\natexlab{}.
\newblock \showarticletitle{{Beam Search Strategies for Neural Machine Translation}}. In \bibinfo{booktitle}{\emph{Proceedings of the 1st Workshop on Neural Machine Translation}} (Vancouver, Canada). \bibinfo{publisher}{Association for Computational Linguistics}, \bibinfo{pages}{56--60}.
\newblock


\bibitem[Fu et~al\mbox{.}(2024a)]%
        {fu2023drive}
\bibfield{author}{\bibinfo{person}{Daocheng Fu}, \bibinfo{person}{Xin Li}, \bibinfo{person}{Licheng Wen}, \bibinfo{person}{Min Dou}, \bibinfo{person}{Pinlong Cai}, \bibinfo{person}{Botian Shi}, {and} \bibinfo{person}{Yu Qiao}.} \bibinfo{year}{2024}\natexlab{a}.
\newblock \showarticletitle{{Drive Like a Human: Rethinking Autonomous Driving with Large Language Models}}. In \bibinfo{booktitle}{\emph{Proceedings of the 2024 IEEE/CVF Winter Conference on Applications of Computer Vision Workshops (WACVW)}} (Waikola, HI, USA). \bibinfo{publisher}{Institute of Electrical and Electronics Engineers}, \bibinfo{pages}{910--919}.
\newblock


\bibitem[Fu et~al\mbox{.}(2024b)]%
        {fu2023gptscore}
\bibfield{author}{\bibinfo{person}{Jinlan Fu}, \bibinfo{person}{See-Kiong Ng}, \bibinfo{person}{Zhengbao Jiang}, {and} \bibinfo{person}{Pengfei Liu}.} \bibinfo{year}{2024}\natexlab{b}.
\newblock \showarticletitle{{{GPTS}core: Evaluate as You Desire}}. In \bibinfo{booktitle}{\emph{Proceedings of the 2024 Conference of the North American Chapter of the Association for Computational Linguistics: Human Language Technologies}} (Mexico City, Mexico). \bibinfo{publisher}{Association for Computational Linguistics}, \bibinfo{pages}{6556--6576}.
\newblock


\bibitem[Gallifant et~al\mbox{.}(2024)]%
        {gallifant2024peer}
\bibfield{author}{\bibinfo{person}{Jack Gallifant}, \bibinfo{person}{Amelia Fiske}, \bibinfo{person}{Yulia~A. Levites~Strekalova}, \bibinfo{person}{Juan~S. Osorio-Valencia}, \bibinfo{person}{Rachael Parke}, \bibinfo{person}{Rogers Mwavu}, \bibinfo{person}{Nicole Martinez}, \bibinfo{person}{Judy~Wawira Gichoya}, \bibinfo{person}{Marzyeh Ghassemi}, \bibinfo{person}{Dina Demner-Fushman}, {et~al\mbox{.}}} \bibinfo{year}{2024}\natexlab{}.
\newblock \showarticletitle{{Peer review of GPT-4 technical report and systems card}}.
\newblock \bibinfo{journal}{\emph{PLOS Digital Health}} \bibinfo{volume}{3}, \bibinfo{number}{1} (\bibinfo{date}{01} \bibinfo{year}{2024}), \bibinfo{pages}{1--15}.
\newblock


\bibitem[Geiger et~al\mbox{.}(2012)]%
        {geiger2012kitti}
\bibfield{author}{\bibinfo{person}{Andreas Geiger}, \bibinfo{person}{Philip Lenz}, {and} \bibinfo{person}{Raquel Urtasun}.} \bibinfo{year}{2012}\natexlab{}.
\newblock \showarticletitle{{Are we ready for autonomous driving? The KITTI vision benchmark suite}}. In \bibinfo{booktitle}{\emph{Proceedings of the 2012 IEEE Conference on Computer Vision and Pattern Recognition (CVPR)}} (Providence, RI, USA). \bibinfo{publisher}{Institute of Electrical and Electronics Engineers}, \bibinfo{pages}{3354--3361}.
\newblock


\bibitem[Geva et~al\mbox{.}(2021)]%
        {geva2021aristotle}
\bibfield{author}{\bibinfo{person}{Mor Geva}, \bibinfo{person}{Daniel Khashabi}, \bibinfo{person}{Elad Segal}, \bibinfo{person}{Tushar Khot}, \bibinfo{person}{Dan Roth}, {and} \bibinfo{person}{Jonathan Berant}.} \bibinfo{year}{2021}\natexlab{}.
\newblock \showarticletitle{{Did Aristotle Use a Laptop? A Question Answering Benchmark with Implicit Reasoning Strategies}}.
\newblock \bibinfo{journal}{\emph{Transactions of the Association for Computational Linguistics}}  \bibinfo{volume}{9} (\bibinfo{date}{04} \bibinfo{year}{2021}), \bibinfo{pages}{346--361}.
\newblock
\showISSN{2307-387X}


\bibitem[Gupta and Nau(1992)]%
        {gupta1992complexity}
\bibfield{author}{\bibinfo{person}{Naresh Gupta} {and} \bibinfo{person}{Dana~S. Nau}.} \bibinfo{year}{1992}\natexlab{}.
\newblock \showarticletitle{On the complexity of blocks-world planning}.
\newblock \bibinfo{journal}{\emph{Artificial Intelligence}} \bibinfo{volume}{56}, \bibinfo{number}{2} (\bibinfo{year}{1992}), \bibinfo{pages}{223--254}.
\newblock
\showISSN{0004-3702}


\bibitem[Han et~al\mbox{.}(2025)]%
        {han2025vfusion}
\bibfield{author}{\bibinfo{person}{Junlin Han}, \bibinfo{person}{Filippos Kokkinos}, {and} \bibinfo{person}{Philip Torr}.} \bibinfo{year}{2025}\natexlab{}.
\newblock \showarticletitle{{VFusion3D: Learning Scalable 3D Generative Models from Video Diffusion Models}}. In \bibinfo{booktitle}{\emph{Proceedings of the 18th European Conference on Computer Vision (ECCV)}} (Milan, Italy). \bibinfo{publisher}{Springer Nature Switzerland}, \bibinfo{pages}{333--350}.
\newblock


\bibitem[Hanna and Bojar(2021)]%
        {hanna2021fine}
\bibfield{author}{\bibinfo{person}{Michael Hanna} {and} \bibinfo{person}{Ond{\v{r}}ej Bojar}.} \bibinfo{year}{2021}\natexlab{}.
\newblock \showarticletitle{{A Fine-Grained Analysis of {BERTS}core}}. In \bibinfo{booktitle}{\emph{Proceedings of the Sixth Conference on Machine Translation}} (Virtual). \bibinfo{publisher}{Association for Computational Linguistics}, \bibinfo{address}{Online}, \bibinfo{pages}{507--517}.
\newblock


\bibitem[Hayes-Roth(1985)]%
        {hayes1985rule}
\bibfield{author}{\bibinfo{person}{Frederick Hayes-Roth}.} \bibinfo{year}{1985}\natexlab{}.
\newblock \showarticletitle{Rule-based systems}.
\newblock \bibinfo{journal}{\emph{Commun. ACM}} \bibinfo{volume}{28}, \bibinfo{number}{9} (\bibinfo{date}{sep} \bibinfo{year}{1985}), \bibinfo{pages}{921–932}.
\newblock
\showISSN{0001-0782}


\bibitem[Hazra et~al\mbox{.}(2024)]%
        {hazra2024saycanpay}
\bibfield{author}{\bibinfo{person}{Rishi Hazra}, \bibinfo{person}{Pedro Zuidberg Dos~Martires}, {and} \bibinfo{person}{Luc De~Raedt}.} \bibinfo{year}{2024}\natexlab{}.
\newblock \showarticletitle{{SayCanPay: Heuristic Planning with Large Language Models Using Learnable Domain Knowledge}}. In \bibinfo{booktitle}{\emph{Proceedings of the 38th AAAI Conference on Artificial Intelligence}} (Vancouver, Canada), Vol.~\bibinfo{volume}{38}. \bibinfo{publisher}{AAAI Press}, \bibinfo{pages}{20123--20133}.
\newblock


\bibitem[He et~al\mbox{.}(2024)]%
        {he2024bridging}
\bibfield{author}{\bibinfo{person}{Haoran He}, \bibinfo{person}{Peilin Wu}, \bibinfo{person}{Chenjia Bai}, \bibinfo{person}{Hang Lai}, \bibinfo{person}{Lingxiao Wang}, \bibinfo{person}{Ling Pan}, \bibinfo{person}{Xiaolin Hu}, {and} \bibinfo{person}{Weinan Zhang}.} \bibinfo{year}{2024}\natexlab{}.
\newblock \showarticletitle{{Bridging the Sim-to-Real Gap from the Information Bottleneck Perspective}}. In \bibinfo{booktitle}{\emph{8th Annual Conference on Robot Learning}} (Munich, Germany).
\newblock


\bibitem[Henderson et~al\mbox{.}(2018)]%
        {henderson2018deep}
\bibfield{author}{\bibinfo{person}{Peter Henderson}, \bibinfo{person}{Riashat Islam}, \bibinfo{person}{Philip Bachman}, \bibinfo{person}{Joelle Pineau}, \bibinfo{person}{Doina Precup}, {and} \bibinfo{person}{David Meger}.} \bibinfo{year}{2018}\natexlab{}.
\newblock \showarticletitle{{Deep Reinforcement Learning That Matters}}. In \bibinfo{booktitle}{\emph{Proceedings of the 32nd AAAI Conference on Artificial Intelligence}} (New Orleans, LA, USA). \bibinfo{publisher}{AAAI Press}, Article \bibinfo{articleno}{392}, \bibinfo{numpages}{8}~pages.
\newblock
\showISBNx{978-1-57735-800-8}


\bibitem[Hendrycks et~al\mbox{.}(2021)]%
        {hendrycks2021measuring}
\bibfield{author}{\bibinfo{person}{Dan Hendrycks}, \bibinfo{person}{Collin Burns}, \bibinfo{person}{Steven Basart}, \bibinfo{person}{Andy Zou}, \bibinfo{person}{Mantas Mazeika}, \bibinfo{person}{Dawn Song}, {and} \bibinfo{person}{Jacob Steinhardt}.} \bibinfo{year}{2021}\natexlab{}.
\newblock \showarticletitle{{Measuring Massive Multitask Language Understanding}}. In \bibinfo{booktitle}{\emph{Proceedings of the 9th International Conference on Learning Representations}} (Virtual).
\newblock


\bibitem[Hermann et~al\mbox{.}(2015)]%
        {hermann2015teaching}
\bibfield{author}{\bibinfo{person}{Karl~Moritz Hermann}, \bibinfo{person}{Tomas Kocisky}, \bibinfo{person}{Edward Grefenstette}, \bibinfo{person}{Lasse Espeholt}, \bibinfo{person}{Will Kay}, \bibinfo{person}{Mustafa Suleyman}, {and} \bibinfo{person}{Phil Blunsom}.} \bibinfo{year}{2015}\natexlab{}.
\newblock \showarticletitle{{Teaching Machines to Read and Comprehend}}. In \bibinfo{booktitle}{\emph{Proceedings of the 2015 Conference on Neural Information Processing Systems}} (Montreal, Canada). \bibinfo{publisher}{Curran Associates, Inc.}
\newblock


\bibitem[Hessel et~al\mbox{.}(2021)]%
        {hessel2021clipscore}
\bibfield{author}{\bibinfo{person}{Jack Hessel}, \bibinfo{person}{Ari Holtzman}, \bibinfo{person}{Maxwell Forbes}, \bibinfo{person}{Ronan Le~Bras}, {and} \bibinfo{person}{Yejin Choi}.} \bibinfo{year}{2021}\natexlab{}.
\newblock \showarticletitle{{{CLIPS}core: A Reference-free Evaluation Metric for Image Captioning}}. In \bibinfo{booktitle}{\emph{Proceedings of the 2021 Conference on Empirical Methods in Natural Language Processing}} (Punta Cana, Dominican Republic). \bibinfo{publisher}{Association for Computational Linguistics}, \bibinfo{pages}{7514--7528}.
\newblock


\bibitem[Hu et~al\mbox{.}(2021)]%
        {hu2021model}
\bibfield{author}{\bibinfo{person}{Xia Hu}, \bibinfo{person}{Lingyang Chu}, \bibinfo{person}{Jian Pei}, \bibinfo{person}{Weiqing Liu}, {and} \bibinfo{person}{Jiang Bian}.} \bibinfo{year}{2021}\natexlab{}.
\newblock \showarticletitle{{Model complexity of deep learning: a survey}}.
\newblock \bibinfo{journal}{\emph{Knowledge and Information Systems}} \bibinfo{volume}{63}, \bibinfo{number}{10} (\bibinfo{date}{Oct.} \bibinfo{year}{2021}), \bibinfo{pages}{2585--2619}.
\newblock


\bibitem[Hu et~al\mbox{.}(2024)]%
        {hu2024deploying}
\bibfield{author}{\bibinfo{person}{Zichao Hu}, \bibinfo{person}{Francesca Lucchetti}, \bibinfo{person}{Claire Schlesinger}, \bibinfo{person}{Yash Saxena}, \bibinfo{person}{Anders Freeman}, \bibinfo{person}{Sadanand Modak}, \bibinfo{person}{Arjun Guha}, {and} \bibinfo{person}{Joydeep Biswas}.} \bibinfo{year}{2024}\natexlab{}.
\newblock \showarticletitle{{Deploying and Evaluating LLMs to Program Service Mobile Robots}}.
\newblock \bibinfo{journal}{\emph{IEEE Robotics and Automation Letters}} \bibinfo{volume}{9}, \bibinfo{number}{3} (\bibinfo{year}{2024}), \bibinfo{pages}{2853--2860}.
\newblock


\bibitem[Huang et~al\mbox{.}(2023)]%
        {huang2023survey}
\bibfield{author}{\bibinfo{person}{Lei Huang}, \bibinfo{person}{Weijiang Yu}, \bibinfo{person}{Weitao Ma}, \bibinfo{person}{Weihong Zhong}, \bibinfo{person}{Zhangyin Feng}, \bibinfo{person}{Haotian Wang}, \bibinfo{person}{Qianglong Chen}, \bibinfo{person}{Weihua Peng}, \bibinfo{person}{Xiaocheng Feng}, \bibinfo{person}{Bing Qin}, {et~al\mbox{.}}} \bibinfo{year}{2023}\natexlab{}.
\newblock \showarticletitle{{A Survey on Hallucination in Large Language Models: Principles, Taxonomy, Challenges, and Open Questions}}.
\newblock \bibinfo{journal}{\emph{arXiv preprint arXiv:2311.05232}} (\bibinfo{year}{2023}).
\newblock


\bibitem[Huang et~al\mbox{.}(2024a)]%
        {huang2024opera}
\bibfield{author}{\bibinfo{person}{Qidong Huang}, \bibinfo{person}{Xiaoyi Dong}, \bibinfo{person}{Pan Zhang}, \bibinfo{person}{Bin Wang}, \bibinfo{person}{Conghui He}, \bibinfo{person}{Jiaqi Wang}, \bibinfo{person}{Dahua Lin}, \bibinfo{person}{Weiming Zhang}, {and} \bibinfo{person}{Nenghai Yu}.} \bibinfo{year}{2024}\natexlab{a}.
\newblock \showarticletitle{{OPERA: Alleviating Hallucination in Multi-Modal Large Language Models via Over-Trust Penalty and Retrospection-Allocation}}. In \bibinfo{booktitle}{\emph{Proceedings of the 2024 IEEE/CVF Conference on Computer Vision and Pattern Recognition (CVPR)}} (Seattle, WA, USA). \bibinfo{publisher}{Institute of Electrical and Electronics Engineers}, \bibinfo{pages}{13418--13427}.
\newblock


\bibitem[Huang et~al\mbox{.}(2024c)]%
        {huang2024visual}
\bibfield{author}{\bibinfo{person}{Wen Huang}, \bibinfo{person}{Hongbin Liu}, \bibinfo{person}{Minxin Guo}, {and} \bibinfo{person}{Neil~Zhenqiang Gong}.} \bibinfo{year}{2024}\natexlab{c}.
\newblock \showarticletitle{{Visual Hallucinations of Multi-modal Large Language Models}}.
\newblock \bibinfo{journal}{\emph{arXiv preprint arXiv:2402.14683}} (\bibinfo{year}{2024}).
\newblock


\bibitem[Huang et~al\mbox{.}(2024b)]%
        {huang2024understanding}
\bibfield{author}{\bibinfo{person}{Xu Huang}, \bibinfo{person}{Weiwen Liu}, \bibinfo{person}{Xiaolong Chen}, \bibinfo{person}{Xingmei Wang}, \bibinfo{person}{Hao Wang}, \bibinfo{person}{Defu Lian}, \bibinfo{person}{Yasheng Wang}, \bibinfo{person}{Ruiming Tang}, {and} \bibinfo{person}{Enhong Chen}.} \bibinfo{year}{2024}\natexlab{b}.
\newblock \showarticletitle{{Understanding the planning of LLM agents: A survey}}.
\newblock \bibinfo{journal}{\emph{arXiv preprint arXiv:2402.02716}} (\bibinfo{year}{2024}).
\newblock


\bibitem[Hudson and Manning(2019)]%
        {hudson2019visual}
\bibfield{author}{\bibinfo{person}{Drew~A. Hudson} {and} \bibinfo{person}{Christopher~D. Manning}.} \bibinfo{year}{2019}\natexlab{}.
\newblock \showarticletitle{{GQA: A New Dataset for Real-World Visual Reasoning and Compositional Question Answering}}. In \bibinfo{booktitle}{\emph{Proceedings of the 2019 IEEE/CVF Conference on Computer Vision and Pattern Recognition (CVPR)}} (Long Beach, CA, USA). \bibinfo{publisher}{Institute of Electrical and Electronics Engineers}, \bibinfo{pages}{6693--6702}.
\newblock


\bibitem[Ichter et~al\mbox{.}(2023)]%
        {ichter2023do}
\bibfield{author}{\bibinfo{person}{Brian Ichter}, \bibinfo{person}{Anthony Brohan}, \bibinfo{person}{Yevgen Chebotar}, \bibinfo{person}{Chelsea Finn}, \bibinfo{person}{Karol Hausman}, \bibinfo{person}{Alexander Herzog}, \bibinfo{person}{Daniel Ho}, \bibinfo{person}{Julian Ibarz}, \bibinfo{person}{Alex Irpan}, \bibinfo{person}{Eric Jang}, {et~al\mbox{.}}} \bibinfo{year}{2023}\natexlab{}.
\newblock \showarticletitle{{Do As I Can, Not As I Say: Grounding Language in Robotic Affordances}}. In \bibinfo{booktitle}{\emph{Proceedings of The 6th Conference on Robot Learning}} (Atlanta, GA, USA) \emph{(\bibinfo{series}{Proceedings of Machine Learning Research}, Vol.~\bibinfo{volume}{205})}. \bibinfo{publisher}{PMLR}, \bibinfo{pages}{287--318}.
\newblock


\bibitem[Jain et~al\mbox{.}(2022)]%
        {jain2022zero}
\bibfield{author}{\bibinfo{person}{Ajay Jain}, \bibinfo{person}{Ben Mildenhall}, \bibinfo{person}{Jonathan~T. Barron}, \bibinfo{person}{Pieter Abbeel}, {and} \bibinfo{person}{Ben Poole}.} \bibinfo{year}{2022}\natexlab{}.
\newblock \showarticletitle{{Zero-Shot Text-Guided Object Generation with Dream Fields}}. In \bibinfo{booktitle}{\emph{Proceedings of the 2022 IEEE/CVF Conference on Computer Vision and Pattern Recognition (CVPR)}} (New Orleans, LA, USA). \bibinfo{publisher}{Institute of Electrical and Electronics Engineers}, \bibinfo{pages}{857--866}.
\newblock


\bibitem[Janai et~al\mbox{.}(2020)]%
        {janai2020computer}
\bibfield{author}{\bibinfo{person}{Joel Janai}, \bibinfo{person}{Fatma Güney}, \bibinfo{person}{Aseem Behl}, {and} \bibinfo{person}{Andreas Geiger}.} \bibinfo{year}{2020}\natexlab{}.
\newblock \showarticletitle{{Computer Vision for Autonomous Vehicles: Problems, Datasets and State of the Art}}.
\newblock \bibinfo{journal}{\emph{Foundations and Trends in Computer Graphics and Vision}} \bibinfo{volume}{12}, \bibinfo{number}{1-3} (\bibinfo{year}{2020}), \bibinfo{pages}{1--308}.
\newblock
\showISSN{1572-2740}


\bibitem[Jha et~al\mbox{.}(2023)]%
        {jha2023counterexample}
\bibfield{author}{\bibinfo{person}{Sumit~Kumar Jha}, \bibinfo{person}{Susmit Jha}, \bibinfo{person}{Patrick Lincoln}, \bibinfo{person}{Nathaniel~D. Bastian}, \bibinfo{person}{Alvaro Velasquez}, \bibinfo{person}{Rickard Ewetz}, {and} \bibinfo{person}{Sandeep Neema}.} \bibinfo{year}{2023}\natexlab{}.
\newblock \showarticletitle{{Counterexample Guided Inductive Synthesis Using Large Language Models and Satisfiability Solving}}. In \bibinfo{booktitle}{\emph{Proceedings of the 2023 IEEE Military Communications Conference (MILCOM)}} (Boston, MA, USA). \bibinfo{publisher}{Institute of Electrical and Electronics Engineers}, \bibinfo{pages}{944--949}.
\newblock


\bibitem[Ji et~al\mbox{.}(2023)]%
        {ji2023survey}
\bibfield{author}{\bibinfo{person}{Ziwei Ji}, \bibinfo{person}{Nayeon Lee}, \bibinfo{person}{Rita Frieske}, \bibinfo{person}{Tiezheng Yu}, \bibinfo{person}{Dan Su}, \bibinfo{person}{Yan Xu}, \bibinfo{person}{Etsuko Ishii}, \bibinfo{person}{Ye~Jin Bang}, \bibinfo{person}{Andrea Madotto}, {and} \bibinfo{person}{Pascale Fung}.} \bibinfo{year}{2023}\natexlab{}.
\newblock \showarticletitle{{Survey of Hallucination in Natural Language Generation}}.
\newblock \bibinfo{journal}{\emph{Comput. Surveys}} \bibinfo{volume}{55}, \bibinfo{number}{12}, Article \bibinfo{articleno}{248} (\bibinfo{date}{mar} \bibinfo{year}{2023}), \bibinfo{numpages}{38}~pages.
\newblock
\showISSN{0360-0300}


\bibitem[Johnson et~al\mbox{.}(2019)]%
        {johnson2019mimic}
\bibfield{author}{\bibinfo{person}{Alistair~EW Johnson}, \bibinfo{person}{Tom~J Pollard}, \bibinfo{person}{Seth~J Berkowitz}, \bibinfo{person}{Nathaniel~R Greenbaum}, \bibinfo{person}{Matthew~P Lungren}, \bibinfo{person}{Chih-ying Deng}, \bibinfo{person}{Roger~G Mark}, {and} \bibinfo{person}{Steven Horng}.} \bibinfo{year}{2019}\natexlab{}.
\newblock \showarticletitle{{MIMIC-CXR, a de-identified publicly available database of chest radiographs with free-text reports}}.
\newblock \bibinfo{journal}{\emph{Scientific Data}} \bibinfo{volume}{6}, \bibinfo{number}{1} (\bibinfo{year}{2019}), \bibinfo{pages}{317}.
\newblock


\bibitem[Joshi et~al\mbox{.}(2017)]%
        {joshi2017triviaqa}
\bibfield{author}{\bibinfo{person}{Mandar Joshi}, \bibinfo{person}{Eunsol Choi}, \bibinfo{person}{Daniel Weld}, {and} \bibinfo{person}{Luke Zettlemoyer}.} \bibinfo{year}{2017}\natexlab{}.
\newblock \showarticletitle{{{T}rivia{QA}: A Large Scale Distantly Supervised Challenge Dataset for Reading Comprehension}}. In \bibinfo{booktitle}{\emph{Proceedings of the 55th Annual Meeting of the Association for Computational Linguistics}} (Vancouver, Canada). \bibinfo{publisher}{Association for Computational Linguistics}, \bibinfo{pages}{1601--1611}.
\newblock


\bibitem[Kang and Liu(2024)]%
        {kang2024deficiency}
\bibfield{author}{\bibinfo{person}{Haoqiang Kang} {and} \bibinfo{person}{Xiao-Yang Liu}.} \bibinfo{year}{2024}\natexlab{}.
\newblock \showarticletitle{{Deficiency of Large Language Models in Finance: An Empirical Examination of Hallucination}}. In \bibinfo{booktitle}{\emph{I Can't Believe It's Not Better Workshop: Failure Modes in the Age of Foundation Models}}.
\newblock


\bibitem[Kasai et~al\mbox{.}(2023)]%
        {kasai2023realtime}
\bibfield{author}{\bibinfo{person}{Jungo Kasai}, \bibinfo{person}{Keisuke Sakaguchi}, \bibinfo{person}{Yoichi Takahashi}, \bibinfo{person}{Ronan~Le Bras}, \bibinfo{person}{Akari Asai}, \bibinfo{person}{Xinyan~Velocity Yu}, \bibinfo{person}{Dragomir Radev}, \bibinfo{person}{Noah~A. Smith}, \bibinfo{person}{Yejin Choi}, {and} \bibinfo{person}{Kentaro Inui}.} \bibinfo{year}{2023}\natexlab{}.
\newblock \showarticletitle{{RealTime {QA}: What's the Answer Right Now?}}. In \bibinfo{booktitle}{\emph{Proceedings of the 37th Conference on Neural Information Processing Systems Datasets and Benchmarks Track}} (New Orleans, LA, USA).
\newblock


\bibitem[Kemker et~al\mbox{.}(2018)]%
        {kemker2018measuring}
\bibfield{author}{\bibinfo{person}{Ronald Kemker}, \bibinfo{person}{Marc McClure}, \bibinfo{person}{Angelina Abitino}, \bibinfo{person}{Tyler Hayes}, {and} \bibinfo{person}{Christopher Kanan}.} \bibinfo{year}{2018}\natexlab{}.
\newblock \showarticletitle{{Measuring Catastrophic Forgetting in Neural Networks}}. In \bibinfo{booktitle}{\emph{Proceedings of the 32nd AAAI Conference on Artificial Intelligence}} (New Orleans, LA, USA). \bibinfo{publisher}{AAAI Press}, \bibinfo{pages}{3390--3398}.
\newblock


\bibitem[Keysan et~al\mbox{.}(2023)]%
        {keysan2023can}
\bibfield{author}{\bibinfo{person}{Ali Keysan}, \bibinfo{person}{Andreas Look}, \bibinfo{person}{Eitan Kosman}, \bibinfo{person}{Gonca Gürsun}, \bibinfo{person}{Jörg Wagner}, \bibinfo{person}{Yu Yao}, {and} \bibinfo{person}{Barbara Rakitsch}.} \bibinfo{year}{2023}\natexlab{}.
\newblock \showarticletitle{{Can you text what is happening? Integrating pre-trained language encoders into trajectory prediction models for autonomous driving}}.
\newblock \bibinfo{journal}{\emph{arXiv preprint arXiv:2309.05282}} (\bibinfo{year}{2023}).
\newblock


\bibitem[Kim et~al\mbox{.}(2018)]%
        {kim2018textual}
\bibfield{author}{\bibinfo{person}{Jinkyu Kim}, \bibinfo{person}{Anna Rohrbach}, \bibinfo{person}{Trevor Darrell}, \bibinfo{person}{John Canny}, {and} \bibinfo{person}{Zeynep Akata}.} \bibinfo{year}{2018}\natexlab{}.
\newblock \showarticletitle{{Textual Explanations for Self-Driving Vehicles}}. In \bibinfo{booktitle}{\emph{Proceedings of the 15th European Conference on Computer Vision (ECCV)}} (Munich, Germany). \bibinfo{publisher}{Springer International Publishing}, \bibinfo{pages}{577--593}.
\newblock
\showISBNx{978-3-030-01216-8}


\bibitem[Kim et~al\mbox{.}(2023)]%
        {kim2023task}
\bibfield{author}{\bibinfo{person}{Seokhwan Kim}, \bibinfo{person}{Spandana Gella}, \bibinfo{person}{Chao Zhao}, \bibinfo{person}{Di Jin}, \bibinfo{person}{Alexandros Papangelis}, \bibinfo{person}{Behnam Hedayatnia}, \bibinfo{person}{Yang Liu}, {and} \bibinfo{person}{Dilek Z~Hakkani-Tur}.} \bibinfo{year}{2023}\natexlab{}.
\newblock \showarticletitle{{Task-Oriented Conversational Modeling with Subjective Knowledge Track in {DSTC}11}}. In \bibinfo{booktitle}{\emph{Proceedings of The Eleventh Dialog System Technology Challenge}} (Prague, Czech Republic). \bibinfo{publisher}{Association for Computational Linguistics}, \bibinfo{pages}{274--281}.
\newblock


\bibitem[Kirillov et~al\mbox{.}(2023)]%
        {kirillov2023segment}
\bibfield{author}{\bibinfo{person}{Alexander Kirillov}, \bibinfo{person}{Eric Mintun}, \bibinfo{person}{Nikhila Ravi}, \bibinfo{person}{Hanzi Mao}, \bibinfo{person}{Chloe Rolland}, \bibinfo{person}{Laura Gustafson}, \bibinfo{person}{Tete Xiao}, \bibinfo{person}{Spencer Whitehead}, \bibinfo{person}{Alexander~C. Berg}, \bibinfo{person}{Wan-Yen Lo}, {et~al\mbox{.}}} \bibinfo{year}{2023}\natexlab{}.
\newblock \showarticletitle{{Segment Anything}}. In \bibinfo{booktitle}{\emph{Proceedings of the 2023 IEEE/CVF International Conference on Computer Vision (ICCV)}} (Paris, France). \bibinfo{publisher}{Institute of Electrical and Electronics Engineers}, \bibinfo{pages}{3992--4003}.
\newblock


\bibitem[Krishna et~al\mbox{.}(2017)]%
        {krishna2017visual}
\bibfield{author}{\bibinfo{person}{Ranjay Krishna}, \bibinfo{person}{Yuke Zhu}, \bibinfo{person}{Oliver Groth}, \bibinfo{person}{Justin Johnson}, \bibinfo{person}{Kenji Hata}, \bibinfo{person}{Joshua Kravitz}, \bibinfo{person}{Stephanie Chen}, \bibinfo{person}{Yannis Kalantidis}, \bibinfo{person}{Li-Jia Li}, \bibinfo{person}{David~A Shamma}, {et~al\mbox{.}}} \bibinfo{year}{2017}\natexlab{}.
\newblock \showarticletitle{{Visual Genome: Connecting Language and Vision Using Crowdsourced Dense Image Annotations}}.
\newblock \bibinfo{journal}{\emph{International Journal of Computer Vision}}  \bibinfo{volume}{123} (\bibinfo{year}{2017}), \bibinfo{pages}{32--73}.
\newblock


\bibitem[Kuhn et~al\mbox{.}(2023)]%
        {kuhn2023semantic}
\bibfield{author}{\bibinfo{person}{Lorenz Kuhn}, \bibinfo{person}{Yarin Gal}, {and} \bibinfo{person}{Sebastian Farquhar}.} \bibinfo{year}{2023}\natexlab{}.
\newblock \showarticletitle{{Semantic Uncertainty: Linguistic Invariances for Uncertainty Estimation in Natural Language Generation}}. In \bibinfo{booktitle}{\emph{Proceedings of the 11th International Conference on Learning Representations}} (Kigali, Rwanda).
\newblock


\bibitem[Kumar et~al\mbox{.}(2023)]%
        {kumar2023conformal}
\bibfield{author}{\bibinfo{person}{Bhawesh Kumar}, \bibinfo{person}{Charlie Lu}, \bibinfo{person}{Gauri Gupta}, \bibinfo{person}{Anil Palepu}, \bibinfo{person}{David Bellamy}, \bibinfo{person}{Ramesh Raskar}, {and} \bibinfo{person}{Andrew Beam}.} \bibinfo{year}{2023}\natexlab{}.
\newblock \showarticletitle{Conformal Prediction with Large Language Models for Multi-Choice Question Answering}. In \bibinfo{booktitle}{\emph{Proceedings of the `Neural Conversational AI Workshop - What’s left to TEACH (Trustworthy, Enhanced, Adaptable, Capable and Human-centric) chatbots?' at the 40th International Conference on Machine Learning}} (Honolulu, HI, USA).
\newblock


\bibitem[Kwiatkowski et~al\mbox{.}(2019)]%
        {kwiatkowski2019natural}
\bibfield{author}{\bibinfo{person}{Tom Kwiatkowski}, \bibinfo{person}{Jennimaria Palomaki}, \bibinfo{person}{Olivia Redfield}, \bibinfo{person}{Michael Collins}, \bibinfo{person}{Ankur Parikh}, \bibinfo{person}{Chris Alberti}, \bibinfo{person}{Danielle Epstein}, \bibinfo{person}{Illia Polosukhin}, \bibinfo{person}{Jacob Devlin}, \bibinfo{person}{Kenton Lee}, {et~al\mbox{.}}} \bibinfo{year}{2019}\natexlab{}.
\newblock \showarticletitle{{Natural Questions: A Benchmark for Question Answering Research}}.
\newblock \bibinfo{journal}{\emph{Transactions of the Association for Computational Linguistics}}  \bibinfo{volume}{7} (\bibinfo{year}{2019}), \bibinfo{pages}{452--466}.
\newblock


\bibitem[Kwon et~al\mbox{.}(2023)]%
        {kwon2023reward}
\bibfield{author}{\bibinfo{person}{Minae Kwon}, \bibinfo{person}{Sang~Michael Xie}, \bibinfo{person}{Kalesha Bullard}, {and} \bibinfo{person}{Dorsa Sadigh}.} \bibinfo{year}{2023}\natexlab{}.
\newblock \showarticletitle{{Reward Design with Language Models}}. In \bibinfo{booktitle}{\emph{Proceedings of the The 11th International Conference on Learning Representations}} (Kigali, Rwanda).
\newblock


\bibitem[Laban et~al\mbox{.}(2022)]%
        {laban2022summac}
\bibfield{author}{\bibinfo{person}{Philippe Laban}, \bibinfo{person}{Tobias Schnabel}, \bibinfo{person}{Paul~N. Bennett}, {and} \bibinfo{person}{Marti~A. Hearst}.} \bibinfo{year}{2022}\natexlab{}.
\newblock \showarticletitle{{{S}umma{C}: Re-Visiting {NLI}-based Models for Inconsistency Detection in Summarization}}.
\newblock \bibinfo{journal}{\emph{Transactions of the Association for Computational Linguistics}}  \bibinfo{volume}{10} (\bibinfo{year}{2022}), \bibinfo{pages}{163--177}.
\newblock


\bibitem[Lai et~al\mbox{.}(2023)]%
        {lai2023large}
\bibfield{author}{\bibinfo{person}{Jinqi Lai}, \bibinfo{person}{Wensheng Gan}, \bibinfo{person}{Jiayang Wu}, \bibinfo{person}{Zhenlian Qi}, {and} \bibinfo{person}{Philip~S Yu}.} \bibinfo{year}{2023}\natexlab{}.
\newblock \showarticletitle{{Large Language Models in Law: A Survey}}.
\newblock \bibinfo{journal}{\emph{arXiv preprint arXiv:2312.03718}} (\bibinfo{year}{2023}).
\newblock


\bibitem[Lebret et~al\mbox{.}(2016)]%
        {lebret2016neural}
\bibfield{author}{\bibinfo{person}{R{\'e}mi Lebret}, \bibinfo{person}{David Grangier}, {and} \bibinfo{person}{Michael Auli}.} \bibinfo{year}{2016}\natexlab{}.
\newblock \showarticletitle{{Neural Text Generation from Structured Data with Application to the Biography Domain}}. In \bibinfo{booktitle}{\emph{Proceedings of the 2016 Conference on Empirical Methods in Natural Language Processing}}. \bibinfo{publisher}{Association for Computational Linguistics}, \bibinfo{address}{Austin, TX, USA}, \bibinfo{pages}{1203--1213}.
\newblock


\bibitem[Lee et~al\mbox{.}(2023)]%
        {lee2023platypus}
\bibfield{author}{\bibinfo{person}{Ariel~N Lee}, \bibinfo{person}{Cole~J Hunter}, {and} \bibinfo{person}{Nataniel Ruiz}.} \bibinfo{year}{2023}\natexlab{}.
\newblock \showarticletitle{{Platypus: Quick, Cheap, and Powerful Refinement of LLMs}}.
\newblock \bibinfo{journal}{\emph{arXiv preprint arXiv:2308.07317}} (\bibinfo{year}{2023}).
\newblock


\bibitem[Lee et~al\mbox{.}(2020)]%
        {lee2020neural}
\bibfield{author}{\bibinfo{person}{Yoonho Lee}, \bibinfo{person}{Juho Lee}, \bibinfo{person}{Sung~Ju Hwang}, \bibinfo{person}{Eunho Yang}, {and} \bibinfo{person}{Seungjin Choi}.} \bibinfo{year}{2020}\natexlab{}.
\newblock \showarticletitle{{Neural Complexity Measures}}. In \bibinfo{booktitle}{\emph{Proceedings of the 2020 Conference on Neural Information Processing Systems}} (Virtual). \bibinfo{publisher}{Curran Associates, Inc.}, \bibinfo{pages}{9713--9724}.
\newblock


\bibitem[Lehmann et~al\mbox{.}(2015)]%
        {lehmann2015dbpedia}
\bibfield{author}{\bibinfo{person}{Jens Lehmann}, \bibinfo{person}{Robert Isele}, \bibinfo{person}{Max Jakob}, \bibinfo{person}{Anja Jentzsch}, \bibinfo{person}{Dimitris Kontokostas}, \bibinfo{person}{Pablo~N. Mendes}, \bibinfo{person}{Sebastian Hellmann}, \bibinfo{person}{Mohamed Morsey}, \bibinfo{person}{Patrick van Kleef}, \bibinfo{person}{S{\"o}ren Auer}, {and} \bibinfo{person}{Christian Bizer}.} \bibinfo{year}{2015}\natexlab{}.
\newblock \showarticletitle{{DBpedia -- A Large-scale, Multilingual Knowledge Base Extracted from Wikipedia}}.
\newblock \bibinfo{journal}{\emph{Semantic Web}}  \bibinfo{volume}{6} (\bibinfo{year}{2015}), \bibinfo{pages}{167--195}.
\newblock
\newblock
\shownote{2}.


\bibitem[Leurent(2018)]%
        {leurent2018highway}
\bibfield{author}{\bibinfo{person}{Edouard Leurent}.} \bibinfo{year}{2018}\natexlab{}.
\newblock \bibinfo{title}{An Environment for Autonomous Driving Decision-Making}.
\newblock \bibinfo{howpublished}{\url{https://github.com/eleurent/highway-env}}.
\newblock


\bibitem[Lewis et~al\mbox{.}(2020a)]%
        {lewis2020bart}
\bibfield{author}{\bibinfo{person}{Mike Lewis}, \bibinfo{person}{Yinhan Liu}, \bibinfo{person}{Naman Goyal}, \bibinfo{person}{Marjan Ghazvininejad}, \bibinfo{person}{Abdelrahman Mohamed}, \bibinfo{person}{Omer Levy}, \bibinfo{person}{Veselin Stoyanov}, {and} \bibinfo{person}{Luke Zettlemoyer}.} \bibinfo{year}{2020}\natexlab{a}.
\newblock \showarticletitle{{BART}: Denoising Sequence-to-Sequence Pre-training for Natural Language Generation, Translation, and Comprehension}. In \bibinfo{booktitle}{\emph{Proceedings of the 58th Annual Meeting of the Association for Computational Linguistics}} (Virtual). \bibinfo{publisher}{Association for Computational Linguistics}, \bibinfo{pages}{7871--7880}.
\newblock


\bibitem[Lewis et~al\mbox{.}(2020b)]%
        {lewis2020retrieval}
\bibfield{author}{\bibinfo{person}{Patrick Lewis}, \bibinfo{person}{Ethan Perez}, \bibinfo{person}{Aleksandra Piktus}, \bibinfo{person}{Fabio Petroni}, \bibinfo{person}{Vladimir Karpukhin}, \bibinfo{person}{Naman Goyal}, \bibinfo{person}{Heinrich K\"{u}ttler}, \bibinfo{person}{Mike Lewis}, \bibinfo{person}{Wen-tau Yih}, \bibinfo{person}{Tim Rockt\"{a}schel}, {et~al\mbox{.}}} \bibinfo{year}{2020}\natexlab{b}.
\newblock \showarticletitle{{Retrieval-Augmented Generation for Knowledge-Intensive NLP Tasks}}. In \bibinfo{booktitle}{\emph{Proceedings of the 2020 Conference on Neural Information Processing Systems}} (Virtual). \bibinfo{publisher}{Curran Associates, Inc.}, \bibinfo{pages}{9459--9474}.
\newblock


\bibitem[Li et~al\mbox{.}(2023e)]%
        {li2023generative}
\bibfield{author}{\bibinfo{person}{Chenghao Li}, \bibinfo{person}{Chaoning Zhang}, \bibinfo{person}{Joseph Cho}, \bibinfo{person}{Atish Waghwase}, \bibinfo{person}{Lik-Hang Lee}, \bibinfo{person}{Francois Rameau}, \bibinfo{person}{Yang Yang}, \bibinfo{person}{Sung-Ho Bae}, {and} \bibinfo{person}{Choong~Seon Hong}.} \bibinfo{year}{2023}\natexlab{e}.
\newblock \showarticletitle{{Generative AI meets 3D: A Survey on Text-to-3D in AIGC Era}}.
\newblock \bibinfo{journal}{\emph{arXiv preprint arXiv:2305.06131}} (\bibinfo{year}{2023}).
\newblock


\bibitem[Li et~al\mbox{.}(2023c)]%
        {li2023large}
\bibfield{author}{\bibinfo{person}{Daliang Li}, \bibinfo{person}{Ankit~Singh Rawat}, \bibinfo{person}{Manzil Zaheer}, \bibinfo{person}{Xin Wang}, \bibinfo{person}{Michal Lukasik}, \bibinfo{person}{Andreas Veit}, \bibinfo{person}{Felix Yu}, {and} \bibinfo{person}{Sanjiv Kumar}.} \bibinfo{year}{2023}\natexlab{c}.
\newblock \showarticletitle{{Large Language Models with Controllable Working Memory}}. In \bibinfo{booktitle}{\emph{Findings of the 61st Annual Meeting of the Association for Computational Linguistics}} (Toronto, Canada). \bibinfo{publisher}{Association for Computational Linguistics}, \bibinfo{pages}{1774--1793}.
\newblock


\bibitem[Li et~al\mbox{.}(2022)]%
        {li2022multispanqa}
\bibfield{author}{\bibinfo{person}{Haonan Li}, \bibinfo{person}{Martin Tomko}, \bibinfo{person}{Maria Vasardani}, {and} \bibinfo{person}{Timothy Baldwin}.} \bibinfo{year}{2022}\natexlab{}.
\newblock \showarticletitle{{{M}ulti{S}pan{QA}: A Dataset for Multi-Span Question Answering}}. In \bibinfo{booktitle}{\emph{Proceedings of the 2022 Conference of the North American Chapter of the Association for Computational Linguistics: Human Language Technologies}} (Seattle, WA, USA). \bibinfo{publisher}{Association for Computational Linguistics}, \bibinfo{pages}{1250--1260}.
\newblock


\bibitem[Li et~al\mbox{.}(2023a)]%
        {li2023halueval}
\bibfield{author}{\bibinfo{person}{Junyi Li}, \bibinfo{person}{Xiaoxue Cheng}, \bibinfo{person}{Xin Zhao}, \bibinfo{person}{Jian-Yun Nie}, {and} \bibinfo{person}{Ji-Rong Wen}.} \bibinfo{year}{2023}\natexlab{a}.
\newblock \showarticletitle{{{H}alu{E}val: A Large-Scale Hallucination Evaluation Benchmark for Large Language Models}}. In \bibinfo{booktitle}{\emph{Proceedings of the 2023 Conference on Empirical Methods in Natural Language Processing}} (Singapore). \bibinfo{publisher}{Association for Computational Linguistics}, \bibinfo{pages}{6449--6464}.
\newblock


\bibitem[Li et~al\mbox{.}(2024)]%
        {li2024chain}
\bibfield{author}{\bibinfo{person}{Xingxuan Li}, \bibinfo{person}{Ruochen Zhao}, \bibinfo{person}{Yew~Ken Chia}, \bibinfo{person}{Bosheng Ding}, \bibinfo{person}{Shafiq Joty}, \bibinfo{person}{Soujanya Poria}, {and} \bibinfo{person}{Lidong Bing}.} \bibinfo{year}{2024}\natexlab{}.
\newblock \showarticletitle{{Chain-of-Knowledge: Grounding Large Language Models via Dynamic Knowledge Adapting over Heterogeneous Sources}}. In \bibinfo{booktitle}{\emph{Proceedings of the 12th International Conference on Learning Representations}} (Vienna, Austria).
\newblock


\bibitem[Li et~al\mbox{.}(2023b)]%
        {li2023evaluating}
\bibfield{author}{\bibinfo{person}{Yifan Li}, \bibinfo{person}{Yifan Du}, \bibinfo{person}{Kun Zhou}, \bibinfo{person}{Jinpeng Wang}, \bibinfo{person}{Xin Zhao}, {and} \bibinfo{person}{Ji-Rong Wen}.} \bibinfo{year}{2023}\natexlab{b}.
\newblock \showarticletitle{{Evaluating Object Hallucination in Large Vision-Language Models}}. In \bibinfo{booktitle}{\emph{Proceedings of the 2023 Conference on Empirical Methods in Natural Language Processing}} (Singapore). \bibinfo{publisher}{Association for Computational Linguistics}, \bibinfo{pages}{292--305}.
\newblock


\bibitem[Li et~al\mbox{.}(2023d)]%
        {li2023language}
\bibfield{author}{\bibinfo{person}{Yinheng Li}, \bibinfo{person}{Shaofei Wang}, \bibinfo{person}{Han Ding}, {and} \bibinfo{person}{Hang Chen}.} \bibinfo{year}{2023}\natexlab{d}.
\newblock \showarticletitle{{Large Language Models in Finance: A Survey}}. In \bibinfo{booktitle}{\emph{Proceedings of the Fourth ACM International Conference on AI in Finance}} (Brooklyn, NY, USA). \bibinfo{publisher}{Association for Computing Machinery}, \bibinfo{address}{New York, NY, USA}, \bibinfo{pages}{374–382}.
\newblock
\showISBNx{9798400702402}


\bibitem[Liang et~al\mbox{.}(2023)]%
        {liang2023code}
\bibfield{author}{\bibinfo{person}{Jacky Liang}, \bibinfo{person}{Wenlong Huang}, \bibinfo{person}{Fei Xia}, \bibinfo{person}{Peng Xu}, \bibinfo{person}{Karol Hausman}, \bibinfo{person}{Brian Ichter}, \bibinfo{person}{Pete Florence}, {and} \bibinfo{person}{Andy Zeng}.} \bibinfo{year}{2023}\natexlab{}.
\newblock \showarticletitle{{Code as Policies: Language Model Programs for Embodied Control}}. In \bibinfo{booktitle}{\emph{Proceedings of the 2023 IEEE International Conference on Robotics and Automation (ICRA)}} (London, United Kingdom). \bibinfo{publisher}{Institute of Electrical and Electronics Engineers}, \bibinfo{pages}{9493--9500}.
\newblock


\bibitem[Liang et~al\mbox{.}(2024)]%
        {liang2024introspective}
\bibfield{author}{\bibinfo{person}{Kaiqu Liang}, \bibinfo{person}{Zixu Zhang}, {and} \bibinfo{person}{Jaime~Fernández Fisac}.} \bibinfo{year}{2024}\natexlab{}.
\newblock \showarticletitle{{Introspective Planning: Guiding Language-Enabled Agents to Refine Their Own Uncertainty}}.
\newblock \bibinfo{journal}{\emph{arXiv preprint arXiv:2402.06529}} (\bibinfo{year}{2024}).
\newblock


\bibitem[Lim and Shim(2024)]%
        {lim2024addressing}
\bibfield{author}{\bibinfo{person}{Youngsun Lim} {and} \bibinfo{person}{Hyunjung Shim}.} \bibinfo{year}{2024}\natexlab{}.
\newblock \showarticletitle{{Addressing Image Hallucination in Text-to-Image Generation through Factual Image Retrieval}}.
\newblock \bibinfo{journal}{\emph{arXiv preprint arXiv:2407.10683}} (\bibinfo{year}{2024}).
\newblock


\bibitem[Lin et~al\mbox{.}(2022a)]%
        {lin2022teaching}
\bibfield{author}{\bibinfo{person}{Stephanie Lin}, \bibinfo{person}{Jacob Hilton}, {and} \bibinfo{person}{Owain Evans}.} \bibinfo{year}{2022}\natexlab{a}.
\newblock \showarticletitle{Teaching Models to Express Their Uncertainty in Words}.
\newblock \bibinfo{journal}{\emph{Transactions on Machine Learning Research}} (\bibinfo{year}{2022}).
\newblock
\showISSN{2835-8856}


\bibitem[Lin et~al\mbox{.}(2022b)]%
        {lin2022truthful}
\bibfield{author}{\bibinfo{person}{Stephanie Lin}, \bibinfo{person}{Jacob Hilton}, {and} \bibinfo{person}{Owain Evans}.} \bibinfo{year}{2022}\natexlab{b}.
\newblock \showarticletitle{{{T}ruthful{QA}: Measuring How Models Mimic Human Falsehoods}}. In \bibinfo{booktitle}{\emph{Proceedings of the 60th Annual Meeting of the Association for Computational Linguistics}} (Dublin, Ireland). \bibinfo{publisher}{Association for Computational Linguistics}, \bibinfo{pages}{3214--3252}.
\newblock


\bibitem[Lin et~al\mbox{.}(2014)]%
        {lin2014mscoco}
\bibfield{author}{\bibinfo{person}{Tsung-Yi Lin}, \bibinfo{person}{Michael Maire}, \bibinfo{person}{Serge Belongie}, \bibinfo{person}{James Hays}, \bibinfo{person}{Pietro Perona}, \bibinfo{person}{Deva Ramanan}, \bibinfo{person}{Piotr Doll{\'a}r}, {and} \bibinfo{person}{C.~Lawrence Zitnick}.} \bibinfo{year}{2014}\natexlab{}.
\newblock \showarticletitle{{Microsoft COCO: Common Objects in Context}}. In \bibinfo{booktitle}{\emph{Proceedings of the 13th European Conference on Computer Vision (ECCV)}} (Zurich, Switzerland). \bibinfo{publisher}{Springer International Publishing}, \bibinfo{pages}{740--755}.
\newblock


\bibitem[Liu et~al\mbox{.}(2024b)]%
        {liu2024survey}
\bibfield{author}{\bibinfo{person}{Hanchao Liu}, \bibinfo{person}{Wenyuan Xue}, \bibinfo{person}{Yifei Chen}, \bibinfo{person}{Dapeng Chen}, \bibinfo{person}{Xiutian Zhao}, \bibinfo{person}{Ke Wang}, \bibinfo{person}{Liping Hou}, \bibinfo{person}{Rongjun Li}, {and} \bibinfo{person}{Wei Peng}.} \bibinfo{year}{2024}\natexlab{b}.
\newblock \showarticletitle{{A survey on hallucination in large vision-language models}}.
\newblock \bibinfo{journal}{\emph{arXiv preprint arXiv:2402.00253}} (\bibinfo{year}{2024}).
\newblock


\bibitem[Liu et~al\mbox{.}(2023)]%
        {liu2023mtd}
\bibfield{author}{\bibinfo{person}{Jiaqi Liu}, \bibinfo{person}{Peng Hang}, \bibinfo{person}{Xiao Qi}, \bibinfo{person}{Jianqiang Wang}, {and} \bibinfo{person}{Jian Sun}.} \bibinfo{year}{2023}\natexlab{}.
\newblock \showarticletitle{{MTD-GPT: A Multi-Task Decision-Making GPT Model for Autonomous Driving at Unsignalized Intersections}}. In \bibinfo{booktitle}{\emph{Proceedings of the 2023 IEEE International Conference on Intelligent Transportation Systems (ITSC)}} (Bilbao, Spain). \bibinfo{publisher}{Institute of Electrical and Electronics Engineers}, \bibinfo{pages}{5154--5161}.
\newblock


\bibitem[Liu et~al\mbox{.}(2024a)]%
        {liu2024comprehensive}
\bibfield{author}{\bibinfo{person}{Jian Liu}, \bibinfo{person}{Xiaoshui Huang}, \bibinfo{person}{Tianyu Huang}, \bibinfo{person}{Lu Chen}, \bibinfo{person}{Yuenan Hou}, \bibinfo{person}{Shixiang Tang}, \bibinfo{person}{Ziwei Liu}, \bibinfo{person}{Wanli Ouyang}, \bibinfo{person}{Wangmeng Zuo}, \bibinfo{person}{Junjun Jiang}, {and} \bibinfo{person}{Xianming Liu}.} \bibinfo{year}{2024}\natexlab{a}.
\newblock \showarticletitle{{A Comprehensive Survey on 3D Content Generation}}.
\newblock \bibinfo{journal}{\emph{arXiv preprint arXiv:2402.01166}} (\bibinfo{year}{2024}).
\newblock


\bibitem[Liu et~al\mbox{.}(2022)]%
        {liu2022token}
\bibfield{author}{\bibinfo{person}{Tianyu Liu}, \bibinfo{person}{Yizhe Zhang}, \bibinfo{person}{Chris Brockett}, \bibinfo{person}{Yi Mao}, \bibinfo{person}{Zhifang Sui}, \bibinfo{person}{Weizhu Chen}, {and} \bibinfo{person}{Bill Dolan}.} \bibinfo{year}{2022}\natexlab{}.
\newblock \showarticletitle{{A Token-level Reference-free Hallucination Detection Benchmark for Free-form Text Generation}}. In \bibinfo{booktitle}{\emph{Proceedings of the 60th Annual Meeting of the Association for Computational Linguistics}}. \bibinfo{publisher}{Association for Computational Linguistics}, \bibinfo{address}{Dublin, Ireland}, \bibinfo{pages}{6723--6737}.
\newblock


\bibitem[Lopez et~al\mbox{.}(2018)]%
        {lopez2018sumo}
\bibfield{author}{\bibinfo{person}{Pablo~Alvarez Lopez}, \bibinfo{person}{Michael Behrisch}, \bibinfo{person}{Laura Bieker-Walz}, \bibinfo{person}{Jakob Erdmann}, \bibinfo{person}{Yun-Pang Flötteröd}, \bibinfo{person}{Robert Hilbrich}, \bibinfo{person}{Leonhard Lücken}, \bibinfo{person}{Johannes Rummel}, \bibinfo{person}{Peter Wagner}, {and} \bibinfo{person}{Evamarie Wiessner}.} \bibinfo{year}{2018}\natexlab{}.
\newblock \showarticletitle{{Microscopic Traffic Simulation using SUMO}}. In \bibinfo{booktitle}{\emph{Proceedings of the 2018 International Conference on Intelligent Transportation Systems (ITSC)}} (Maui, HI, USA). \bibinfo{publisher}{Institute of Electrical and Electronics Engineers}, \bibinfo{pages}{2575--2582}.
\newblock


\bibitem[Lu et~al\mbox{.}(2022)]%
        {lu2022learn}
\bibfield{author}{\bibinfo{person}{Pan Lu}, \bibinfo{person}{Swaroop Mishra}, \bibinfo{person}{Tanglin Xia}, \bibinfo{person}{Liang Qiu}, \bibinfo{person}{Kai-Wei Chang}, \bibinfo{person}{Song-Chun Zhu}, \bibinfo{person}{Oyvind Tafjord}, \bibinfo{person}{Peter Clark}, {and} \bibinfo{person}{Ashwin Kalyan}.} \bibinfo{year}{2022}\natexlab{}.
\newblock \showarticletitle{{Learn to Explain: Multimodal Reasoning via Thought Chains for Science Question Answering}}. In \bibinfo{booktitle}{\emph{Proceedings of the 2022 Conference on Neural Information Processing Systems}} (New Orleans, LA, USA). \bibinfo{publisher}{Curran Associates, Inc.}, \bibinfo{pages}{2507--2521}.
\newblock


\bibitem[Lynch et~al\mbox{.}(2023)]%
        {lynch2023interactive}
\bibfield{author}{\bibinfo{person}{Corey Lynch}, \bibinfo{person}{Ayzaan Wahid}, \bibinfo{person}{Jonathan Tompson}, \bibinfo{person}{Tianli Ding}, \bibinfo{person}{James Betker}, \bibinfo{person}{Robert Baruch}, \bibinfo{person}{Travis Armstrong}, {and} \bibinfo{person}{Pete Florence}.} \bibinfo{year}{2023}\natexlab{}.
\newblock \showarticletitle{{Interactive Language: Talking to Robots in Real Time}}.
\newblock \bibinfo{journal}{\emph{IEEE Robotics and Automation Letters}} (\bibinfo{year}{2023}).
\newblock


\bibitem[Malaviya et~al\mbox{.}(2023)]%
        {malaviya2023quest}
\bibfield{author}{\bibinfo{person}{Chaitanya Malaviya}, \bibinfo{person}{Peter Shaw}, \bibinfo{person}{Ming-Wei Chang}, \bibinfo{person}{Kenton Lee}, {and} \bibinfo{person}{Kristina Toutanova}.} \bibinfo{year}{2023}\natexlab{}.
\newblock \showarticletitle{{{QUEST}: A Retrieval Dataset of Entity-Seeking Queries with Implicit Set Operations}}. In \bibinfo{booktitle}{\emph{Proceedings of the 61st Annual Meeting of the Association for Computational Linguistics}} (Toronto, Canada). \bibinfo{publisher}{Association for Computational Linguistics}, \bibinfo{pages}{14032--14047}.
\newblock


\bibitem[Malla et~al\mbox{.}(2023)]%
        {malla2023drama}
\bibfield{author}{\bibinfo{person}{Srikanth Malla}, \bibinfo{person}{Chiho Choi}, \bibinfo{person}{Isht Dwivedi}, \bibinfo{person}{Joon Hee~Choi}, {and} \bibinfo{person}{Jiachen Li}.} \bibinfo{year}{2023}\natexlab{}.
\newblock \showarticletitle{{DRAMA: Joint Risk Localization and Captioning in Driving}}. In \bibinfo{booktitle}{\emph{Proceedings of the 2023 IEEE/CVF Winter Conference on Applications of Computer Vision (WACV)}} (Waikola, HI, USA). \bibinfo{publisher}{Institute of Electrical and Electronics Engineers}, \bibinfo{pages}{1043--1052}.
\newblock


\bibitem[Manakul et~al\mbox{.}(2023)]%
        {manakul2023self}
\bibfield{author}{\bibinfo{person}{Potsawee Manakul}, \bibinfo{person}{Adian Liusie}, {and} \bibinfo{person}{Mark Gales}.} \bibinfo{year}{2023}\natexlab{}.
\newblock \showarticletitle{{{S}elf{C}heck{GPT}: Zero-Resource Black-Box Hallucination Detection for Generative Large Language Models}}. In \bibinfo{booktitle}{\emph{Proceedings of the 2023 Conference on Empirical Methods in Natural Language Processing}} (Singapore). \bibinfo{publisher}{Association for Computational Linguistics}, \bibinfo{pages}{9004--9017}.
\newblock


\bibitem[Mao et~al\mbox{.}(2024)]%
        {mao2023language}
\bibfield{author}{\bibinfo{person}{Jiageng Mao}, \bibinfo{person}{Junjie Ye}, \bibinfo{person}{Yuxi Qian}, \bibinfo{person}{Marco Pavone}, {and} \bibinfo{person}{Yue Wang}.} \bibinfo{year}{2024}\natexlab{}.
\newblock \showarticletitle{{A Language Agent for Autonomous Driving}}. In \bibinfo{booktitle}{\emph{Proceedings of the First Conference on Language Modeling}}.
\newblock


\bibitem[Mehrabi et~al\mbox{.}(2023)]%
        {mehrabi2023flirt}
\bibfield{author}{\bibinfo{person}{Ninareh Mehrabi}, \bibinfo{person}{Palash Goyal}, \bibinfo{person}{Christophe Dupuy}, \bibinfo{person}{Qian Hu}, \bibinfo{person}{Shalini Ghosh}, \bibinfo{person}{Richard Zemel}, \bibinfo{person}{Kai-Wei Chang}, \bibinfo{person}{Aram Galstyan}, {and} \bibinfo{person}{Rahul Gupta}.} \bibinfo{year}{2023}\natexlab{}.
\newblock \showarticletitle{{FLIRT: Feedback Loop In-context Red Teaming}}.
\newblock \bibinfo{journal}{\emph{arXiv preprint arXiv:2308.04265}} (\bibinfo{year}{2023}).
\newblock


\bibitem[Meng et~al\mbox{.}(2022)]%
        {meng2024locating}
\bibfield{author}{\bibinfo{person}{Kevin Meng}, \bibinfo{person}{David Bau}, \bibinfo{person}{Alex Andonian}, {and} \bibinfo{person}{Yonatan Belinkov}.} \bibinfo{year}{2022}\natexlab{}.
\newblock \showarticletitle{{Locating and Editing Factual Associations in GPT}}. In \bibinfo{booktitle}{\emph{Proceedings of the 2022 Conference on Neural Information Processing Systems}} (New Orleans, LA, USA), Vol.~\bibinfo{volume}{35}. \bibinfo{publisher}{Curran Associates, Inc.}, \bibinfo{pages}{17359--17372}.
\newblock


\bibitem[Meta(2024)]%
        {meta2024model}
\bibfield{author}{\bibinfo{person}{Meta}.} \bibinfo{year}{2024}\natexlab{}.
\newblock \showarticletitle{{Model Information}}.
\newblock \bibinfo{journal}{\emph{GitHub}} (\bibinfo{year}{2024}).
\newblock
\urldef\tempurl%
\url{https://github.com/meta-llama/llama-models/blob/main/models/llama3_1/MODEL_CARD.md}
\showURL{%
\tempurl}


\bibitem[Mialon et~al\mbox{.}(2023)]%
        {mialon2023augmented}
\bibfield{author}{\bibinfo{person}{Gr{\'e}goire Mialon}, \bibinfo{person}{Roberto Dessi}, \bibinfo{person}{Maria Lomeli}, \bibinfo{person}{Christoforos Nalmpantis}, \bibinfo{person}{Ramakanth Pasunuru}, \bibinfo{person}{Roberta Raileanu}, \bibinfo{person}{Baptiste Roziere}, \bibinfo{person}{Timo Schick}, \bibinfo{person}{Jane Dwivedi-Yu}, \bibinfo{person}{Asli Celikyilmaz}, {et~al\mbox{.}}} \bibinfo{year}{2023}\natexlab{}.
\newblock \showarticletitle{{Augmented Language Models: a Survey}}.
\newblock \bibinfo{journal}{\emph{Transactions on Machine Learning Research}} (\bibinfo{year}{2023}).
\newblock
\showISSN{2835-8856}
\newblock
\shownote{Survey Certification}.


\bibitem[Min et~al\mbox{.}(2023)]%
        {min2023factscore}
\bibfield{author}{\bibinfo{person}{Sewon Min}, \bibinfo{person}{Kalpesh Krishna}, \bibinfo{person}{Xinxi Lyu}, \bibinfo{person}{Mike Lewis}, \bibinfo{person}{Wen-tau Yih}, \bibinfo{person}{Pang Koh}, \bibinfo{person}{Mohit Iyyer}, \bibinfo{person}{Luke Zettlemoyer}, {and} \bibinfo{person}{Hannaneh Hajishirzi}.} \bibinfo{year}{2023}\natexlab{}.
\newblock \showarticletitle{{{FA}ct{S}core: Fine-grained Atomic Evaluation of Factual Precision in Long Form Text Generation}}. In \bibinfo{booktitle}{\emph{Proceedings of the 2023 Conference on Empirical Methods in Natural Language Processing}} (Singapore). \bibinfo{publisher}{Association for Computational Linguistics}, \bibinfo{pages}{12076--12100}.
\newblock


\bibitem[Mishkin et~al\mbox{.}(2022)]%
        {mishkin2022risks}
\bibfield{author}{\bibinfo{person}{Pamela Mishkin}, \bibinfo{person}{Lama Ahmad}, \bibinfo{person}{Miles Brundage}, \bibinfo{person}{Gretchen Krueger}, {and} \bibinfo{person}{Girish Sastry}.} \bibinfo{year}{2022}\natexlab{}.
\newblock \showarticletitle{{DALL·E 2 Preview - Risks and Limitations}}.
\newblock \bibinfo{journal}{\emph{GitHub}} (\bibinfo{year}{2022}).
\newblock
\urldef\tempurl%
\url{https://github.com/openai/dalle-2-preview/blob/main/system-card.md}
\showURL{%
\tempurl}


\bibitem[M{\"u}ndler et~al\mbox{.}(2024)]%
        {mundler2024self}
\bibfield{author}{\bibinfo{person}{Niels M{\"u}ndler}, \bibinfo{person}{Jingxuan He}, \bibinfo{person}{Slobodan Jenko}, {and} \bibinfo{person}{Martin Vechev}.} \bibinfo{year}{2024}\natexlab{}.
\newblock \showarticletitle{{Self-contradictory Hallucinations of Large Language Models: Evaluation, Detection and Mitigation}}. In \bibinfo{booktitle}{\emph{Proceedings of the 12th International Conference on Learning Representations}} (Vienna, Austria).
\newblock


\bibitem[Murray et~al\mbox{.}(2012)]%
        {murray2012ava}
\bibfield{author}{\bibinfo{person}{Naila Murray}, \bibinfo{person}{Luca Marchesotti}, {and} \bibinfo{person}{Florent Perronnin}.} \bibinfo{year}{2012}\natexlab{}.
\newblock \showarticletitle{{AVA: A Large-Scale Database for Aesthetic Visual Analysis}}. In \bibinfo{booktitle}{\emph{Proceedings of the 2012 IEEE Conference on Computer Vision and Pattern Recognition (CVPR)}} (Providence, RI, USA). \bibinfo{publisher}{Institute of Electrical and Electronics Engineers}, \bibinfo{pages}{2408--2415}.
\newblock


\bibitem[Omiye et~al\mbox{.}(2024)]%
        {omiye2024large}
\bibfield{author}{\bibinfo{person}{Jesutofunmi~A. Omiye}, \bibinfo{person}{Haiwen Gui}, \bibinfo{person}{Shawheen~J. Rezaei}, \bibinfo{person}{James Zou}, {and} \bibinfo{person}{Roxana Daneshjou}.} \bibinfo{year}{2024}\natexlab{}.
\newblock \showarticletitle{{Large Language Models in Medicine: The Potentials and Pitfalls}}.
\newblock \bibinfo{journal}{\emph{Annals of Internal Medicine}} \bibinfo{volume}{177}, \bibinfo{number}{2} (\bibinfo{date}{20 Feb} \bibinfo{year}{2024}), \bibinfo{pages}{210--220}.
\newblock
\showISSN{0003-4819}


\bibitem[OpenAI(2022)]%
        {openai2022chatgpt}
\bibfield{author}{\bibinfo{person}{OpenAI}.} \bibinfo{year}{2022}\natexlab{}.
\newblock \showarticletitle{{Introducing ChatGPT}}.
\newblock \bibinfo{journal}{\emph{OpenAI blog}} (\bibinfo{year}{2022}).
\newblock
\urldef\tempurl%
\url{https://openai.com/blog/chatgpt}
\showURL{%
\tempurl}


\bibitem[OpenAI(2023)]%
        {dalle32024system}
\bibfield{author}{\bibinfo{person}{OpenAI}.} \bibinfo{year}{2023}\natexlab{}.
\newblock \showarticletitle{{DALL·E 3 System Card}}.
\newblock \bibinfo{journal}{\emph{OpenAI blog}} (\bibinfo{year}{2023}).
\newblock
\urldef\tempurl%
\url{https://cdn.openai.com/papers/DALL_E_3_System_Card.pdf}
\showURL{%
\tempurl}


\bibitem[OpenAI(2024)]%
        {openai2024video}
\bibfield{author}{\bibinfo{person}{OpenAI}.} \bibinfo{year}{2024}\natexlab{}.
\newblock \showarticletitle{{Video generation models as world simulators}}.
\newblock \bibinfo{journal}{\emph{OpenAI blog}} (\bibinfo{year}{2024}).
\newblock
\urldef\tempurl%
\url{https://openai.com/index/video-generation-models-as-world-simulators/}
\showURL{%
\tempurl}


\bibitem[Oquab et~al\mbox{.}(2024)]%
        {oquab2024dinov2}
\bibfield{author}{\bibinfo{person}{Maxime Oquab}, \bibinfo{person}{Timoth{\'e}e Darcet}, \bibinfo{person}{Th{\'e}o Moutakanni}, \bibinfo{person}{Huy~V. Vo}, \bibinfo{person}{Marc Szafraniec}, \bibinfo{person}{Vasil Khalidov}, \bibinfo{person}{Pierre Fernandez}, \bibinfo{person}{Daniel HAZIZA}, \bibinfo{person}{Francisco Massa}, \bibinfo{person}{Alaaeldin El-Nouby}, {et~al\mbox{.}}} \bibinfo{year}{2024}\natexlab{}.
\newblock \showarticletitle{{{DINO}v2: Learning Robust Visual Features without Supervision}}.
\newblock \bibinfo{journal}{\emph{Transactions on Machine Learning Research}} (\bibinfo{year}{2024}).
\newblock
\showISSN{2835-8856}


\bibitem[Paa{\ss} and Giesselbach(2023)]%
        {paab2023foundation}
\bibfield{author}{\bibinfo{person}{Gerhard Paa{\ss}} {and} \bibinfo{person}{Sven Giesselbach}.} \bibinfo{year}{2023}\natexlab{}.
\newblock \bibinfo{booktitle}{\emph{{Foundation Models for Speech, Images, Videos, and Control}}}.
\newblock \bibinfo{publisher}{Springer International Publishing}, \bibinfo{address}{Cham}, \bibinfo{pages}{313--382}.
\newblock
\showISBNx{978-3-031-23190-2}


\bibitem[Pacchiardi et~al\mbox{.}(2024)]%
        {pacchiardi2024how}
\bibfield{author}{\bibinfo{person}{Lorenzo Pacchiardi}, \bibinfo{person}{Alex~James Chan}, \bibinfo{person}{S{\"o}ren Mindermann}, \bibinfo{person}{Ilan Moscovitz}, \bibinfo{person}{Alexa~Yue Pan}, \bibinfo{person}{Yarin Gal}, \bibinfo{person}{Owain Evans}, {and} \bibinfo{person}{Jan~M. Brauner}.} \bibinfo{year}{2024}\natexlab{}.
\newblock \showarticletitle{{How to Catch an {AI} Liar: Lie Detection in Black-Box {LLM}s by Asking Unrelated Questions}}. In \bibinfo{booktitle}{\emph{Proceedings of the 12th International Conference on Learning Representations}} (Vienna, Austria).
\newblock


\bibitem[Papineni et~al\mbox{.}(2002)]%
        {papineni2002bleu}
\bibfield{author}{\bibinfo{person}{Kishore Papineni}, \bibinfo{person}{Salim Roukos}, \bibinfo{person}{Todd Ward}, {and} \bibinfo{person}{Wei-Jing Zhu}.} \bibinfo{year}{2002}\natexlab{}.
\newblock \showarticletitle{{{BLEU}: a Method for Automatic Evaluation of Machine Translation}}. In \bibinfo{booktitle}{\emph{Proceedings of the 40th Annual Meeting of the Association for Computational Linguistics}} (Philadelphia, PA, USA). \bibinfo{publisher}{Association for Computational Linguistics}, \bibinfo{pages}{311--318}.
\newblock


\bibitem[Park et~al\mbox{.}(2024)]%
        {park2024clara}
\bibfield{author}{\bibinfo{person}{Jeongeun Park}, \bibinfo{person}{Seungwon Lim}, \bibinfo{person}{Joonhyung Lee}, \bibinfo{person}{Sangbeom Park}, \bibinfo{person}{Minsuk Chang}, \bibinfo{person}{Youngjae Yu}, {and} \bibinfo{person}{Sungjoon Choi}.} \bibinfo{year}{2024}\natexlab{}.
\newblock \showarticletitle{{CLARA: Classifying and Disambiguating User Commands for Reliable Interactive Robotic Agents}}.
\newblock \bibinfo{journal}{\emph{IEEE Robotics and Automation Letters}} \bibinfo{volume}{9}, \bibinfo{number}{2} (\bibinfo{year}{2024}), \bibinfo{pages}{1059--1066}.
\newblock


\bibitem[Park et~al\mbox{.}(2023)]%
        {park2023generative}
\bibfield{author}{\bibinfo{person}{Joon~Sung Park}, \bibinfo{person}{Joseph O'Brien}, \bibinfo{person}{Carrie~Jun Cai}, \bibinfo{person}{Meredith~Ringel Morris}, \bibinfo{person}{Percy Liang}, {and} \bibinfo{person}{Michael~S. Bernstein}.} \bibinfo{year}{2023}\natexlab{}.
\newblock \showarticletitle{{Generative Agents: Interactive Simulacra of Human Behavior}}. In \bibinfo{booktitle}{\emph{Proceedings of the 36th Annual ACM Symposium on User Interface Software and Technology}} (San Francisco, CA, USA). \bibinfo{publisher}{Association for Computing Machinery}, \bibinfo{address}{New York, NY, USA}, Article \bibinfo{articleno}{2}, \bibinfo{numpages}{22}~pages.
\newblock
\showISBNx{9798400701320}


\bibitem[Patel et~al\mbox{.}(2021)]%
        {patel2021nlp}
\bibfield{author}{\bibinfo{person}{Arkil Patel}, \bibinfo{person}{Satwik Bhattamishra}, {and} \bibinfo{person}{Navin Goyal}.} \bibinfo{year}{2021}\natexlab{}.
\newblock \showarticletitle{{Are {NLP} Models really able to Solve Simple Math Word Problems?}}. In \bibinfo{booktitle}{\emph{Proceedings of the 2021 Conference of the North American Chapter of the Association for Computational Linguistics: Human Language Technologies}} (Virtual). \bibinfo{publisher}{Association for Computational Linguistics}, \bibinfo{pages}{2080--2094}.
\newblock


\bibitem[Peng et~al\mbox{.}(2023)]%
        {peng2023check}
\bibfield{author}{\bibinfo{person}{Baolin Peng}, \bibinfo{person}{Michel Galley}, \bibinfo{person}{Pengcheng He}, \bibinfo{person}{Hao Cheng}, \bibinfo{person}{Yujia Xie}, \bibinfo{person}{Yu Hu}, \bibinfo{person}{Qiuyuan Huang}, \bibinfo{person}{Lars Liden}, \bibinfo{person}{Zhou Yu}, \bibinfo{person}{Weizhu Chen}, {and} \bibinfo{person}{Jianfeng Gao}.} \bibinfo{year}{2023}\natexlab{}.
\newblock \showarticletitle{{Check Your Facts and Try Again: Improving Large Language Models with External Knowledge and Automated Feedback}}.
\newblock \bibinfo{journal}{\emph{arXiv preprint arXiv:2302.12813}} (\bibinfo{year}{2023}).
\newblock


\bibitem[Perez et~al\mbox{.}(2023)]%
        {perez2023discovering}
\bibfield{author}{\bibinfo{person}{Ethan Perez}, \bibinfo{person}{Sam Ringer}, \bibinfo{person}{Kamile Lukosiute}, \bibinfo{person}{Karina Nguyen}, \bibinfo{person}{Edwin Chen}, \bibinfo{person}{Scott Heiner}, \bibinfo{person}{Craig Pettit}, \bibinfo{person}{Catherine Olsson}, \bibinfo{person}{Sandipan Kundu}, \bibinfo{person}{Saurav Kadavath}, {et~al\mbox{.}}} \bibinfo{year}{2023}\natexlab{}.
\newblock \showarticletitle{{Discovering Language Model Behaviors with Model-Written Evaluations}}. In \bibinfo{booktitle}{\emph{Findings of the 61st Annual Meeting of the Association for Computational Linguistics}} (Toronto, Canada). \bibinfo{publisher}{Association for Computational Linguistics}, \bibinfo{pages}{13387--13434}.
\newblock


\bibitem[Poggio et~al\mbox{.}(2020)]%
        {poggio2020theoretical}
\bibfield{author}{\bibinfo{person}{Tomaso Poggio}, \bibinfo{person}{Andrzej Banburski}, {and} \bibinfo{person}{Qianli Liao}.} \bibinfo{year}{2020}\natexlab{}.
\newblock \showarticletitle{{Theoretical issues in deep networks}}.
\newblock \bibinfo{journal}{\emph{Proceedings of the National Academy of Sciences}} \bibinfo{volume}{117}, \bibinfo{number}{48} (\bibinfo{year}{2020}), \bibinfo{pages}{30039--30045}.
\newblock


\bibitem[Poole et~al\mbox{.}(2023)]%
        {poole2023dreamfusion}
\bibfield{author}{\bibinfo{person}{Ben Poole}, \bibinfo{person}{Ajay Jain}, \bibinfo{person}{Jonathan~T. Barron}, {and} \bibinfo{person}{Ben Mildenhall}.} \bibinfo{year}{2023}\natexlab{}.
\newblock \showarticletitle{{DreamFusion: Text-to-3D using 2D Diffusion}}. In \bibinfo{booktitle}{\emph{Proceedings of the 11th International Conference on Learning Representations}} (Kigali, Rwanda).
\newblock


\bibitem[Puig et~al\mbox{.}(2018)]%
        {puig2018virtual}
\bibfield{author}{\bibinfo{person}{Xavier Puig}, \bibinfo{person}{Kevin Ra}, \bibinfo{person}{Marko Boben}, \bibinfo{person}{Jiaman Li}, \bibinfo{person}{Tingwu Wang}, \bibinfo{person}{Sanja Fidler}, {and} \bibinfo{person}{Antonio Torralba}.} \bibinfo{year}{2018}\natexlab{}.
\newblock \showarticletitle{VirtualHome: Simulating Household Activities Via Programs}. In \bibinfo{booktitle}{\emph{Proceedings of the 2018 IEEE/CVF Conference on Computer Vision and Pattern Recognition (CVPR)}} (Salt Lake City, UT, USA). \bibinfo{publisher}{Institute of Electrical and Electronics Engineers}, \bibinfo{pages}{8494--8502}.
\newblock


\bibitem[Puthumanaillam et~al\mbox{.}(2024)]%
        {puthumanaillam2024moral}
\bibfield{author}{\bibinfo{person}{Gokul Puthumanaillam}, \bibinfo{person}{Manav Vora}, \bibinfo{person}{Pranay Thangeda}, {and} \bibinfo{person}{Melkior Ornik}.} \bibinfo{year}{2024}\natexlab{}.
\newblock \showarticletitle{{A Moral Imperative: The Need for Continual Superalignment of Large Language Models}}.
\newblock \bibinfo{journal}{\emph{arXiv preprint arXiv:2403.14683}} (\bibinfo{year}{2024}).
\newblock


\bibitem[Qian et~al\mbox{.}(2024)]%
        {qian2024nuscenesqa}
\bibfield{author}{\bibinfo{person}{Tianwen Qian}, \bibinfo{person}{Jingjing Chen}, \bibinfo{person}{Linhai Zhuo}, \bibinfo{person}{Yang Jiao}, {and} \bibinfo{person}{Yu-Gang Jiang}.} \bibinfo{year}{2024}\natexlab{}.
\newblock \showarticletitle{{NuScenes-QA: A Multi-Modal Visual Question Answering Benchmark for Autonomous Driving Scenario}}. In \bibinfo{booktitle}{\emph{Proceedings of the 38th AAAI Conference on Artificial Intelligence}} (Vancouver, Canada), Vol.~\bibinfo{volume}{38}. \bibinfo{publisher}{AAAI Press}, \bibinfo{pages}{4542--4550}.
\newblock


\bibitem[Qiu et~al\mbox{.}(2023)]%
        {qiu2023latent}
\bibfield{author}{\bibinfo{person}{Huachuan Qiu}, \bibinfo{person}{Shuai Zhang}, \bibinfo{person}{Anqi Li}, \bibinfo{person}{Hongliang He}, {and} \bibinfo{person}{Zhenzhong Lan}.} \bibinfo{year}{2023}\natexlab{}.
\newblock \showarticletitle{{Latent Jailbreak: A Benchmark for Evaluating Text Safety and Output Robustness of Large Language Models}}.
\newblock \bibinfo{journal}{\emph{arXiv preprint arXiv:2307.08487}} (\bibinfo{year}{2023}).
\newblock


\bibitem[Quach et~al\mbox{.}(2024)]%
        {quach2024conformal}
\bibfield{author}{\bibinfo{person}{Victor Quach}, \bibinfo{person}{Adam Fisch}, \bibinfo{person}{Tal Schuster}, \bibinfo{person}{Adam Yala}, \bibinfo{person}{Jae~Ho Sohn}, \bibinfo{person}{Tommi~S. Jaakkola}, {and} \bibinfo{person}{Regina Barzilay}.} \bibinfo{year}{2024}\natexlab{}.
\newblock \showarticletitle{{Conformal Language Modeling}}. In \bibinfo{booktitle}{\emph{Proceedings of the 12th International Conference on Learning Representations}} (Vienna, Austria).
\newblock


\bibitem[Quiñonero-Candela et~al\mbox{.}(2008)]%
        {quinonero2008dataset}
\bibfield{author}{\bibinfo{person}{Joaquin Quiñonero-Candela}, \bibinfo{person}{Masashi Sugiyama}, \bibinfo{person}{Anton Schwaighofer}, {and} \bibinfo{person}{Neil~D. Lawrence}.} \bibinfo{year}{2008}\natexlab{}.
\newblock \bibinfo{booktitle}{\emph{{Dataset Shift in Machine Learning}}}.
\newblock \bibinfo{publisher}{The MIT Press}.
\newblock
\showISBNx{9780262255103}


\bibitem[Radford et~al\mbox{.}(2021)]%
        {radford2021learning}
\bibfield{author}{\bibinfo{person}{Alec Radford}, \bibinfo{person}{Jong~Wook Kim}, \bibinfo{person}{Chris Hallacy}, \bibinfo{person}{Aditya Ramesh}, \bibinfo{person}{Gabriel Goh}, \bibinfo{person}{Sandhini Agarwal}, \bibinfo{person}{Girish Sastry}, \bibinfo{person}{Amanda Askell}, \bibinfo{person}{Pamela Mishkin}, \bibinfo{person}{Jack Clark}, {et~al\mbox{.}}} \bibinfo{year}{2021}\natexlab{}.
\newblock \showarticletitle{{Learning Transferable Visual Models From Natural Language Supervision}}. In \bibinfo{booktitle}{\emph{Proceedings of the 38th International Conference on Machine Learning}} (Virtual) \emph{(\bibinfo{series}{Proceedings of Machine Learning Research}, Vol.~\bibinfo{volume}{139})}. \bibinfo{publisher}{PMLR}, \bibinfo{pages}{8748--8763}.
\newblock


\bibitem[Radford et~al\mbox{.}(2019)]%
        {radford2019language}
\bibfield{author}{\bibinfo{person}{Alec Radford}, \bibinfo{person}{Jeffrey Wu}, \bibinfo{person}{Rewon Child}, \bibinfo{person}{David Luan}, \bibinfo{person}{Dario Amodei}, \bibinfo{person}{Ilya Sutskever}, {et~al\mbox{.}}} \bibinfo{year}{2019}\natexlab{}.
\newblock \showarticletitle{Language models are unsupervised multitask learners}.
\newblock \bibinfo{journal}{\emph{OpenAI blog}} (\bibinfo{year}{2019}).
\newblock
\urldef\tempurl%
\url{https://openai.com/research/better-language-models}
\showURL{%
\tempurl}


\bibitem[Rajpurkar et~al\mbox{.}(2016)]%
        {rajpurkar2016squad}
\bibfield{author}{\bibinfo{person}{Pranav Rajpurkar}, \bibinfo{person}{Jian Zhang}, \bibinfo{person}{Konstantin Lopyrev}, {and} \bibinfo{person}{Percy Liang}.} \bibinfo{year}{2016}\natexlab{}.
\newblock \showarticletitle{{{SQ}u{AD}: 100,000+ Questions for Machine Comprehension of Text}}. In \bibinfo{booktitle}{\emph{Proceedings of the 2016 Conference on Empirical Methods in Natural Language Processing}} (Austin, TX, USA). \bibinfo{publisher}{Association for Computational Linguistics}, \bibinfo{pages}{2383--2392}.
\newblock


\bibitem[Ramakrishna et~al\mbox{.}(2023)]%
        {ramakrishna2023invite}
\bibfield{author}{\bibinfo{person}{Anil Ramakrishna}, \bibinfo{person}{Rahul Gupta}, \bibinfo{person}{Jens Lehmann}, {and} \bibinfo{person}{Morteza Ziyadi}.} \bibinfo{year}{2023}\natexlab{}.
\newblock \showarticletitle{{{INVITE}: a Testbed of Automatically Generated Invalid Questions to Evaluate Large Language Models for Hallucinations}}. In \bibinfo{booktitle}{\emph{Findings of the 2023 Conference on Empirical Methods in Natural Language Processing}} (Singapore). \bibinfo{publisher}{Association for Computational Linguistics}, \bibinfo{pages}{5422--5429}.
\newblock


\bibitem[Ramesh et~al\mbox{.}(2022)]%
        {ramesh2022hierarchical}
\bibfield{author}{\bibinfo{person}{Aditya Ramesh}, \bibinfo{person}{Prafulla Dhariwal}, \bibinfo{person}{Alex Nichol}, \bibinfo{person}{Casey Chu}, {and} \bibinfo{person}{Mark Chen}.} \bibinfo{year}{2022}\natexlab{}.
\newblock \showarticletitle{{Hierarchical Text-Conditional Image Generation with CLIP Latents}}.
\newblock \bibinfo{journal}{\emph{arXiv preprint arXiv:2204.06125}} (\bibinfo{year}{2022}).
\newblock


\bibitem[Rando et~al\mbox{.}(2022)]%
        {rando2022redteaming}
\bibfield{author}{\bibinfo{person}{Javier Rando}, \bibinfo{person}{Daniel Paleka}, \bibinfo{person}{David Lindner}, \bibinfo{person}{Lennart Heim}, {and} \bibinfo{person}{Florian Tramer}.} \bibinfo{year}{2022}\natexlab{}.
\newblock \showarticletitle{{Red-Teaming the Stable Diffusion Safety Filter}}. In \bibinfo{booktitle}{\emph{NeurIPS ML Safety Workshop}}.
\newblock


\bibitem[Rawte et~al\mbox{.}(2023)]%
        {rawte2023survey}
\bibfield{author}{\bibinfo{person}{Vipula Rawte}, \bibinfo{person}{Amit Sheth}, {and} \bibinfo{person}{Amitava Das}.} \bibinfo{year}{2023}\natexlab{}.
\newblock \showarticletitle{{A Survey of Hallucination in Large Foundation Models}}.
\newblock \bibinfo{journal}{\emph{arXiv preprint arXiv:2309.05922}} (\bibinfo{year}{2023}).
\newblock


\bibitem[Ren et~al\mbox{.}(2023)]%
        {ren2023robots}
\bibfield{author}{\bibinfo{person}{Allen~Z. Ren}, \bibinfo{person}{Anushri Dixit}, \bibinfo{person}{Alexandra Bodrova}, \bibinfo{person}{Sumeet Singh}, \bibinfo{person}{Stephen Tu}, \bibinfo{person}{Noah Brown}, \bibinfo{person}{Peng Xu}, \bibinfo{person}{Leila Takayama}, \bibinfo{person}{Fei Xia}, \bibinfo{person}{Jake Varley}, {et~al\mbox{.}}} \bibinfo{year}{2023}\natexlab{}.
\newblock \showarticletitle{{Robots That Ask For Help: Uncertainty Alignment for Large Language Model Planners}}. In \bibinfo{booktitle}{\emph{Proceedings of The 7th Conference on Robot Learning}} (Atlanta, GA, USA) \emph{(\bibinfo{series}{Proceedings of Machine Learning Research}, Vol.~\bibinfo{volume}{229})}. \bibinfo{publisher}{PMLR}, \bibinfo{pages}{661--682}.
\newblock


\bibitem[Rohrbach et~al\mbox{.}(2018)]%
        {rohrbach2018object}
\bibfield{author}{\bibinfo{person}{Anna Rohrbach}, \bibinfo{person}{Lisa~Anne Hendricks}, \bibinfo{person}{Kaylee Burns}, \bibinfo{person}{Trevor Darrell}, {and} \bibinfo{person}{Kate Saenko}.} \bibinfo{year}{2018}\natexlab{}.
\newblock \showarticletitle{{Object Hallucination in Image Captioning}}. In \bibinfo{booktitle}{\emph{Proceedings of the 2018 Conference on Empirical Methods in Natural Language Processing}} (Brussels, Belgium). \bibinfo{publisher}{Association for Computational Linguistics}, \bibinfo{pages}{4035--4045}.
\newblock


\bibitem[Rombach et~al\mbox{.}(2022)]%
        {rombach2022high}
\bibfield{author}{\bibinfo{person}{Robin Rombach}, \bibinfo{person}{Andreas Blattmann}, \bibinfo{person}{Dominik Lorenz}, \bibinfo{person}{Patrick Esser}, {and} \bibinfo{person}{Bj\"orn Ommer}.} \bibinfo{year}{2022}\natexlab{}.
\newblock \showarticletitle{{High-Resolution Image Synthesis With Latent Diffusion Models}}. In \bibinfo{booktitle}{\emph{Proceedings of the 2022 IEEE/CVF Conference on Computer Vision and Pattern Recognition (CVPR)}} (New Orleans, LA, USA). \bibinfo{publisher}{Institute of Electrical and Electronics Engineers}, \bibinfo{pages}{10684--10695}.
\newblock


\bibitem[Saharia et~al\mbox{.}(2022)]%
        {saharia2024photorealistic}
\bibfield{author}{\bibinfo{person}{Chitwan Saharia}, \bibinfo{person}{William Chan}, \bibinfo{person}{Saurabh Saxena}, \bibinfo{person}{Lala Li}, \bibinfo{person}{Jay Whang}, \bibinfo{person}{Emily~L Denton}, \bibinfo{person}{Kamyar Ghasemipour}, \bibinfo{person}{Raphael Gontijo~Lopes}, \bibinfo{person}{Burcu Karagol~Ayan}, \bibinfo{person}{Tim Salimans}, {et~al\mbox{.}}} \bibinfo{year}{2022}\natexlab{}.
\newblock \showarticletitle{{Photorealistic Text-to-Image Diffusion Models with Deep Language Understanding}}. In \bibinfo{booktitle}{\emph{Proceedings of the 2022 Conference on Neural Information Processing Systems}} (New Orleans, LA, USA). \bibinfo{publisher}{Curran Associates, Inc.}, \bibinfo{pages}{36479--36494}.
\newblock


\bibitem[Sahoo et~al\mbox{.}(2024)]%
        {sahoo2024comprehensive}
\bibfield{author}{\bibinfo{person}{Pranab Sahoo}, \bibinfo{person}{Prabhash Meharia}, \bibinfo{person}{Akash Ghosh}, \bibinfo{person}{Sriparna Saha}, \bibinfo{person}{Vinija Jain}, {and} \bibinfo{person}{Aman Chadha}.} \bibinfo{year}{2024}\natexlab{}.
\newblock \showarticletitle{{A Comprehensive Survey of Hallucination in Large Language, Image, Video and Audio Foundation Models}}. In \bibinfo{booktitle}{\emph{Findings of the 2024 Conference on Empirical Methods in Natural Language Processing}} (Miami, FL, USA). \bibinfo{publisher}{Association for Computational Linguistics}, \bibinfo{pages}{11709--11724}.
\newblock


\bibitem[Salvato et~al\mbox{.}(2021)]%
        {salvato2021crossing}
\bibfield{author}{\bibinfo{person}{Erica Salvato}, \bibinfo{person}{Gianfranco Fenu}, \bibinfo{person}{Eric Medvet}, {and} \bibinfo{person}{Felice~Andrea Pellegrino}.} \bibinfo{year}{2021}\natexlab{}.
\newblock \showarticletitle{{Crossing the Reality Gap: A Survey on Sim-to-Real Transferability of Robot Controllers in Reinforcement Learning}}.
\newblock \bibinfo{journal}{\emph{IEEE Access}}  \bibinfo{volume}{9} (\bibinfo{year}{2021}), \bibinfo{pages}{153171--153187}.
\newblock


\bibitem[Sanghi et~al\mbox{.}(2022)]%
        {sanghi2022clip}
\bibfield{author}{\bibinfo{person}{Aditya Sanghi}, \bibinfo{person}{Hang Chu}, \bibinfo{person}{Joseph~G. Lambourne}, \bibinfo{person}{Ye Wang}, \bibinfo{person}{Chin-Yi Cheng}, \bibinfo{person}{Marco Fumero}, {and} \bibinfo{person}{Kamal~Rahimi Malekshan}.} \bibinfo{year}{2022}\natexlab{}.
\newblock \showarticletitle{{CLIP-Forge: Towards Zero-Shot Text-to-Shape Generation}}. In \bibinfo{booktitle}{\emph{Proceedings of the 2022 IEEE/CVF Conference on Computer Vision and Pattern Recognition (CVPR)}} (New Orleans, LA, USA). \bibinfo{publisher}{Institute of Electrical and Electronics Engineers}, \bibinfo{pages}{18582--18592}.
\newblock


\bibitem[Sarmah et~al\mbox{.}(2024)]%
        {sarmah2024towards}
\bibfield{author}{\bibinfo{person}{Bhaskarjit Sarmah}, \bibinfo{person}{Dhagash Mehta}, \bibinfo{person}{Stefano Pasquali}, {and} \bibinfo{person}{Tianjie Zhu}.} \bibinfo{year}{2024}\natexlab{}.
\newblock \showarticletitle{{Towards reducing hallucination in extracting information from financial reports using Large Language Models}}. In \bibinfo{booktitle}{\emph{Proceedings of the Third International Conference on AI-ML Systems}} (Bangalore, India). \bibinfo{publisher}{Association for Computing Machinery}, \bibinfo{address}{New York, NY, USA}, Article \bibinfo{articleno}{39}, \bibinfo{numpages}{5}~pages.
\newblock
\showISBNx{9798400716492}


\bibitem[Schramowski et~al\mbox{.}(2023)]%
        {schramowski2023safe}
\bibfield{author}{\bibinfo{person}{Patrick Schramowski}, \bibinfo{person}{Manuel Brack}, \bibinfo{person}{Björn Deiseroth}, {and} \bibinfo{person}{Kristian Kersting}.} \bibinfo{year}{2023}\natexlab{}.
\newblock \showarticletitle{{Safe Latent Diffusion: Mitigating Inappropriate Degeneration in Diffusion Models}}. In \bibinfo{booktitle}{\emph{Proceedings of the 2023 IEEE/CVF Conference on Computer Vision and Pattern Recognition (CVPR)}} (Vancouver, Canada). \bibinfo{publisher}{Institute of Electrical and Electronics Engineers}, \bibinfo{pages}{22522--22531}.
\newblock


\bibitem[Schreiber et~al\mbox{.}(2023)]%
        {schreiber2023attentional}
\bibfield{author}{\bibinfo{person}{Andre Schreiber}, \bibinfo{person}{Tianchen Ji}, \bibinfo{person}{D.~Livingston McPherson}, {and} \bibinfo{person}{Katherine Driggs-Campbell}.} \bibinfo{year}{2023}\natexlab{}.
\newblock \showarticletitle{{An Attentional Recurrent Neural Network for Occlusion-Aware Proactive Anomaly Detection in Field Robot Navigation}}. In \bibinfo{booktitle}{\emph{Proceedings of the 2023 IEEE/RSJ International Conference on Intelligent Robots and Systems (IROS)}} (Detroit, MI, USA). \bibinfo{publisher}{Institute of Electrical and Electronics Engineers}, \bibinfo{pages}{8038--8045}.
\newblock


\bibitem[Schuhmann et~al\mbox{.}(2021)]%
        {schuhmann2021laion}
\bibfield{author}{\bibinfo{person}{Christoph Schuhmann}, \bibinfo{person}{Richard Vencu}, \bibinfo{person}{Romain Beaumont}, \bibinfo{person}{Robert Kaczmarczyk}, \bibinfo{person}{Clayton Mullis}, \bibinfo{person}{Aarush Katta}, \bibinfo{person}{Theo Coombes}, \bibinfo{person}{Jenia Jitsev}, {and} \bibinfo{person}{Aran Komatsuzaki}.} \bibinfo{year}{2021}\natexlab{}.
\newblock \showarticletitle{{LAION-400M: Open Dataset of CLIP-Filtered 400 Million Image-Text Pairs}}.
\newblock \bibinfo{journal}{\emph{arXiv preprint arXiv:2111.02114}} (\bibinfo{year}{2021}).
\newblock


\bibitem[Schwenk et~al\mbox{.}(2022)]%
        {schwenk2022aokvqa}
\bibfield{author}{\bibinfo{person}{Dustin Schwenk}, \bibinfo{person}{Apoorv Khandelwal}, \bibinfo{person}{Christopher Clark}, \bibinfo{person}{Kenneth Marino}, {and} \bibinfo{person}{Roozbeh Mottaghi}.} \bibinfo{year}{2022}\natexlab{}.
\newblock \showarticletitle{{A-OKVQA: A Benchmark for Visual Question Answering Using World Knowledge}}. In \bibinfo{booktitle}{\emph{Proceedings of the 17th European Conference on Computer Vision (ECCV)}} (Tel Aviv, Israel). \bibinfo{publisher}{Springer Nature Switzerland}, \bibinfo{pages}{146--162}.
\newblock
\showISBNx{978-3-031-20074-8}


\bibitem[Shafer and Vovk(2008)]%
        {shafer2008tutorial}
\bibfield{author}{\bibinfo{person}{Glenn Shafer} {and} \bibinfo{person}{Vladimir Vovk}.} \bibinfo{year}{2008}\natexlab{}.
\newblock \showarticletitle{{A Tutorial on Conformal Prediction}}.
\newblock \bibinfo{journal}{\emph{Journal of Machine Learning Research}} \bibinfo{volume}{9}, \bibinfo{number}{12} (\bibinfo{year}{2008}), \bibinfo{pages}{371--421}.
\newblock


\bibitem[Shah et~al\mbox{.}(2023)]%
        {shah2023lm}
\bibfield{author}{\bibinfo{person}{Dhruv Shah}, \bibinfo{person}{B\l{a}\.zej Osi\'nski}, \bibinfo{person}{Brian Ichter}, {and} \bibinfo{person}{Sergey Levine}.} \bibinfo{year}{2023}\natexlab{}.
\newblock \showarticletitle{{LM-Nav: Robotic Navigation with Large Pre-Trained Models of Language, Vision, and Action}}. In \bibinfo{booktitle}{\emph{Proceedings of The 6th Conference on Robot Learning}} (Atlanta, GA, USA) \emph{(\bibinfo{series}{Proceedings of Machine Learning Research}, Vol.~\bibinfo{volume}{205})}. \bibinfo{publisher}{PMLR}, \bibinfo{pages}{492--504}.
\newblock


\bibitem[Shinn et~al\mbox{.}(2023)]%
        {shinn2023reflexion}
\bibfield{author}{\bibinfo{person}{Noah Shinn}, \bibinfo{person}{Federico Cassano}, \bibinfo{person}{Ashwin Gopinath}, \bibinfo{person}{Karthik Narasimhan}, {and} \bibinfo{person}{Shunyu Yao}.} \bibinfo{year}{2023}\natexlab{}.
\newblock \showarticletitle{{Reflexion: language agents with verbal reinforcement learning}}. In \bibinfo{booktitle}{\emph{Proceedings of the 2023 Conference on Neural Information Processing Systems}} (New Orleans, LA, USA). \bibinfo{publisher}{Curran Associates, Inc.}, \bibinfo{pages}{8634--8652}.
\newblock


\bibitem[Shridhar et~al\mbox{.}(2020)]%
        {shridhar2020alfred}
\bibfield{author}{\bibinfo{person}{Mohit Shridhar}, \bibinfo{person}{Jesse Thomason}, \bibinfo{person}{Daniel Gordon}, \bibinfo{person}{Yonatan Bisk}, \bibinfo{person}{Winson Han}, \bibinfo{person}{Roozbeh Mottaghi}, \bibinfo{person}{Luke Zettlemoyer}, {and} \bibinfo{person}{Dieter Fox}.} \bibinfo{year}{2020}\natexlab{}.
\newblock \showarticletitle{{ALFRED: A Benchmark for Interpreting Grounded Instructions for Everyday Tasks}}. In \bibinfo{booktitle}{\emph{Proceedings of the 2020 IEEE/CVF Conference on Computer Vision and Pattern Recognition (CVPR)}} (Virtual). \bibinfo{publisher}{Institute of Electrical and Electronics Engineers}, \bibinfo{pages}{10737--10746}.
\newblock


\bibitem[Shridhar et~al\mbox{.}(2021)]%
        {shridhar2021alfworld}
\bibfield{author}{\bibinfo{person}{Mohit Shridhar}, \bibinfo{person}{Xingdi Yuan}, \bibinfo{person}{Marc-Alexandre Cote}, \bibinfo{person}{Yonatan Bisk}, \bibinfo{person}{Adam Trischler}, {and} \bibinfo{person}{Matthew Hausknecht}.} \bibinfo{year}{2021}\natexlab{}.
\newblock \showarticletitle{{ALFWorld: Aligning Text and Embodied Environments for Interactive Learning}}. In \bibinfo{booktitle}{\emph{Proceedings of the 9th International Conference on Learning Representations}} (Virtual).
\newblock


\bibitem[Sima et~al\mbox{.}(2023)]%
        {sima2023drivelm}
\bibfield{author}{\bibinfo{person}{Chonghao Sima}, \bibinfo{person}{Katrin Renz}, \bibinfo{person}{Kashyap Chitta}, \bibinfo{person}{Li Chen}, \bibinfo{person}{Hanxue Zhang}, \bibinfo{person}{Chengen Xie}, \bibinfo{person}{Ping Luo}, \bibinfo{person}{Andreas Geiger}, {and} \bibinfo{person}{Hongyang Li}.} \bibinfo{year}{2023}\natexlab{}.
\newblock \showarticletitle{{DriveLM: Driving with Graph Visual Question Answering}}.
\newblock \bibinfo{journal}{\emph{arXiv preprint arXiv:2312.14150}} (\bibinfo{year}{2023}).
\newblock


\bibitem[Singer and Cohen(2021)]%
        {singer2021framework}
\bibfield{author}{\bibinfo{person}{Gonen Singer} {and} \bibinfo{person}{Yuval Cohen}.} \bibinfo{year}{2021}\natexlab{}.
\newblock \showarticletitle{{A framework for smart control using machine-learning modeling for processes with closed-loop control in Industry 4.0}}.
\newblock \bibinfo{journal}{\emph{Engineering Applications of Artificial Intelligence}}  \bibinfo{volume}{102} (\bibinfo{year}{2021}), \bibinfo{pages}{104236}.
\newblock
\showISSN{0952-1976}


\bibitem[Singh et~al\mbox{.}(2023)]%
        {sing2023progprompt}
\bibfield{author}{\bibinfo{person}{Ishika Singh}, \bibinfo{person}{Valts Blukis}, \bibinfo{person}{Arsalan Mousavian}, \bibinfo{person}{Ankit Goyal}, \bibinfo{person}{Danfei Xu}, \bibinfo{person}{Jonathan Tremblay}, \bibinfo{person}{Dieter Fox}, \bibinfo{person}{Jesse Thomason}, {and} \bibinfo{person}{Animesh Garg}.} \bibinfo{year}{2023}\natexlab{}.
\newblock \showarticletitle{{ProgPrompt: Generating Situated Robot Task Plans using Large Language Models}}. In \bibinfo{booktitle}{\emph{Proceedings of the 2023 IEEE International Conference on Robotics and Automation (ICRA)}} (London, United Kingdom). \bibinfo{publisher}{Institute of Electrical and Electronics Engineers}, \bibinfo{pages}{11523--11530}.
\newblock


\bibitem[Song et~al\mbox{.}(2024)]%
        {song2023luna}
\bibfield{author}{\bibinfo{person}{Da Song}, \bibinfo{person}{Xuan Xie}, \bibinfo{person}{Jiayang Song}, \bibinfo{person}{Derui Zhu}, \bibinfo{person}{Yuheng Huang}, \bibinfo{person}{Felix Juefei-Xu}, {and} \bibinfo{person}{Lei Ma}.} \bibinfo{year}{2024}\natexlab{}.
\newblock \showarticletitle{{LUNA: A Model-Based Universal Analysis Framework for Large Language Models}}.
\newblock \bibinfo{journal}{\emph{IEEE Transactions on Software Engineering}} \bibinfo{volume}{50}, \bibinfo{number}{7} (\bibinfo{year}{2024}), \bibinfo{pages}{1921--1948}.
\newblock


\bibitem[Soori et~al\mbox{.}(2023)]%
        {soori2023artificial}
\bibfield{author}{\bibinfo{person}{Mohsen Soori}, \bibinfo{person}{Behrooz Arezoo}, {and} \bibinfo{person}{Roza Dastres}.} \bibinfo{year}{2023}\natexlab{}.
\newblock \showarticletitle{{Artificial intelligence, machine learning and deep learning in advanced robotics, a review}}.
\newblock \bibinfo{journal}{\emph{Cognitive Robotics}}  \bibinfo{volume}{3} (\bibinfo{year}{2023}), \bibinfo{pages}{54--70}.
\newblock
\showISSN{2667-2413}


\bibitem[Srivastava et~al\mbox{.}(2023)]%
        {srivastava2023beyond}
\bibfield{author}{\bibinfo{person}{Aarohi Srivastava}, \bibinfo{person}{Abhinav Rastogi}, \bibinfo{person}{Abhishek Rao}, \bibinfo{person}{Abu Awal~Md Shoeb}, \bibinfo{person}{Abubakar Abid}, \bibinfo{person}{Adam Fisch}, \bibinfo{person}{Adam~R. Brown}, \bibinfo{person}{Adam Santoro}, \bibinfo{person}{Aditya Gupta}, \bibinfo{person}{Adri{\`a} Garriga-Alonso}, {et~al\mbox{.}}} \bibinfo{year}{2023}\natexlab{}.
\newblock \showarticletitle{{Beyond the Imitation Game: Quantifying and extrapolating the capabilities of language models}}.
\newblock \bibinfo{journal}{\emph{Transactions on Machine Learning Research}} (\bibinfo{year}{2023}).
\newblock
\showISSN{2835-8856}


\bibitem[Stremple(2024)]%
        {stremple2024false}
\bibfield{author}{\bibinfo{person}{Claire Stremple}.} \bibinfo{year}{2024}\natexlab{}.
\newblock \showarticletitle{{False citations show Alaska education official relied on generative AI, raising broader questions}}.
\newblock \bibinfo{journal}{\emph{Alaska Beacon}} (\bibinfo{year}{2024}).
\newblock
\urldef\tempurl%
\url{https://alaskabeacon.com/2024/10/28/alaska-education-department-published-false-ai-generated-academic-citations-in-cell-policy-document/}
\showURL{%
\tempurl}


\bibitem[Suganuma and Yoneda(2022)]%
        {suganuma2022current}
\bibfield{author}{\bibinfo{person}{Naoki Suganuma} {and} \bibinfo{person}{Keisuke Yoneda}.} \bibinfo{year}{2022}\natexlab{}.
\newblock \showarticletitle{{Current Status and Issues of Traffic Light Recognition Technology in Autonomous Driving System}}.
\newblock \bibinfo{journal}{\emph{IEICE Transactions on Fundamentals of Electronics, Communications and Computer Sciences}} \bibinfo{volume}{E105.A}, \bibinfo{number}{5} (\bibinfo{year}{2022}), \bibinfo{pages}{763--769}.
\newblock


\bibitem[Suri et~al\mbox{.}(2024)]%
        {suri2024large}
\bibfield{author}{\bibinfo{person}{Gaurav Suri}, \bibinfo{person}{Lily~R. Slater}, \bibinfo{person}{Ali Ziaee}, {and} \bibinfo{person}{Morgan Nguyen}.} \bibinfo{year}{2024}\natexlab{}.
\newblock \showarticletitle{{Do Large Language Models Show Decision Heuristics Similar to Humans? A Case Study Using GPT-3.5}}.
\newblock \bibinfo{journal}{\emph{Journal of Experimental Psychology: General}} \bibinfo{volume}{153}, \bibinfo{number}{4} (\bibinfo{date}{04} \bibinfo{year}{2024}), \bibinfo{pages}{1066--1075}.
\newblock
\showISBNx{0096-3445, 0096-3445}


\bibitem[Swain(2024)]%
        {swain2024patients}
\bibfield{author}{\bibinfo{person}{Gyana Swain}.} \bibinfo{year}{2024}\natexlab{}.
\newblock \showarticletitle{{Patients may suffer from hallucinations of AI medical transcription tools}}.
\newblock \bibinfo{journal}{\emph{CIO}} (\bibinfo{year}{2024}).
\newblock
\urldef\tempurl%
\url{https://www.cio.com/article/3593403/patients-may-suffer-from-hallucinations-of-ai-medical-transcription-tools.html}
\showURL{%
\tempurl}


\bibitem[Taheri et~al\mbox{.}(2021)]%
        {taheri2021statistical}
\bibfield{author}{\bibinfo{person}{Mahsa Taheri}, \bibinfo{person}{Fang Xie}, {and} \bibinfo{person}{Johannes Lederer}.} \bibinfo{year}{2021}\natexlab{}.
\newblock \showarticletitle{{Statistical guarantees for regularized neural networks}}.
\newblock \bibinfo{journal}{\emph{Neural Networks}}  \bibinfo{volume}{142} (\bibinfo{year}{2021}), \bibinfo{pages}{148--161}.
\newblock
\showISSN{0893-6080}


\bibitem[Talmor et~al\mbox{.}(2021)]%
        {talmor2021commonsenseqa}
\bibfield{author}{\bibinfo{person}{Alon Talmor}, \bibinfo{person}{Ori Yoran}, \bibinfo{person}{Ronan Le~Bras}, \bibinfo{person}{Chandra Bhagavatula}, \bibinfo{person}{Yoav Goldberg}, \bibinfo{person}{Yejin Choi}, {and} \bibinfo{person}{Jonathan Berant}.} \bibinfo{year}{2021}\natexlab{}.
\newblock \showarticletitle{{CommonsenseQA 2.0: Exposing the Limits of AI through Gamification}}. In \bibinfo{booktitle}{\emph{Proceedings of the 35th Conference on Neural Information Processing Systems Datasets and Benchmarks Track}} (Virtual), Vol.~\bibinfo{volume}{1}.
\newblock


\bibitem[Thorne et~al\mbox{.}(2018)]%
        {thorne2018fever}
\bibfield{author}{\bibinfo{person}{James Thorne}, \bibinfo{person}{Andreas Vlachos}, \bibinfo{person}{Christos Christodoulopoulos}, {and} \bibinfo{person}{Arpit Mittal}.} \bibinfo{year}{2018}\natexlab{}.
\newblock \showarticletitle{{{FEVER}: a Large-scale Dataset for Fact Extraction and {VER}ification}}. In \bibinfo{booktitle}{\emph{Proceedings of the 2018 Conference of the North {A}merican Chapter of the Association for Computational Linguistics: Human Language Technologies}} (New Orleans, Louisiana). \bibinfo{publisher}{Association for Computational Linguistics}, \bibinfo{pages}{809--819}.
\newblock


\bibitem[Tiedemann(2012)]%
        {tiedemann2012parallel}
\bibfield{author}{\bibinfo{person}{J{\"o}rg Tiedemann}.} \bibinfo{year}{2012}\natexlab{}.
\newblock \showarticletitle{{Parallel Data, Tools and Interfaces in {OPUS}}}. In \bibinfo{booktitle}{\emph{Proceedings of the Eighth International Conference on Language Resources and Evaluation}} ({Istanbul, Turkey}). \bibinfo{publisher}{European Language Resources Association}, \bibinfo{pages}{2214--2218}.
\newblock


\bibitem[Tong et~al\mbox{.}(2024)]%
        {tong2024cambrian}
\bibfield{author}{\bibinfo{person}{Shengbang Tong}, \bibinfo{person}{Ellis Brown}, \bibinfo{person}{Penghao Wu}, \bibinfo{person}{Sanghyun Woo}, \bibinfo{person}{Manoj Middepogu}, \bibinfo{person}{Sai~Charitha Akula}, \bibinfo{person}{Jihan Yang}, \bibinfo{person}{Shusheng Yang}, \bibinfo{person}{Adithya Iyer}, \bibinfo{person}{Xichen Pan}, {et~al\mbox{.}}} \bibinfo{year}{2024}\natexlab{}.
\newblock \showarticletitle{Cambrian-1: A Fully Open, Vision-Centric Exploration of Multimodal LLMs}.
\newblock \bibinfo{journal}{\emph{arXiv preprint arXiv:2406.16860}} (\bibinfo{year}{2024}).
\newblock


\bibitem[Touvron et~al\mbox{.}(2023a)]%
        {touvron2023llama}
\bibfield{author}{\bibinfo{person}{Hugo Touvron}, \bibinfo{person}{Thibaut Lavril}, \bibinfo{person}{Gautier Izacard}, \bibinfo{person}{Xavier Martinet}, \bibinfo{person}{Marie-Anne Lachaux}, \bibinfo{person}{Timothée Lacroix}, \bibinfo{person}{Baptiste Rozière}, \bibinfo{person}{Naman Goyal}, \bibinfo{person}{Eric Hambro}, \bibinfo{person}{Faisal Azhar}, {et~al\mbox{.}}} \bibinfo{year}{2023}\natexlab{a}.
\newblock \showarticletitle{{LLaMA: Open and Efficient Foundation Language Models}}.
\newblock \bibinfo{journal}{\emph{arXiv preprint arXiv:2302.13971}} (\bibinfo{year}{2023}).
\newblock


\bibitem[Touvron et~al\mbox{.}(2023b)]%
        {touvron2023llama2}
\bibfield{author}{\bibinfo{person}{Hugo Touvron}, \bibinfo{person}{Louis Martin}, \bibinfo{person}{Kevin Stone}, \bibinfo{person}{Peter Albert}, \bibinfo{person}{Amjad Almahairi}, \bibinfo{person}{Yasmine Babaei}, \bibinfo{person}{Nikolay Bashlykov}, \bibinfo{person}{Soumya Batra}, \bibinfo{person}{Prajjwal Bhargava}, \bibinfo{person}{Shruti Bhosale}, {et~al\mbox{.}}} \bibinfo{year}{2023}\natexlab{b}.
\newblock \showarticletitle{{Llama 2: Open Foundation and Fine-Tuned Chat Models}}.
\newblock \bibinfo{journal}{\emph{arXiv preprint arXiv:2307.09288}} (\bibinfo{year}{2023}).
\newblock


\bibitem[Trivedi et~al\mbox{.}(2017)]%
        {trivedi2017lcquad}
\bibfield{author}{\bibinfo{person}{Priyansh Trivedi}, \bibinfo{person}{Gaurav Maheshwari}, \bibinfo{person}{Mohnish Dubey}, {and} \bibinfo{person}{Jens Lehmann}.} \bibinfo{year}{2017}\natexlab{}.
\newblock \showarticletitle{{LC-QuAD: A Corpus for Complex Question Answering over Knowledge Graphs}}. In \bibinfo{booktitle}{\emph{Proceedings of the 2017 International Semantic Web Conference}} (Vienna, Austria). \bibinfo{publisher}{Springer International Publishing}, \bibinfo{pages}{210--218}.
\newblock
\showISBNx{978-3-319-68204-4}


\bibitem[Uluoglakci and Temizel(2024)]%
        {uluoglakci2024hypotermqa}
\bibfield{author}{\bibinfo{person}{Cem Uluoglakci} {and} \bibinfo{person}{Tugba Temizel}.} \bibinfo{year}{2024}\natexlab{}.
\newblock \showarticletitle{{{H}ypo{T}erm{QA}: Hypothetical Terms Dataset for Benchmarking Hallucination Tendency of {LLM}s}}. In \bibinfo{booktitle}{\emph{Proceedings of the Student Research Workshop at the 18th Conference of the European Chapter of the Association for Computational Linguistics}} (St. Julian{'}s, Malta). \bibinfo{publisher}{Association for Computational Linguistics}, \bibinfo{pages}{95--136}.
\newblock


\bibitem[Varshney et~al\mbox{.}(2023)]%
        {varshney2023stitch}
\bibfield{author}{\bibinfo{person}{Neeraj Varshney}, \bibinfo{person}{Wenlin Yao}, \bibinfo{person}{Hongming Zhang}, \bibinfo{person}{Jianshu Chen}, {and} \bibinfo{person}{Dong Yu}.} \bibinfo{year}{2023}\natexlab{}.
\newblock \showarticletitle{{A Stitch in Time Saves Nine: Detecting and Mitigating Hallucinations of LLMs by Validating Low-Confidence Generation}}.
\newblock \bibinfo{journal}{\emph{arXiv preprint arXiv:2307.03987}} (\bibinfo{year}{2023}).
\newblock


\bibitem[Vrande\v{c}i\'{c} and Kr\"{o}tzsch(2014)]%
        {vrandevcic2014wikidata}
\bibfield{author}{\bibinfo{person}{Denny Vrande\v{c}i\'{c}} {and} \bibinfo{person}{Markus Kr\"{o}tzsch}.} \bibinfo{year}{2014}\natexlab{}.
\newblock \showarticletitle{{Wikidata: A Free Collaborative Knowledgebase}}.
\newblock \bibinfo{journal}{\emph{Commun. ACM}} \bibinfo{volume}{57}, \bibinfo{number}{10} (\bibinfo{date}{Sept.} \bibinfo{year}{2014}), \bibinfo{pages}{78–85}.
\newblock
\showISSN{0001-0782}


\bibitem[Wang et~al\mbox{.}(2023)]%
        {wang2023decoding}
\bibfield{author}{\bibinfo{person}{Boxin Wang}, \bibinfo{person}{Weixin Chen}, \bibinfo{person}{Hengzhi Pei}, \bibinfo{person}{Chulin Xie}, \bibinfo{person}{Mintong Kang}, \bibinfo{person}{Chenhui Zhang}, \bibinfo{person}{Chejian Xu}, \bibinfo{person}{Zidi Xiong}, \bibinfo{person}{Ritik Dutta}, \bibinfo{person}{Rylan Schaeffer}, {et~al\mbox{.}}} \bibinfo{year}{2023}\natexlab{}.
\newblock \showarticletitle{{DecodingTrust: A Comprehensive Assessment of Trustworthiness in {GPT} Models}}. In \bibinfo{booktitle}{\emph{Proceedings of the 37th Conference on Neural Information Processing Systems Datasets and Benchmarks Track}} (New Orleans, LA, USA).
\newblock


\bibitem[Wang et~al\mbox{.}(2024c)]%
        {wang2023voyager}
\bibfield{author}{\bibinfo{person}{Guanzhi Wang}, \bibinfo{person}{Yuqi Xie}, \bibinfo{person}{Yunfan Jiang}, \bibinfo{person}{Ajay Mandlekar}, \bibinfo{person}{Chaowei Xiao}, \bibinfo{person}{Yuke Zhu}, \bibinfo{person}{Linxi Fan}, {and} \bibinfo{person}{Anima Anandkumar}.} \bibinfo{year}{2024}\natexlab{c}.
\newblock \showarticletitle{{Voyager: An Open-Ended Embodied Agent with Large Language Models}}.
\newblock \bibinfo{journal}{\emph{Transactions on Machine Learning Research}} (\bibinfo{year}{2024}).
\newblock
\showISSN{2835-8856}


\bibitem[Wang et~al\mbox{.}(2024a)]%
        {wang2024hallod}
\bibfield{author}{\bibinfo{person}{Hongbo Wang}, \bibinfo{person}{Jie Cao}, \bibinfo{person}{Jin Liu}, \bibinfo{person}{Xiaoqiang Zhou}, \bibinfo{person}{Huaibo Huang}, {and} \bibinfo{person}{Ran He}.} \bibinfo{year}{2024}\natexlab{a}.
\newblock \showarticletitle{{Hallo3D: Multi-Modal Hallucination Detection and Mitigation for Consistent 3D Content Generation}}. In \bibinfo{booktitle}{\emph{Proceedings of the 2024 Conference on Neural Information Processing Systems}}.
\newblock


\bibitem[Wang et~al\mbox{.}(2024b)]%
        {wang2024conformal}
\bibfield{author}{\bibinfo{person}{Jun Wang}, \bibinfo{person}{Jiaming Tong}, \bibinfo{person}{Kaiyuan Tan}, \bibinfo{person}{Yevgeniy Vorobeychik}, {and} \bibinfo{person}{Yiannis Kantaros}.} \bibinfo{year}{2024}\natexlab{b}.
\newblock \showarticletitle{{Conformal Temporal Logic Planning using Large Language Models}}.
\newblock \bibinfo{journal}{\emph{arXiv preprint arXiv:2309.10092}} (\bibinfo{year}{2024}).
\newblock


\bibitem[Wegener et~al\mbox{.}(2008)]%
        {wegener2008traci}
\bibfield{author}{\bibinfo{person}{Axel Wegener}, \bibinfo{person}{Micha\l{} Pi\'{o}rkowski}, \bibinfo{person}{Maxim Raya}, \bibinfo{person}{Horst Hellbr\"{u}ck}, \bibinfo{person}{Stefan Fischer}, {and} \bibinfo{person}{Jean-Pierre Hubaux}.} \bibinfo{year}{2008}\natexlab{}.
\newblock \showarticletitle{{TraCI: an interface for coupling road traffic and network simulators}}. In \bibinfo{booktitle}{\emph{Proceedings of the 11th Communications and Networking Simulation Symposium}} (Ottawa, Canada). \bibinfo{publisher}{Association for Computing Machinery}, \bibinfo{address}{New York, NY, USA}, \bibinfo{pages}{155–163}.
\newblock
\showISBNx{1565553187}


\bibitem[Wei et~al\mbox{.}(2022)]%
        {wei2022chain}
\bibfield{author}{\bibinfo{person}{Jason Wei}, \bibinfo{person}{Xuezhi Wang}, \bibinfo{person}{Dale Schuurmans}, \bibinfo{person}{Maarten Bosma}, \bibinfo{person}{Brian Ichter}, \bibinfo{person}{Fei Xia}, \bibinfo{person}{Ed Chi}, \bibinfo{person}{Quoc~V Le}, {and} \bibinfo{person}{Denny Zhou}.} \bibinfo{year}{2022}\natexlab{}.
\newblock \showarticletitle{{Chain-of-Thought Prompting Elicits Reasoning in Large Language Models}}. In \bibinfo{booktitle}{\emph{Proceedings of the 2022 Conference on Neural Information Processing Systems}} (New Orleans, LA, USA). \bibinfo{publisher}{Curran Associates, Inc.}, \bibinfo{pages}{24824--24837}.
\newblock


\bibitem[Weiser(2023)]%
        {weiser2023here}
\bibfield{author}{\bibinfo{person}{Benjamin Weiser}.} \bibinfo{year}{2023}\natexlab{}.
\newblock \showarticletitle{{Here’s What Happens When Your Lawyer Uses ChatGPT}}.
\newblock \bibinfo{journal}{\emph{The New York Times}} (\bibinfo{year}{2023}).
\newblock
\urldef\tempurl%
\url{https://www.nytimes.com/2023/05/27/nyregion/avianca-airline-lawsuit-chatgpt.html}
\showURL{%
\tempurl}


\bibitem[Welbl et~al\mbox{.}(2017)]%
        {welbl2017crowdsourcing}
\bibfield{author}{\bibinfo{person}{Johannes Welbl}, \bibinfo{person}{Nelson~F. Liu}, {and} \bibinfo{person}{Matt Gardner}.} \bibinfo{year}{2017}\natexlab{}.
\newblock \showarticletitle{{Crowdsourcing Multiple Choice Science Questions}}. In \bibinfo{booktitle}{\emph{Proceedings of the 3rd Workshop on Noisy User-generated Text}}. \bibinfo{publisher}{Association for Computational Linguistics}, \bibinfo{address}{Copenhagen, Denmark}, \bibinfo{pages}{94--106}.
\newblock


\bibitem[Wen et~al\mbox{.}(2024a)]%
        {wen2024dilu}
\bibfield{author}{\bibinfo{person}{Licheng Wen}, \bibinfo{person}{Daocheng Fu}, \bibinfo{person}{Xin Li}, \bibinfo{person}{Xinyu Cai}, \bibinfo{person}{Tao MA}, \bibinfo{person}{Pinlong Cai}, \bibinfo{person}{Min Dou}, \bibinfo{person}{Botian Shi}, \bibinfo{person}{Liang He}, {and} \bibinfo{person}{Yu Qiao}.} \bibinfo{year}{2024}\natexlab{a}.
\newblock \showarticletitle{{DiLu: A Knowledge-Driven Approach to Autonomous Driving with Large Language Models}}. In \bibinfo{booktitle}{\emph{Proceedings of the 12th International Conference on Learning Representations}} (Vienna, Austria).
\newblock


\bibitem[Wen et~al\mbox{.}(2024b)]%
        {wen2023road}
\bibfield{author}{\bibinfo{person}{Licheng Wen}, \bibinfo{person}{Xuemeng Yang}, \bibinfo{person}{Daocheng Fu}, \bibinfo{person}{Xiaofeng Wang}, \bibinfo{person}{Pinlong Cai}, \bibinfo{person}{Xin Li}, \bibinfo{person}{Tao MA}, \bibinfo{person}{Yingxuan Li}, \bibinfo{person}{Linran XU}, \bibinfo{person}{Dengke Shang}, {et~al\mbox{.}}} \bibinfo{year}{2024}\natexlab{b}.
\newblock \showarticletitle{{On the Road with {GPT}-4V(ision): Explorations of Utilizing Visual-Language Model as Autonomous Driving Agent}}. In \bibinfo{booktitle}{\emph{Proceedings of the Workshop on Large Language Model (LLM) Agents at the 12th International Conference on Learning Representations}} (Vienna, Austria).
\newblock


\bibitem[Williams(1992)]%
        {williams1992simple}
\bibfield{author}{\bibinfo{person}{Ronald~J. Williams}.} \bibinfo{year}{1992}\natexlab{}.
\newblock \showarticletitle{Simple Statistical Gradient-Following Algorithms for Connectionist Reinforcement Learning}.
\newblock \bibinfo{journal}{\emph{Machine Learning}} \bibinfo{volume}{8}, \bibinfo{number}{3–4} (\bibinfo{date}{may} \bibinfo{year}{1992}), \bibinfo{pages}{229–256}.
\newblock
\showISSN{0885-6125}


\bibitem[Wu et~al\mbox{.}(2023a)]%
        {wu2023referring}
\bibfield{author}{\bibinfo{person}{Dongming Wu}, \bibinfo{person}{Wencheng Han}, \bibinfo{person}{Tiancai Wang}, \bibinfo{person}{Xingping Dong}, \bibinfo{person}{Xiangyu Zhang}, {and} \bibinfo{person}{Jianbing Shen}.} \bibinfo{year}{2023}\natexlab{a}.
\newblock \showarticletitle{{Referring Multi-Object Tracking}}. In \bibinfo{booktitle}{\emph{Proceedings of the 2023 IEEE/CVF Conference on Computer Vision and Pattern Recognition (CVPR)}} (Vancouver, Canada). \bibinfo{publisher}{Institute of Electrical and Electronics Engineers}, \bibinfo{pages}{14633--14642}.
\newblock


\bibitem[Wu et~al\mbox{.}(2023b)]%
        {wu2023language}
\bibfield{author}{\bibinfo{person}{Dongming Wu}, \bibinfo{person}{Wencheng Han}, \bibinfo{person}{Tiancai Wang}, \bibinfo{person}{Yingfei Liu}, \bibinfo{person}{Xiangyu Zhang}, {and} \bibinfo{person}{Jianbing Shen}.} \bibinfo{year}{2023}\natexlab{b}.
\newblock \showarticletitle{{Language Prompt for Autonomous Driving}}.
\newblock \bibinfo{journal}{\emph{arXiv preprint arXiv:2309.04379}} (\bibinfo{year}{2023}).
\newblock


\bibitem[Wu et~al\mbox{.}(2024)]%
        {wu2024well}
\bibfield{author}{\bibinfo{person}{Kevin Wu}, \bibinfo{person}{Eric Wu}, \bibinfo{person}{Ally Cassasola}, \bibinfo{person}{Angela Zhang}, \bibinfo{person}{Kevin Wei}, \bibinfo{person}{Teresa Nguyen}, \bibinfo{person}{Sith Riantawan}, \bibinfo{person}{Patricia~Shi Riantawan}, \bibinfo{person}{Daniel~E Ho}, {and} \bibinfo{person}{James Zou}.} \bibinfo{year}{2024}\natexlab{}.
\newblock \showarticletitle{{How well do LLMs cite relevant medical references? An evaluation framework and analyses}}.
\newblock \bibinfo{journal}{\emph{arXiv preprint arXiv:2402.02008}} (\bibinfo{year}{2024}).
\newblock


\bibitem[Wu et~al\mbox{.}(2023c)]%
        {wu2023bloomberggpt}
\bibfield{author}{\bibinfo{person}{Shijie Wu}, \bibinfo{person}{Ozan Irsoy}, \bibinfo{person}{Steven Lu}, \bibinfo{person}{Vadim Dabravolski}, \bibinfo{person}{Mark Dredze}, \bibinfo{person}{Sebastian Gehrmann}, \bibinfo{person}{Prabhanjan Kambadur}, \bibinfo{person}{David Rosenberg}, {and} \bibinfo{person}{Gideon Mann}.} \bibinfo{year}{2023}\natexlab{c}.
\newblock \showarticletitle{{BloombergGPT: A Large Language Model for Finance}}.
\newblock \bibinfo{journal}{\emph{arXiv preprint arXiv:2303.17564}} (\bibinfo{year}{2023}).
\newblock


\bibitem[Xia and Xue(2023)]%
        {xia2023survey}
\bibfield{author}{\bibinfo{person}{Weihao Xia} {and} \bibinfo{person}{Jing-Hao Xue}.} \bibinfo{year}{2023}\natexlab{}.
\newblock \showarticletitle{{A Survey on Deep Generative 3D-aware Image Synthesis}}.
\newblock \bibinfo{journal}{\emph{Comput. Surveys}} \bibinfo{volume}{56}, \bibinfo{number}{4}, Article \bibinfo{articleno}{90} (\bibinfo{date}{nov} \bibinfo{year}{2023}), \bibinfo{numpages}{34}~pages.
\newblock
\showISSN{0360-0300}


\bibitem[Xiao et~al\mbox{.}(2010)]%
        {xiao2010sun}
\bibfield{author}{\bibinfo{person}{Jianxiong Xiao}, \bibinfo{person}{James Hays}, \bibinfo{person}{Krista~A. Ehinger}, \bibinfo{person}{Aude Oliva}, {and} \bibinfo{person}{Antonio Torralba}.} \bibinfo{year}{2010}\natexlab{}.
\newblock \showarticletitle{{SUN Database: Large-scale Scene Recognition from Abbey to Zoo}}. In \bibinfo{booktitle}{\emph{Proceedings of the 2010 IEEE Conference on Computer Vision and Pattern Recognition (CVPR)}} (San Francisco, CA, USA). \bibinfo{publisher}{Institute of Electrical and Electronics Engineers}, \bibinfo{pages}{3485--3492}.
\newblock


\bibitem[Xiong et~al\mbox{.}(2024)]%
        {xiong2024can}
\bibfield{author}{\bibinfo{person}{Miao Xiong}, \bibinfo{person}{Zhiyuan Hu}, \bibinfo{person}{Xinyang Lu}, \bibinfo{person}{YIFEI LI}, \bibinfo{person}{Jie Fu}, \bibinfo{person}{Junxian He}, {and} \bibinfo{person}{Bryan Hooi}.} \bibinfo{year}{2024}\natexlab{}.
\newblock \showarticletitle{{Can {LLM}s Express Their Uncertainty? An Empirical Evaluation of Confidence Elicitation in {LLM}s}}. In \bibinfo{booktitle}{\emph{Proceedings of the 12th International Conference on Learning Representations}} (Vienna, Austria).
\newblock


\bibitem[Xu et~al\mbox{.}(2023)]%
        {xu2023seeavatar}
\bibfield{author}{\bibinfo{person}{Yuanyou Xu}, \bibinfo{person}{Zongxin Yang}, {and} \bibinfo{person}{Yi Yang}.} \bibinfo{year}{2023}\natexlab{}.
\newblock \showarticletitle{{SEEAvatar: Photorealistic Text-to-3D Avatar Generation with Constrained Geometry and Appearance}}.
\newblock \bibinfo{journal}{\emph{arXiv preprint arXiv:2312.08889}} (\bibinfo{year}{2023}).
\newblock


\bibitem[Xu et~al\mbox{.}(2024)]%
        {xu2024drivegpt4}
\bibfield{author}{\bibinfo{person}{Zhenhua Xu}, \bibinfo{person}{Yujia Zhang}, \bibinfo{person}{Enze Xie}, \bibinfo{person}{Zhen Zhao}, \bibinfo{person}{Yong Guo}, \bibinfo{person}{Kwan-Yee~K. Wong}, \bibinfo{person}{Zhenguo Li}, {and} \bibinfo{person}{Hengshuang Zhao}.} \bibinfo{year}{2024}\natexlab{}.
\newblock \showarticletitle{{DriveGPT4: Interpretable End-to-End Autonomous Driving Via Large Language Model}}.
\newblock \bibinfo{journal}{\emph{IEEE Robotics and Automation Letters}} \bibinfo{volume}{9}, \bibinfo{number}{10} (\bibinfo{year}{2024}), \bibinfo{pages}{8186--8193}.
\newblock


\bibitem[Yang et~al\mbox{.}(2024a)]%
        {yang20243d}
\bibfield{author}{\bibinfo{person}{Jianing Yang}, \bibinfo{person}{Xuweiyi Chen}, \bibinfo{person}{Nikhil Madaan}, \bibinfo{person}{Madhavan Iyengar}, \bibinfo{person}{Shengyi Qian}, \bibinfo{person}{David~F Fouhey}, {and} \bibinfo{person}{Joyce Chai}.} \bibinfo{year}{2024}\natexlab{a}.
\newblock \showarticletitle{{3D-GRAND: A Million-Scale Dataset for 3D-LLMs with Better Grounding and Less Hallucination}}.
\newblock \bibinfo{journal}{\emph{arXiv preprint arXiv:2406.05132}} (\bibinfo{year}{2024}).
\newblock


\bibitem[Yang et~al\mbox{.}(2024d)]%
        {yang2024harnessing}
\bibfield{author}{\bibinfo{person}{Jingfeng Yang}, \bibinfo{person}{Hongye Jin}, \bibinfo{person}{Ruixiang Tang}, \bibinfo{person}{Xiaotian Han}, \bibinfo{person}{Qizhang Feng}, \bibinfo{person}{Haoming Jiang}, \bibinfo{person}{Shaochen Zhong}, \bibinfo{person}{Bing Yin}, {and} \bibinfo{person}{Xia Hu}.} \bibinfo{year}{2024}\natexlab{d}.
\newblock \showarticletitle{{Harnessing the Power of LLMs in Practice: A Survey on ChatGPT and Beyond}}.
\newblock \bibinfo{journal}{\emph{ACM Transactions on Knowledge Discovery from Data}} (\bibinfo{date}{feb} \bibinfo{year}{2024}).
\newblock
\showISSN{1556-4681}


\bibitem[Yang et~al\mbox{.}(2024b)]%
        {yang2023alignment}
\bibfield{author}{\bibinfo{person}{Yuqing Yang}, \bibinfo{person}{Ethan Chern}, \bibinfo{person}{Xipeng Qiu}, \bibinfo{person}{Graham Neubig}, {and} \bibinfo{person}{Pengfei Liu}.} \bibinfo{year}{2024}\natexlab{b}.
\newblock \showarticletitle{{Alignment for Honesty}}. In \bibinfo{booktitle}{\emph{Proceedings of the 2024 Conference on Neural Information Processing Systems}} (Vancouver, Canada).
\newblock


\bibitem[Yang et~al\mbox{.}(2024e)]%
        {yang2024holodeck}
\bibfield{author}{\bibinfo{person}{Yue Yang}, \bibinfo{person}{Fan-Yun Sun}, \bibinfo{person}{Luca Weihs}, \bibinfo{person}{Eli Vanderbilt}, \bibinfo{person}{Alvaro Herrasti}, \bibinfo{person}{Winson Han}, \bibinfo{person}{Jiajun Wu}, \bibinfo{person}{Nick Haber}, \bibinfo{person}{Ranjay Krishna}, \bibinfo{person}{Lingjie Liu}, {et~al\mbox{.}}} \bibinfo{year}{2024}\natexlab{e}.
\newblock \showarticletitle{{Holodeck: Language Guided Generation of 3D Embodied AI Environments}}. In \bibinfo{booktitle}{\emph{Proceedings of the 2024 IEEE/CVF Conference on Computer Vision and Pattern Recognition (CVPR)}} (Seattle, WA, USA). \bibinfo{publisher}{Institute of Electrical and Electronics Engineers}, \bibinfo{pages}{16277--16287}.
\newblock


\bibitem[Yang et~al\mbox{.}(2024c)]%
        {yang2023llm4drive}
\bibfield{author}{\bibinfo{person}{Zhenjie Yang}, \bibinfo{person}{Xiaosong Jia}, \bibinfo{person}{Hongyang Li}, {and} \bibinfo{person}{Junchi Yan}.} \bibinfo{year}{2024}\natexlab{c}.
\newblock \showarticletitle{{{LLM}4Drive: A Survey of Large Language Models for Autonomous Driving}}. In \bibinfo{booktitle}{\emph{Proceedings of the Workshop on Open-World Agents at the 2024 Conference on Neural Information Processing Systems}} (Vancouver, Canada).
\newblock


\bibitem[Yang et~al\mbox{.}(2018)]%
        {yang2018hotpotqa}
\bibfield{author}{\bibinfo{person}{Zhilin Yang}, \bibinfo{person}{Peng Qi}, \bibinfo{person}{Saizheng Zhang}, \bibinfo{person}{Yoshua Bengio}, \bibinfo{person}{William Cohen}, \bibinfo{person}{Ruslan Salakhutdinov}, {and} \bibinfo{person}{Christopher~D. Manning}.} \bibinfo{year}{2018}\natexlab{}.
\newblock \showarticletitle{{{H}otpot{QA}: A Dataset for Diverse, Explainable Multi-hop Question Answering}}. In \bibinfo{booktitle}{\emph{Proceedings of the 2018 Conference on Empirical Methods in Natural Language Processing}} (Brussels, Belgium). \bibinfo{publisher}{Association for Computational Linguistics}, \bibinfo{pages}{2369--2380}.
\newblock


\bibitem[Yao et~al\mbox{.}(2023a)]%
        {yao2023llm}
\bibfield{author}{\bibinfo{person}{Jia-Yu Yao}, \bibinfo{person}{Kun-Peng Ning}, \bibinfo{person}{Zhen-Hui Liu}, \bibinfo{person}{Mu-Nan Ning}, {and} \bibinfo{person}{Li Yuan}.} \bibinfo{year}{2023}\natexlab{a}.
\newblock \showarticletitle{{LLM Lies: Hallucinations are not Bugs, but Features as Adversarial Examples}}.
\newblock \bibinfo{journal}{\emph{arXiv preprint arXiv:2310.01469}} (\bibinfo{year}{2023}).
\newblock


\bibitem[Yao et~al\mbox{.}(2023b)]%
        {yao2023react}
\bibfield{author}{\bibinfo{person}{Shunyu Yao}, \bibinfo{person}{Jeffrey Zhao}, \bibinfo{person}{Dian Yu}, \bibinfo{person}{Nan Du}, \bibinfo{person}{Izhak Shafran}, \bibinfo{person}{Karthik~R Narasimhan}, {and} \bibinfo{person}{Yuan Cao}.} \bibinfo{year}{2023}\natexlab{b}.
\newblock \showarticletitle{{ReAct: Synergizing Reasoning and Acting in Language Models}}. In \bibinfo{booktitle}{\emph{Proceedings of the 11th International Conference on Learning Representations}} (Kigali, Rwanda).
\newblock


\bibitem[Ye et~al\mbox{.}(2023)]%
        {ye2023cognitive}
\bibfield{author}{\bibinfo{person}{Hongbin Ye}, \bibinfo{person}{Tong Liu}, \bibinfo{person}{Aijia Zhang}, \bibinfo{person}{Wei Hua}, {and} \bibinfo{person}{Weiqiang Jia}.} \bibinfo{year}{2023}\natexlab{}.
\newblock \showarticletitle{{Cognitive Mirage: A Review of Hallucinations in Large Language Models}}.
\newblock \bibinfo{journal}{\emph{arXiv preprint arXiv:2309.06794}} (\bibinfo{year}{2023}).
\newblock


\bibitem[Yehuda et~al\mbox{.}(2024)]%
        {yehuda2024interrogatellm}
\bibfield{author}{\bibinfo{person}{Yakir Yehuda}, \bibinfo{person}{Itzik Malkiel}, \bibinfo{person}{Oren Barkan}, \bibinfo{person}{Jonathan Weill}, \bibinfo{person}{Royi Ronen}, {and} \bibinfo{person}{Noam Koenigstein}.} \bibinfo{year}{2024}\natexlab{}.
\newblock \showarticletitle{{{I}nterrogate{LLM}: Zero-Resource Hallucination Detection in {LLM}-Generated Answers}}. In \bibinfo{booktitle}{\emph{Proceedings of the 62nd Annual Meeting of the Association for Computational Linguistics}} (Bangkok, Thailand). \bibinfo{publisher}{Association for Computational Linguistics}, \bibinfo{pages}{9333--9347}.
\newblock


\bibitem[Yoshino et~al\mbox{.}(2019)]%
        {yoshino2019dialog}
\bibfield{author}{\bibinfo{person}{Koichiro Yoshino}, \bibinfo{person}{Chiori Hori}, \bibinfo{person}{Julien Perez}, \bibinfo{person}{Luis~Fernando D'Haro}, \bibinfo{person}{Lazaros Polymenakos}, \bibinfo{person}{Chulaka Gunasekara}, \bibinfo{person}{Walter~S Lasecki}, \bibinfo{person}{Jonathan~K Kummerfeld}, \bibinfo{person}{Michel Galley}, \bibinfo{person}{Chris Brockett}, {et~al\mbox{.}}} \bibinfo{year}{2019}\natexlab{}.
\newblock \showarticletitle{{Dialog System Technology Challenge 7}}.
\newblock \bibinfo{journal}{\emph{arXiv preprint arXiv:1901.03461}} (\bibinfo{year}{2019}).
\newblock


\bibitem[Yu et~al\mbox{.}(2015)]%
        {yu2015lsun}
\bibfield{author}{\bibinfo{person}{Fisher Yu}, \bibinfo{person}{Ari Seff}, \bibinfo{person}{Yinda Zhang}, \bibinfo{person}{Shuran Song}, \bibinfo{person}{Thomas Funkhouser}, {and} \bibinfo{person}{Jianxiong Xiao}.} \bibinfo{year}{2015}\natexlab{}.
\newblock \showarticletitle{{LSUN: Construction of a Large-Scale Image Dataset using Deep Learning with Humans in the Loop}}.
\newblock \bibinfo{journal}{\emph{arXiv preprint arXiv:1506.03365}} (\bibinfo{year}{2015}).
\newblock


\bibitem[Yu et~al\mbox{.}(2024)]%
        {yu2023automatic}
\bibfield{author}{\bibinfo{person}{Xiaodong Yu}, \bibinfo{person}{Hao Cheng}, \bibinfo{person}{Xiaodong Liu}, \bibinfo{person}{Dan Roth}, {and} \bibinfo{person}{Jianfeng Gao}.} \bibinfo{year}{2024}\natexlab{}.
\newblock \showarticletitle{{{R}e{E}val: Automatic Hallucination Evaluation for Retrieval-Augmented Large Language Models via Transferable Adversarial Attacks}}. In \bibinfo{booktitle}{\emph{Findings of the 2024 Conference of the North American Chapter of the Association for Computational Linguistics}} (Mexico City, Mexico). \bibinfo{publisher}{Association for Computational Linguistics}, \bibinfo{pages}{1333--1351}.
\newblock


\bibitem[Yuan et~al\mbox{.}(2021)]%
        {yuan2021bartscore}
\bibfield{author}{\bibinfo{person}{Weizhe Yuan}, \bibinfo{person}{Graham Neubig}, {and} \bibinfo{person}{Pengfei Liu}.} \bibinfo{year}{2021}\natexlab{}.
\newblock \showarticletitle{{BARTScore: Evaluating Generated Text as Text Generation}}. In \bibinfo{booktitle}{\emph{Proceedings of the 2021 Conference on Neural Information Processing Systems}} (Virtual). \bibinfo{publisher}{Curran Associates, Inc.}, \bibinfo{pages}{27263--27277}.
\newblock


\bibitem[Zeng et~al\mbox{.}(2021)]%
        {zeng2021transporter}
\bibfield{author}{\bibinfo{person}{Andy Zeng}, \bibinfo{person}{Pete Florence}, \bibinfo{person}{Jonathan Tompson}, \bibinfo{person}{Stefan Welker}, \bibinfo{person}{Jonathan Chien}, \bibinfo{person}{Maria Attarian}, \bibinfo{person}{Travis Armstrong}, \bibinfo{person}{Ivan Krasin}, \bibinfo{person}{Dan Duong}, \bibinfo{person}{Vikas Sindhwani}, {and} \bibinfo{person}{Johnny Lee}.} \bibinfo{year}{2021}\natexlab{}.
\newblock \showarticletitle{{Transporter Networks: Rearranging the Visual World for Robotic Manipulation}}. In \bibinfo{booktitle}{\emph{Proceedings of the 4th Conference on Robot Learning}} (London, United Kingdom) \emph{(\bibinfo{series}{Proceedings of Machine Learning Research}, Vol.~\bibinfo{volume}{155})}. \bibinfo{publisher}{PMLR}, \bibinfo{pages}{726--747}.
\newblock


\bibitem[Zeng et~al\mbox{.}(2023)]%
        {zeng2023large}
\bibfield{author}{\bibinfo{person}{Fanlong Zeng}, \bibinfo{person}{Wensheng Gan}, \bibinfo{person}{Yongheng Wang}, \bibinfo{person}{Ning Liu}, {and} \bibinfo{person}{Philip~S. Yu}.} \bibinfo{year}{2023}\natexlab{}.
\newblock \showarticletitle{{Large Language Models for Robotics: A Survey}}.
\newblock \bibinfo{journal}{\emph{arXiv preprint arXiv:2311.07226}} (\bibinfo{year}{2023}).
\newblock


\bibitem[Zha et~al\mbox{.}(2023)]%
        {zha2023alignscore}
\bibfield{author}{\bibinfo{person}{Yuheng Zha}, \bibinfo{person}{Yichi Yang}, \bibinfo{person}{Ruichen Li}, {and} \bibinfo{person}{Zhiting Hu}.} \bibinfo{year}{2023}\natexlab{}.
\newblock \showarticletitle{{{A}lign{S}core: Evaluating Factual Consistency with A Unified Alignment Function}}. In \bibinfo{booktitle}{\emph{Proceedings of the 61st Annual Meeting of the Association for Computational Linguistics}} (Toronto, Canada). \bibinfo{publisher}{Association for Computational Linguistics}, \bibinfo{pages}{11328--11348}.
\newblock


\bibitem[Zhang et~al\mbox{.}(2023a)]%
        {zhang2023large}
\bibfield{author}{\bibinfo{person}{Ceng Zhang}, \bibinfo{person}{Junxin Chen}, \bibinfo{person}{Jiatong Li}, \bibinfo{person}{Yanhong Peng}, {and} \bibinfo{person}{Zebing Mao}.} \bibinfo{year}{2023}\natexlab{a}.
\newblock \showarticletitle{{Large language models for human-robot interaction: A review}}.
\newblock \bibinfo{journal}{\emph{Biomimetic Intelligence and Robotics}} \bibinfo{volume}{3}, \bibinfo{number}{4} (\bibinfo{year}{2023}), \bibinfo{pages}{100131}.
\newblock
\showISSN{2667-3797}


\bibitem[Zhang et~al\mbox{.}(2024b)]%
        {zhang2023language}
\bibfield{author}{\bibinfo{person}{Muru Zhang}, \bibinfo{person}{Ofir Press}, \bibinfo{person}{William Merrill}, \bibinfo{person}{Alisa Liu}, {and} \bibinfo{person}{Noah~A. Smith}.} \bibinfo{year}{2024}\natexlab{b}.
\newblock \showarticletitle{{How Language Model Hallucinations Can Snowball}}. In \bibinfo{booktitle}{\emph{Proceedings of the 41st International Conference on Machine Learning}} (Vienna, Austria) \emph{(\bibinfo{series}{Proceedings of Machine Learning Research}, Vol.~\bibinfo{volume}{235})}. \bibinfo{publisher}{PMLR}, \bibinfo{pages}{59670--59684}.
\newblock


\bibitem[Zhang et~al\mbox{.}(2024a)]%
        {zhang2023knowledge}
\bibfield{author}{\bibinfo{person}{Shuo Zhang}, \bibinfo{person}{Liangming Pan}, \bibinfo{person}{Junzhou Zhao}, {and} \bibinfo{person}{William~Yang Wang}.} \bibinfo{year}{2024}\natexlab{a}.
\newblock \showarticletitle{{The Knowledge Alignment Problem: Bridging Human and External Knowledge for Large Language Models}}. In \bibinfo{booktitle}{\emph{Findings of the 62nd Annual Meeting of the Association for Computational Linguistics}} (Bangkok, Thailand). \bibinfo{publisher}{Association for Computational Linguistics}, \bibinfo{pages}{2025--2038}.
\newblock


\bibitem[Zhang et~al\mbox{.}(2020)]%
        {zhang2020bertscore}
\bibfield{author}{\bibinfo{person}{Tianyi Zhang}, \bibinfo{person}{Varsha Kishore}, \bibinfo{person}{Felix Wu}, \bibinfo{person}{Kilian~Q. Weinberger}, {and} \bibinfo{person}{Yoav Artzi}.} \bibinfo{year}{2020}\natexlab{}.
\newblock \showarticletitle{{BERTScore: Evaluating Text Generation with BERT}}. In \bibinfo{booktitle}{\emph{Proceedings of the 8th International Conference on Learning Representations}} (Virtual).
\newblock


\bibitem[Zhang et~al\mbox{.}(2023b)]%
        {zhang2023siren}
\bibfield{author}{\bibinfo{person}{Yue Zhang}, \bibinfo{person}{Yafu Li}, \bibinfo{person}{Leyang Cui}, \bibinfo{person}{Deng Cai}, \bibinfo{person}{Lemao Liu}, \bibinfo{person}{Tingchen Fu}, \bibinfo{person}{Xinting Huang}, \bibinfo{person}{Enbo Zhao}, \bibinfo{person}{Yu Zhang}, \bibinfo{person}{Yulong Chen}, {et~al\mbox{.}}} \bibinfo{year}{2023}\natexlab{b}.
\newblock \showarticletitle{{Siren's Song in the AI Ocean: A Survey on Hallucination in Large Language Models}}.
\newblock \bibinfo{journal}{\emph{arXiv preprint arXiv:2309.01219}} (\bibinfo{year}{2023}).
\newblock


\bibitem[Zhao et~al\mbox{.}(2024a)]%
        {zhao2024revolutionizing}
\bibfield{author}{\bibinfo{person}{Huaqin Zhao}, \bibinfo{person}{Zhengliang Liu}, \bibinfo{person}{Zihao Wu}, \bibinfo{person}{Yiwei Li}, \bibinfo{person}{Tianze Yang}, \bibinfo{person}{Peng Shu}, \bibinfo{person}{Shaochen Xu}, \bibinfo{person}{Haixing Dai}, \bibinfo{person}{Lin Zhao}, \bibinfo{person}{Gengchen Mai}, {et~al\mbox{.}}} \bibinfo{year}{2024}\natexlab{a}.
\newblock \showarticletitle{{Revolutionizing Finance with LLMs: An Overview of Applications and Insights}}.
\newblock \bibinfo{journal}{\emph{arXiv preprint arXiv:2401.11641}} (\bibinfo{year}{2024}).
\newblock


\bibitem[Zhao et~al\mbox{.}(2020)]%
        {zhao2020sim}
\bibfield{author}{\bibinfo{person}{Wenshuai Zhao}, \bibinfo{person}{Jorge~Peña Queralta}, {and} \bibinfo{person}{Tomi Westerlund}.} \bibinfo{year}{2020}\natexlab{}.
\newblock \showarticletitle{{Sim-to-Real Transfer in Deep Reinforcement Learning for Robotics: a Survey}}. In \bibinfo{booktitle}{\emph{2020 IEEE Symposium Series on Computational Intelligence (SSCI)}} (Canberra, ACT, Australia). \bibinfo{pages}{737--744}.
\newblock


\bibitem[Zhao et~al\mbox{.}(2023)]%
        {zhao2023survey}
\bibfield{author}{\bibinfo{person}{Wayne~Xin Zhao}, \bibinfo{person}{Kun Zhou}, \bibinfo{person}{Junyi Li}, \bibinfo{person}{Tianyi Tang}, \bibinfo{person}{Xiaolei Wang}, \bibinfo{person}{Yupeng Hou}, \bibinfo{person}{Yingqian Min}, \bibinfo{person}{Beichen Zhang}, \bibinfo{person}{Junjie Zhang}, \bibinfo{person}{Zican Dong}, {et~al\mbox{.}}} \bibinfo{year}{2023}\natexlab{}.
\newblock \showarticletitle{{A Survey of Large Language Models}}.
\newblock \bibinfo{journal}{\emph{arXiv preprint arXiv:2303.18223}} (\bibinfo{year}{2023}).
\newblock


\bibitem[Zhao et~al\mbox{.}(2024b)]%
        {zhao2024hallucinations}
\bibfield{author}{\bibinfo{person}{Zhiyuan Zhao}, \bibinfo{person}{Bin Wang}, \bibinfo{person}{Linke Ouyang}, \bibinfo{person}{Xiaoyi Dong}, \bibinfo{person}{Jiaqi Wang}, {and} \bibinfo{person}{Conghui He}.} \bibinfo{year}{2024}\natexlab{b}.
\newblock \showarticletitle{{Beyond Hallucinations: Enhancing LVLMs through Hallucination-Aware Direct Preference Optimization}}.
\newblock \bibinfo{journal}{\emph{arXiv preprint arXiv:2311.16839}} (\bibinfo{year}{2024}).
\newblock


\bibitem[Zhou et~al\mbox{.}(2023)]%
        {zhou2023comprehensive}
\bibfield{author}{\bibinfo{person}{Ce Zhou}, \bibinfo{person}{Qian Li}, \bibinfo{person}{Chen Li}, \bibinfo{person}{Jun Yu}, \bibinfo{person}{Yixin Liu}, \bibinfo{person}{Guangjing Wang}, \bibinfo{person}{Kai Zhang}, \bibinfo{person}{Cheng Ji}, \bibinfo{person}{Qiben Yan}, \bibinfo{person}{Lifang He}, {et~al\mbox{.}}} \bibinfo{year}{2023}\natexlab{}.
\newblock \showarticletitle{{A Comprehensive Survey on Pretrained Foundation Models: A History from BERT to ChatGPT}}.
\newblock \bibinfo{journal}{\emph{arXiv preprint arXiv:2302.09419}} (\bibinfo{year}{2023}).
\newblock


\bibitem[Zhou et~al\mbox{.}(2024)]%
        {zhou2024analyzing}
\bibfield{author}{\bibinfo{person}{Yiyang Zhou}, \bibinfo{person}{Chenhang Cui}, \bibinfo{person}{Jaehong Yoon}, \bibinfo{person}{Linjun Zhang}, \bibinfo{person}{Zhun Deng}, \bibinfo{person}{Chelsea Finn}, \bibinfo{person}{Mohit Bansal}, {and} \bibinfo{person}{Huaxiu Yao}.} \bibinfo{year}{2024}\natexlab{}.
\newblock \showarticletitle{{Analyzing and Mitigating Object Hallucination in Large Vision-Language Models}}. In \bibinfo{booktitle}{\emph{Proceedings of the 12th International Conference on Learning Representations}} (Vienna, Austria).
\newblock


\bibitem[Ziegler(2020)]%
        {kaggle2017books}
\bibfield{author}{\bibinfo{person}{Cai-Nicolas Ziegler}.} \bibinfo{year}{2020}\natexlab{}.
\newblock \showarticletitle{{Books Dataset}}.
\newblock \bibinfo{journal}{\emph{Kaggle}} (\bibinfo{year}{2020}).
\newblock
\urldef\tempurl%
\url{https://www.kaggle.com/datasets/saurabhbagchi/books-dataset}
\showURL{%
\tempurl}


\end{thebibliography}

\end{document}